\documentclass{article}[11pt] 
\usepackage[vmargin=1.3in,hmargin=0.8in]{geometry}

\usepackage{amsmath,amssymb,amsfonts,graphicx,amsthm,nicefrac,mathtools,bm}
\usepackage{hyperref} 
\usepackage[numbers]{natbib}
\usepackage[english]{babel}
\usepackage{times}
\usepackage[T1]{fontenc}
\usepackage{bbm,mdframed}
\usepackage[dvipsnames]{xcolor}
\usepackage{xspace}
\usepackage{rotating,multirow} 
\usepackage{tikz}
\usetikzlibrary{arrows}
\usetikzlibrary{shapes} 
\usepackage{algpseudocode,algorithm,algorithmicx} 
\usepackage{fancyhdr} 
\usepackage{tabularx}
\usepackage{subcaption}
\usepackage{multirow}

\usepackage{float}
\usepackage{soul} 
\usepackage{color}
\usepackage{enumitem}
\usepackage{comment}
\date{}
\author{\theauthor}
\title{\thetitle} 
\fancypagestyle{plain}{%
  \fancyhf{}%
  \lhead{\thepage}
}

\newcommand{\theauthor}{}
\newcommand{\thetitle}{ML-LOO: Detecting Adversarial Examples with Feature Attribution}

\makeatletter
\long\def\@makecaption#1#2{
        \vskip 0.8ex
        \setbox\@tempboxa\hbox{\small {\bf #1:} #2}
        \parindent 1.5em  
        \dimen0=\hsize
        \advance\dimen0 by -3em
        \ifdim \wd\@tempboxa >\dimen0
                \hbox to \hsize{
                        \parindent 0em
                        \hfil 
                        \parbox{\dimen0}{\def\baselinestretch{0.96}\small
                                {\bf #1.} #2
                                } 
                        \hfil}
        \else \hbox to \hsize{\hfil \box\@tempboxa \hfil}
        \fi
        }
\makeatother

\makeatletter
\newcommand\footnoteref[1]{\protected@xdef\@thefnmark{\ref{#1}}\@footnotemark}
\makeatother

\makeatletter \renewcommand*{\@fnsymbol}[1]{\ensuremath{\ifcase#1\or
    \dagger\or \dagger\or \ddagger\or \mathsection\or
    \mathparagraph\or \|\or **\or \dagger\dagger \or \ddagger\ddagger
    \else\@ctrerr\fi}} \makeatother

\begin{document}
\author{ 
Puyudi Yang$^{*}$, Jianbo Chen$^{\dagger}$, Cho-Jui Hsieh$^{\ddagger}$, Jane-Ling Wang$^{*}$, Michael I. Jordan$^{\dagger}$\\
University of California, Davis$^{*}$\\
University of California, Berkeley$^{\dagger}$\\
University of California, Los Angeles$^{\ddagger}$\\
}


  \maketitle
\begin{abstract}
Deep neural networks obtain state-of-the-art performance on a series of tasks. However, they are easily fooled by adding a small adversarial perturbation to input. The perturbation is often human imperceptible on image data. We observe a significant difference in feature attributions of adversarially crafted examples from those of original ones. Based on this observation, we introduce a new framework to detect adversarial examples through thresholding a scale estimate of feature attribution scores. Furthermore, we extend our method to include multi-layer feature attributions in order to tackle the attacks with mixed confidence levels. Through vast experiments, our method achieves superior performances in distinguishing adversarial examples from popular attack methods on a variety of real data sets among state-of-the-art detection methods. In particular, our method is able to detect adversarial examples of mixed confidence levels, and transfer between different attacking methods.
\end{abstract}

\section{Introduction}
Deep neural networks have achieved state-of-the-art performance on a variety of tasks, including image classification, object detection, speech recognition and machine translation. However, they have been shown to be vulnerable to adversarial examples. This incurs a security risk when DNNs are applied to sensitive areas such as finance, medicine, criminal justice and transportation. Adversarial examples are inputs to machine learning models that an attacker constructs intentionally to fool the model~\cite{openai2017attacking}. \citet{szegedy2013intriguing} observed that a visually indistinguishable perturbation in pixel space to the original image can alter the prediction of a neural network. Later, a series of papers \citep{goodfellow2014explaining, kurakin2016adversarial, moosavi2016deepfool,papernot2016limitations, carlini2017towards,madry2018towards, chen2017zoo, ilyas2018black, ilyas2018prior, liu2016delving, papernot2016transferability, papernot2017practical, brendel2018decisionbased, brunner2018guessing} designed more sophisticated methods for the worst-case perturbation within a restricted set, often a small $L_p$ ball with $p=0,2,\infty$. 

While a line of work tries to explain why adversarial examples exist~\cite{goodfellow2014explaining, tanay2016boundary, fawzi2018analysis, fawzi2016robustness}, a comprehensive analysis of underlying reasons has so far been an open problem, mainly because deep neural networks have complex function forms that a complete mathematical analysis is difficult to achieve. 
On the other hand, there has been a growing interest in developing tools for tackling the black-box nature of neural networks, among which feature attribution is a widely studied approach \citep{shrikumar2016not, bach2015pixel, simonyan2013deep, ribeiro2016should, li2015visualizing, baehrens2010explain, lipton2016mythos, li2016understanding, lundberg2017unified, vstrumbelj2010efficient, datta2016algorithmic, sundararajan2017axiomatic, chen2018lshapley}. Given a predictive model, such a method outputs, for each instance to which the model is applied, a vector of importance scores associated with the underlying features. Feature attribution has been used to improve transparency and fairness of machine learning models~\cite{ribeiro2016should, datta2016algorithmic}.

\begin{figure}[!bt]
\centering 
\includegraphics[width=0.9\linewidth]{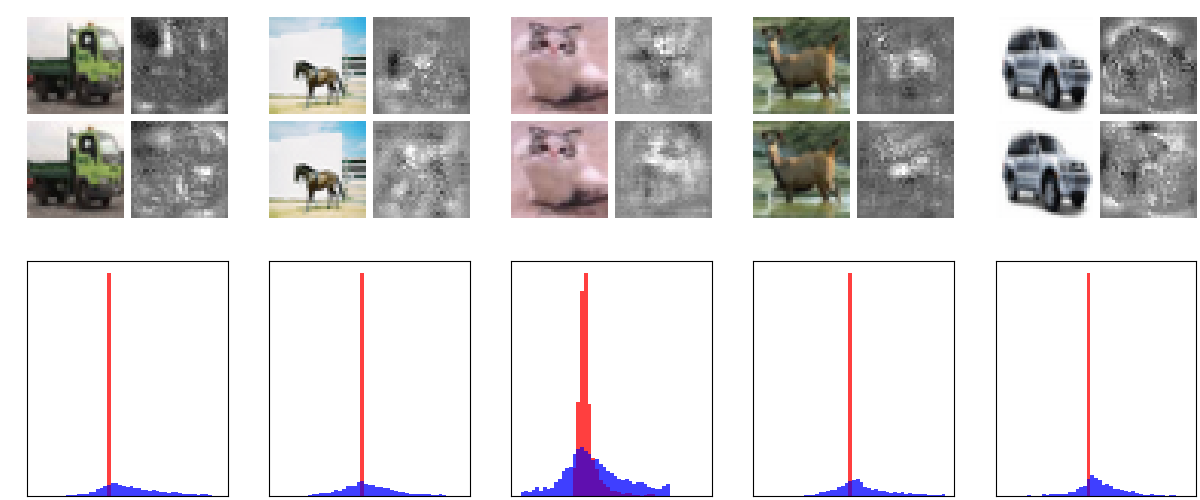} 
\includegraphics[width=1.0\linewidth]{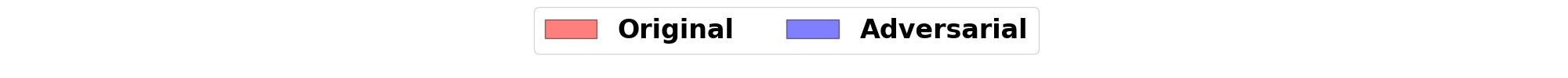} 
\caption{The first row shows the original CIFAR-10 examples and their corresponding feature attributions. The second row shows the adversarial examples and their corresponding feature attributions. The third row plots the histograms of the original and adversarial feature attributions.}
\label{fig:cifar10examples}
\end{figure}

In this paper, we investigate the application of feature attribution to detecting adversarial examples. In particular, we observe that the feature attribution map of an adversarial example near the boundary always differs from that of the corresponding original example. A motivating example is shown in Figure~\ref{fig:cifar10examples}, which demonstrates images in CIFAR-10 to be fed into a residual neural network and the corresponding feature attribution from Leave-One-Out (LOO)~
\citep{li2016understanding}. The latter interprets decisions from a
neural model by observing the effects on the
model of erasing each pixel of input 
before and after the worst-case perturbation by C\&W attack. While the perturbation on the original image is visually imperceptible, the feature attribution is altered drastically. We further observe that the difference can be summarized by simple statistics that characterize \textit{feature disagreement}, which are capable of distinguishing adversarial examples
from natural images. We conjecture that this is because adversarial attacks tend to perturb samples into an unstable region on the decision surface. 

The above observation led to an effective method for detecting adversarial examples near the decision boundary. On the other hand, there also exists adversarial examples in which the model has high confidence~\cite{carlini2017towards}. Previous work has observed several state-of-the-art detection methods are vulnerable to such attacks~\citep{lu2018on, athalye2018obfuscated}. However, we observe an interesting phenomenon: middle layers of neural networks still contain information of uncertainty even for high-confidence adversarial examples.
Based on this observation, we generalize our method to incorporate multi-layer feature attribution, where attribution scores for intermediate layers are computed without incurring extra model queries.

In numerical experiments, our method achieves superior performance in detecting adversarial examples generated from popular attack methods on MNIST, CIFAR-10 and CIFAR-100 among state-of-the-art detection methods.
We also show the proposed method is capable of detecting mixed-confidence adversarial examples, 
transferring between adversarial examples of different confidence levels, and adversarial examples generated by various types of attacks. 
\section{Related Work}
In this section, we review related work in feature attribution, adversarial attack, adversarial defense and detection. 
\paragraph{Feature attribution}
A variety of methods have been proposed to assign feature attribution scores. For each specific instance where the model is applied, an attribution method assigns an importance score for each feature, by approximating the target model via a linear model locally around the instance.
One popular class of methods assumes the differentiability of the model, and propagates the prediction to features through gradients. Examples include direct use of gradient (Saliency Map) \cite{simonyan2013deep}, Layer-wise Relevance Propagation (LWRP) \cite{bach2015pixel} and its improved version DeepLIFT \cite{shrikumar2016not}, and Integrated Gradients \cite{sundararajan2017axiomatic}. 

Another class is perturbation-based and thus model-agnostic. 
Given an instance, multiple perturbed samples are generated by masking different groups of features with a pre-specified reference value.
The feature attribution of the instance is computed according to the prediction scores of a model on these samples. Popular perturbation based methods include Leave-One-Out~\cite{zeiler2014visualizing, li2016understanding}, LIME \cite{ribeiro2016should} and KernelSHAP \cite{lundberg2017unified}. 

It has been observed in~\citet{ghorbani2017interpretation} that gradient-based feature attribution maps are sensitive to small perturbations. 
Adversarial attack to feature attribution is designed to characterize the fragility. On the contrary, robustness of an attribution method has been observed on a robust model. In fact, \citet{yeh2019sensitive} observed that gradient based explanations of an adversarially trained network are less sensitive, and \citet{chalasani2018adversarial} established theoretical results for the robustness of attribution map on an adversarially trained logistic regression. These observations indicate that the sensitivity of a feature attribution might be rooted in the sensitivity of the model, instead of the attribution method. This motivates the detection of adversarial examples via attribution methods.
\paragraph{Adversarial attack}
Adversarial attacks try to alter, with minimal perturbation, the prediction of an original instance from a given model, which leads to adversarial examples. Adversarial examples can be categorized as targeted or untargeted, depending on whether the goal is to classify the perturbed instance into a given target class or an arbitrary class different from the correct one. Attacks also differ by the type of distance they use to characterize minimal perturbation. $L_\infty, L_0,$ and $L_2$ distances are the most commonly used distances. Fast Gradient Sign Method (FGSM) by \citet{goodfellow2014explaining} is an efficient method to minimize the $L_\infty$ distance. \citet{kurakin2016adversarial} and \citet{madry2018towards} proposed $L_\infty$-PGD (BIM), an iterative version of FGSM, which achieves a higher success rate with a smaller size of perturbation.  
DeepFool presented by \citet{moosavi2016deepfool} minimizes $L_2$ distance through an iterative linearization procedure. \citet{carlini2017towards} proposed effective algorithms to generate adversarial examples for each of the three distances. In particular, \citet{carlini2017towards} proposed a loss function that is capable of controlling the confidence level of adversarial examples. The Jacobian-based Saliency Map Attack (JSMA) by \cite{papernot2016limitations} is a greedy method for perturbation with $L_0$ metric. 
Recently, several black-box adversarial attacks that solely depend on probability scores or decisions have been introduced. \citet{chen2017zoo} and \citet{ilyas2018black, ilyas2018prior} introduced score-based methods using zeroth-order gradient estimation to craft adversarial examples. \citet{brendel2018decisionbased} introduced Boundary Attack, as a black-box method to minimize the $L_2$ distance, that does not need access to gradient information and relies solely on the model decision. We demonstrate in our experiments that our method is capable of detecting adversarial examples generated by these attacks, regardless of the distance, confidence level, or whether the gradient information is used.
\paragraph{Adversarial defense and detection}
To improve the robustness of neural networks, various approaches have been proposed to defend against adversarial attacks, including adversarial training \cite{goodfellow2014explaining, kurakin2016adversarial, madry2018towards, tramer2018ensemble,liu2019robgan}, distributional smoothing \cite{miyato2015distributional}, defensive distillation \cite{papernot2016distillation}, generative models \cite{song2018pixeldefend}, feature squeezing \cite{xu2018feature}, randomized models \cite{liu2018towards,lecuyer2019certified,liu2018adv}, and verifiable defense \cite{wong2018provable,dvijotham2018training}. These defenses often involve modifications in the training process of a model, which often require higher computational or sample complexity~\cite{schmidt2018adversarially}, and lead to loss of accuracy \cite{tsipras2018there}. 

Complimentary to the previous defending techniques, an alternative line of work focuses on screening out adversarial examples in the test stage without touching the training of the original model. Data transformations such as PCA have been used to extract features from the input and layers of neural networks for adversarial detection \cite{li2017adversarial, bhagoji2018enhancing,hendrycks2016early}. Alternative neural networks are used to classify adversarial and original images \cite{grosse2017statistical, gong2017adversarial, metzen2017detecting}. \citet{feinman2017detecting} proposed to use kernel density estimate (KD) and Bayesian uncertainty (BU) in hidden layers of the neural network for detection. \citet{ma2018characterizing} observed Local Intrinsic Dimension (LID) of hidden-layer outputs differ between the original and adversarial examples. 
\citet{lee2018simple} obtained the class conditional
Gaussian distributions with respect to lower-level and upper-level features of the deep neural network under Gaussian discriminant analysis, which result in a confidence score based on the Mahalanobis distance (MAHA), followed by a logistic regression model on the confidence scores to detect adversarial examples.
Through vast experiments, we show that our method achieves comparable or superior performance than these detection methods across various attacks. 
Furthermore, we show that our method achieves competitive performance for attacks with a varied confidence level, a setting where the other detection methods fail to work~\cite{lu2018on, athalye2018obfuscated}.

Most related to our work, \citet{tao2018attacks} proposed to identify neurons critical for individual attributes to detect adversarial examples, but their method is restricted to models in face recognition. Instead, our method is applicable across different types of image data. \citet{zhang2018detecting} proposed to identify adversarial perturbations by training a neural network on the saliency map of inputs. However, their method depends on additional neural networks, which are vulnerable to white-box attacks when attackers perturb the image to fool the original model and the new neural network simultaneously. 

\begin{figure}[!bt]
\centering 
\includegraphics[width=0.3\linewidth]{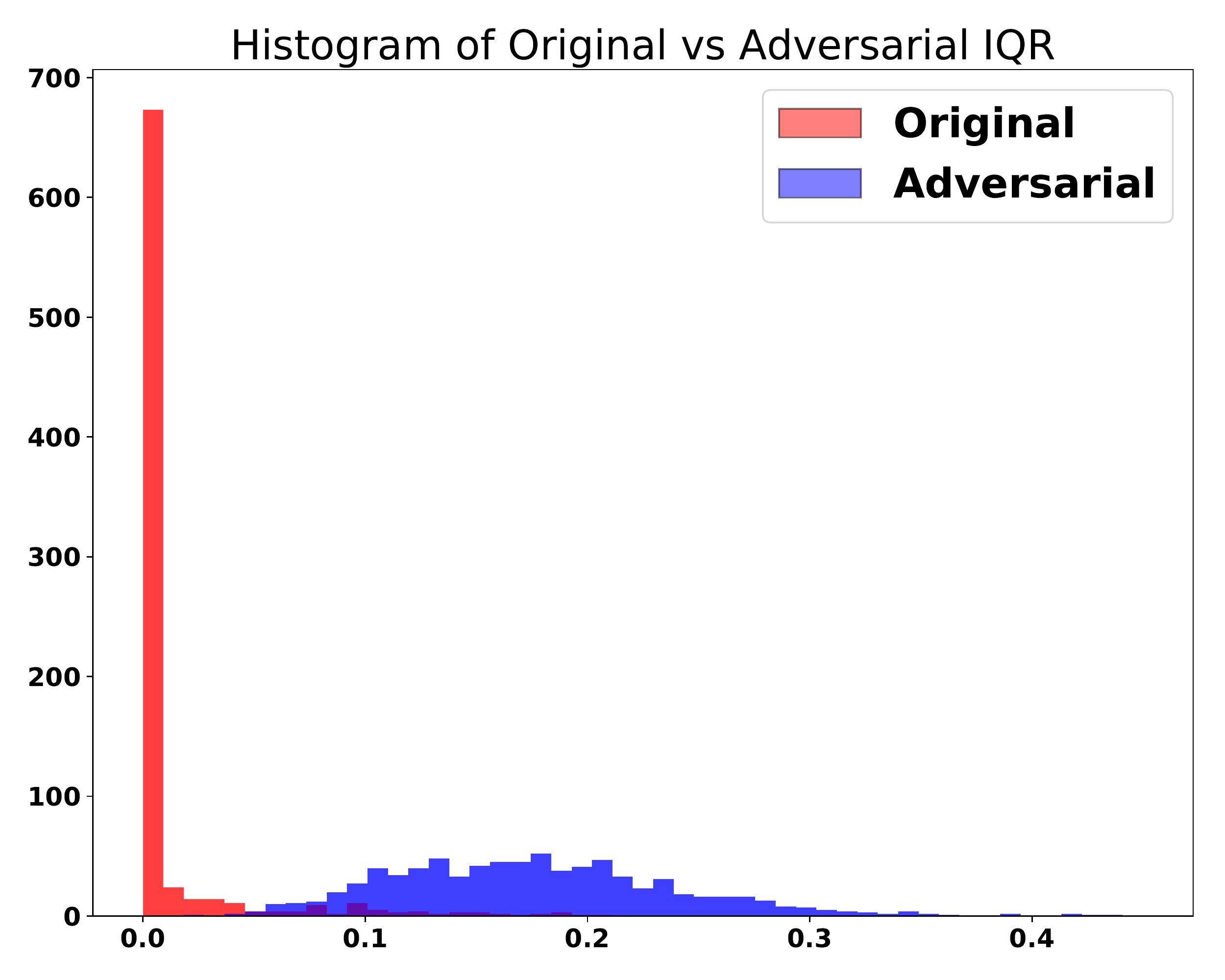} 
\includegraphics[width=0.3\linewidth]{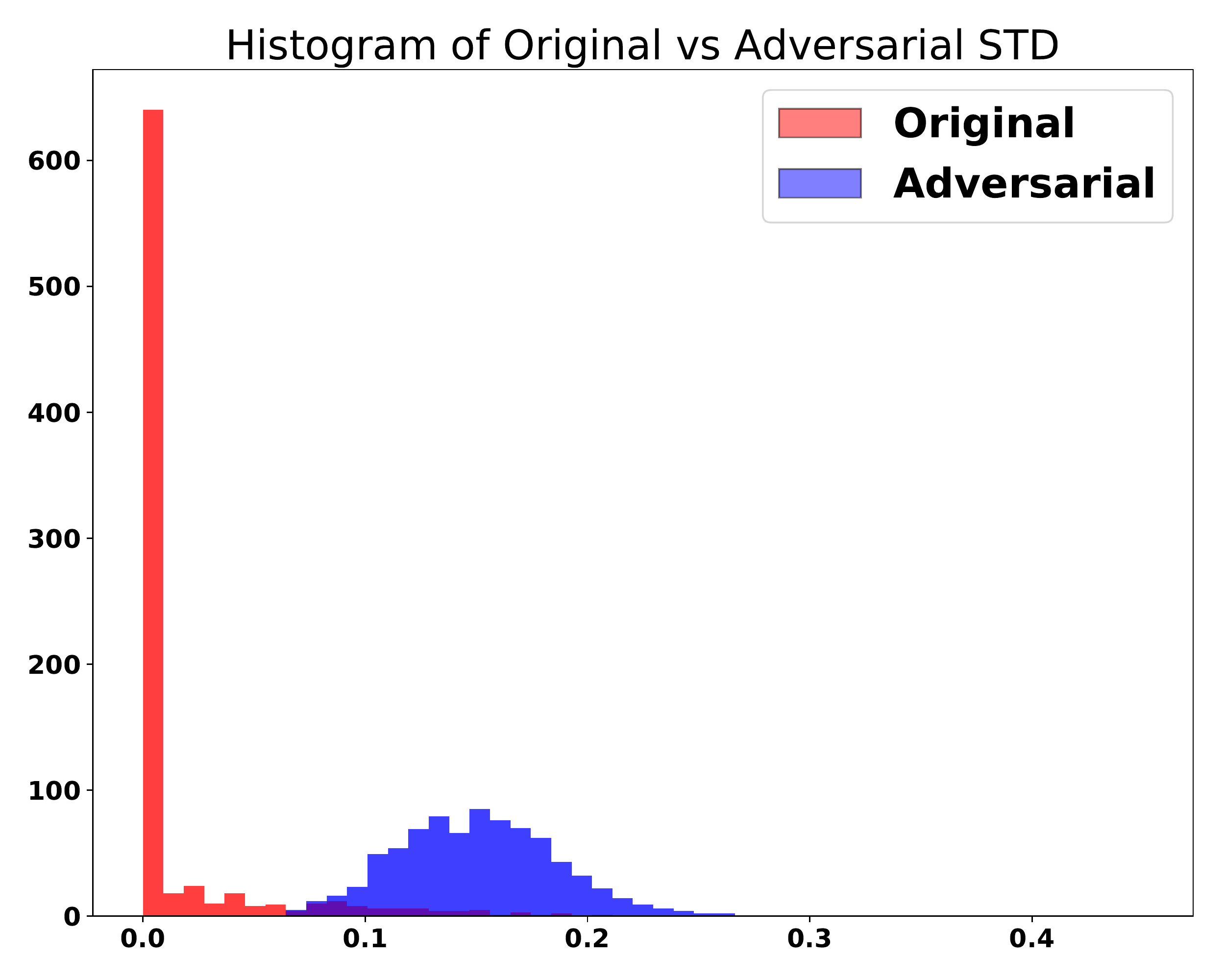} 
\includegraphics[width=0.3\linewidth]{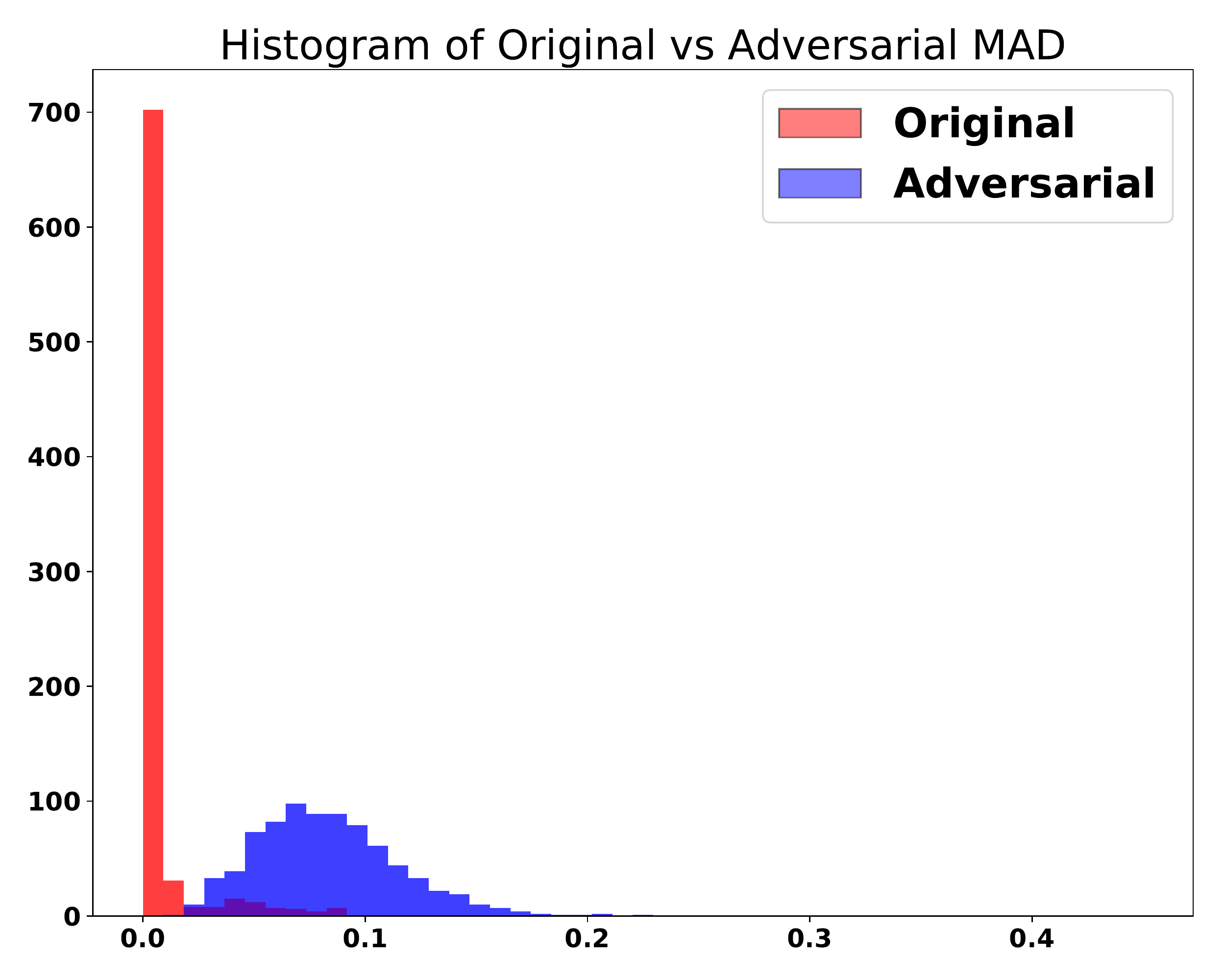} 
\caption{Histogram of dispersion measures}
\label{fig:iqr}
\end{figure}
\section{Adversarial detection with feature attribution}
\subsection{Feature attribution before and after perturbation}
Assume that the model is a function $f:\mathbb R^{d}\to [0,1]^C$ which maps an image $x$ of dimension $d = h\times w \times c$ to a probability vector $f(x)$ of dimension $C$, where $C$ is the number of classes. A feature attribution method $\phi$ maps an input image $x\in\mathbb R^{d}$ to an attribution vector of the same shape as the image: $\phi(x)\in\mathbb R^{d}$, such that the $i$th dimension of $\phi(x)$ is the contribution of feature $i$ in the prediction of the model on the specific image $x$. We suppress the dependence of $\phi$ on the model $f$ for notational convenience. We focus on the leave-One-Out (LOO) method~\cite{zeiler2014visualizing, li2016understanding} throughout the paper, which assigns to each feature the reduction in the probability of the selected class when the feature in consideration is masked by some reference value, e.g. $0$. 
Denoting the example with the $i$th feature masked by $0$ as $x_{(i)}$, LOO defines $\phi$ as
\begin{equation}
\phi(x)_i := f(x)_c - f(x_{(i)})_c, \text{ where } c = \arg\max_{j\in C}f(x)_j.
\end{equation}
Adversarial attacks aim to change the prediction of a model with minimal perturbation of a sample, so that human is not able to detect the difference between an original image $x$ and its perturbed version $x'$. Yet we observed that $\phi$ is sensitive to the small difference between $x$ and $x'$. Figure~\ref{fig:cifar10examples} shows the attribution maps $\phi(x), \phi(x')$ with the original image $x$ and its adversarially perturbed counterpart $x'$ by C\&W attack. 
Even with human eyes, we can observe an explicit difference in the attribution maps of the original and adversarial images. 
In particular, adversarial images have a larger dispersion in its importance scores, as demonstrated in Figure~\ref{fig:cifar10examples}. We comment here that our proposed framework of adversarial detection via feature attribution is generic to popular feature attribution methods. As an example, we show the performance of Integrated Gradients~\cite{sundararajan2017axiomatic} for adversarial detection in Appendix~\ref{app:ig}. LOO achieves the best performance among all attribution methods across different data sets.
\subsection{Quantify the dispersion in feature attribution maps}
Motivated by the apparent differences in the distributions of importance scores between the original and adversarial images, as demonstrated in Figure~\ref{fig:cifar10examples}, we propose to use measures of statistical dispersion in feature attribution to detect adversarial examples. In particular, we tried standard deviation (STD), median absolute deviation (MAD), which is the median of absolute differences between entries and their median, and interquartile range (IQR), which is the difference between the 75th percentile and the 25th percentile among all entries of $\phi(x)\in\mathbb R^{d}$:
\begin{equation}
\text{IQR}(\phi(x)) = Q_{\phi(x)}(0.75) - Q_{\phi(x)}(0.25), 
\text{ where } Q_{\phi(x)}(p):=\min\{\beta: \frac{\# \{i:\phi(x)_i < \beta\}}{d} \geq p\}.
\end{equation}
We observe there is a larger dispersion, which we call \textit{feature disagreement}, between feature contribution to a model for an adversarially perturbed image. The difference is universal across different images. Figure~\ref{fig:iqr} compares the histograms of these three dispersion measures of feature attributions for ResNet on natural test images from CIFAR-10 with those on adversarially perturbed images, where the adversarial perturbation is carried out by C\&W Attack. We can see there is a significant difference in the distributions of STD, MAD and IQR between natural and adversarial images.  
A majority of adversarially perturbed images have a larger dispersion in feature attribution than an arbitrary natural image, besides the corresponding original images. 
We propose to distinguish adversarial images from natural images by thresholding the IQR of feature attribution maps. In Appendix~\ref{sec:roc_stat}, we show the ROC curves of adversarial detection using the three dispersion measures on CIFAR-10 data set with ResNet across three different attacks. All the three measures yield competitive performance. We stick to IQR for the rest of the paper, which is robust and has a slightly superior performance among the three.


\begin{figure}[!bt]
\centering 
\includegraphics[width=0.3\linewidth]{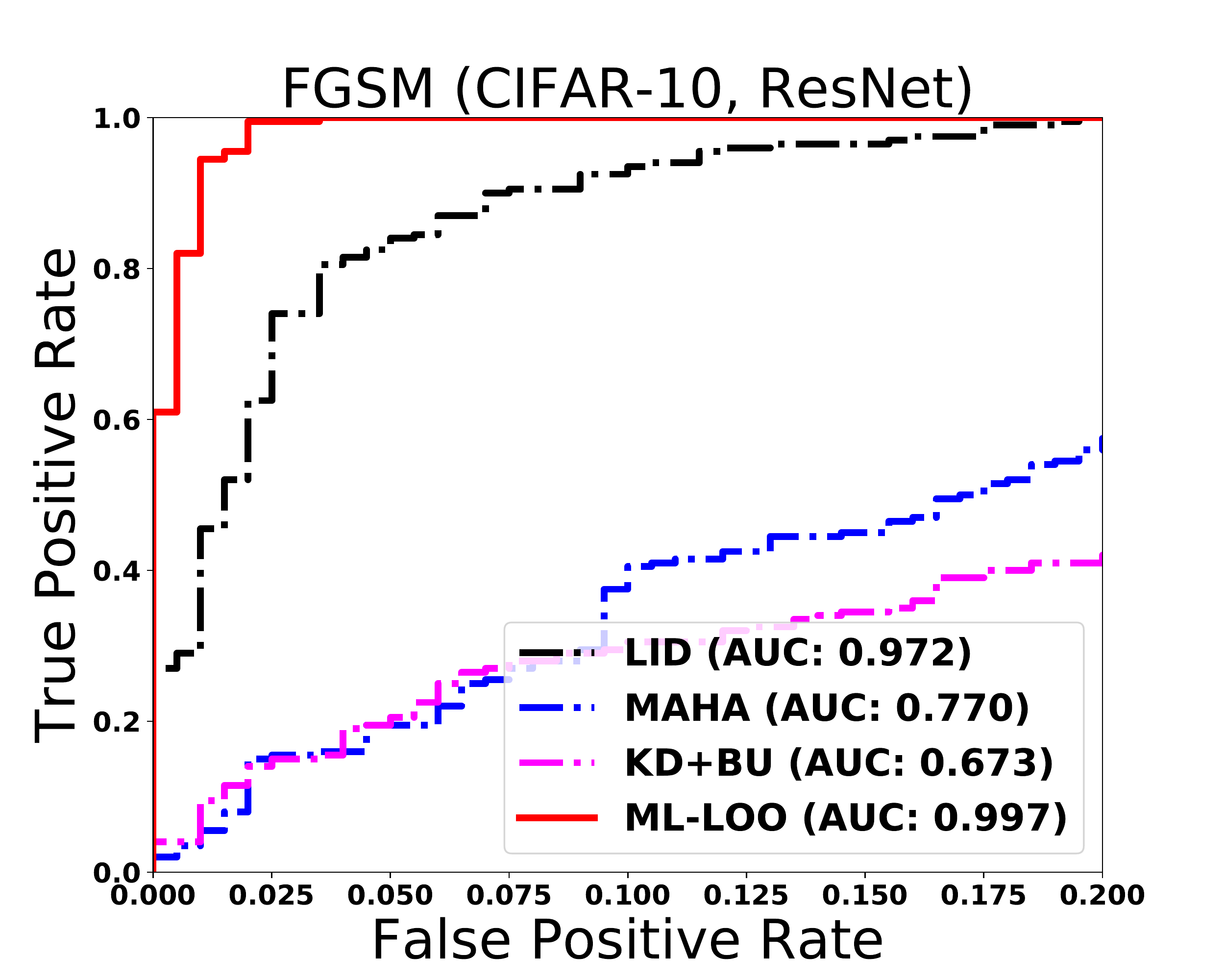} 
\includegraphics[width=0.3\linewidth]{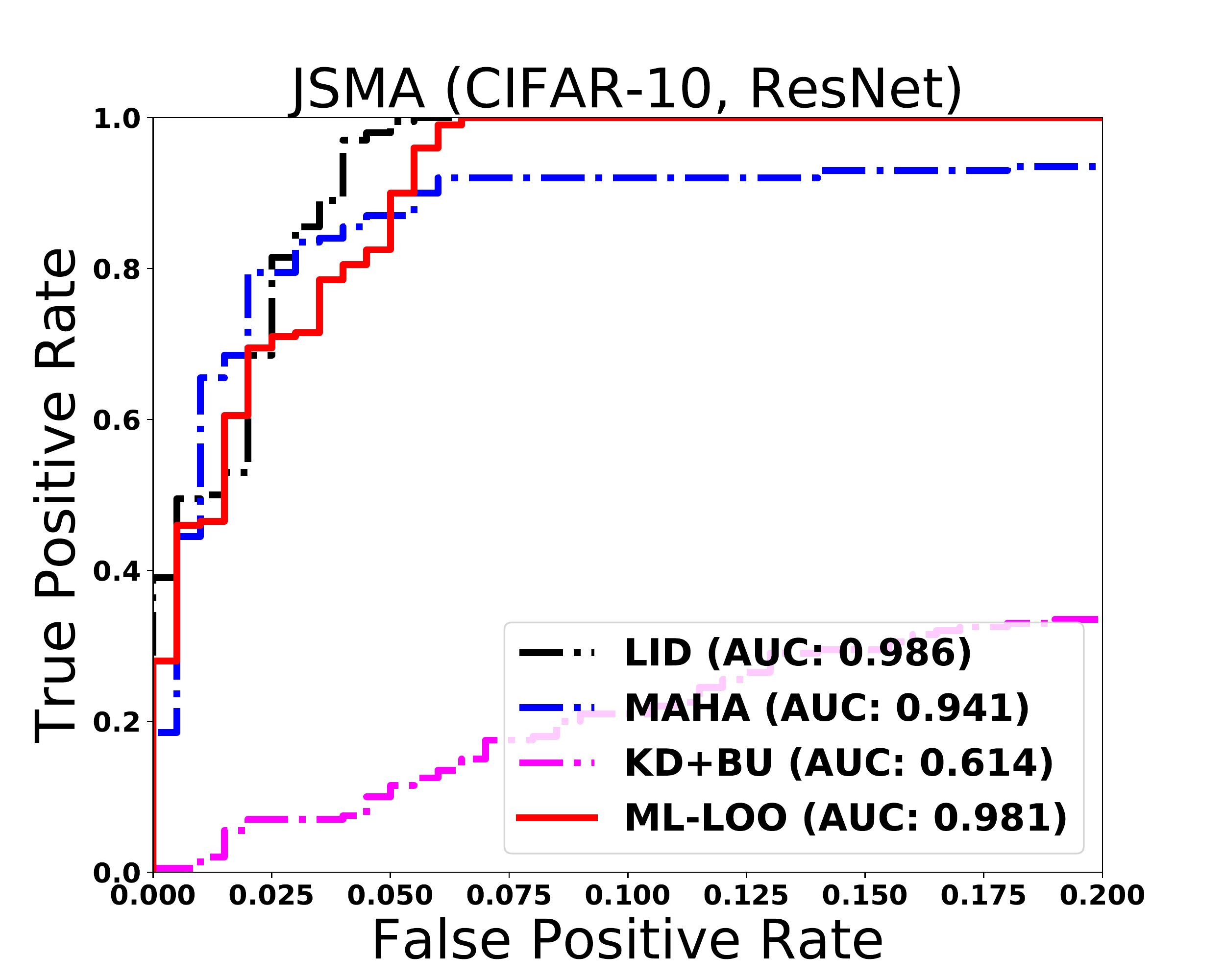} 
\includegraphics[width=0.3\linewidth]{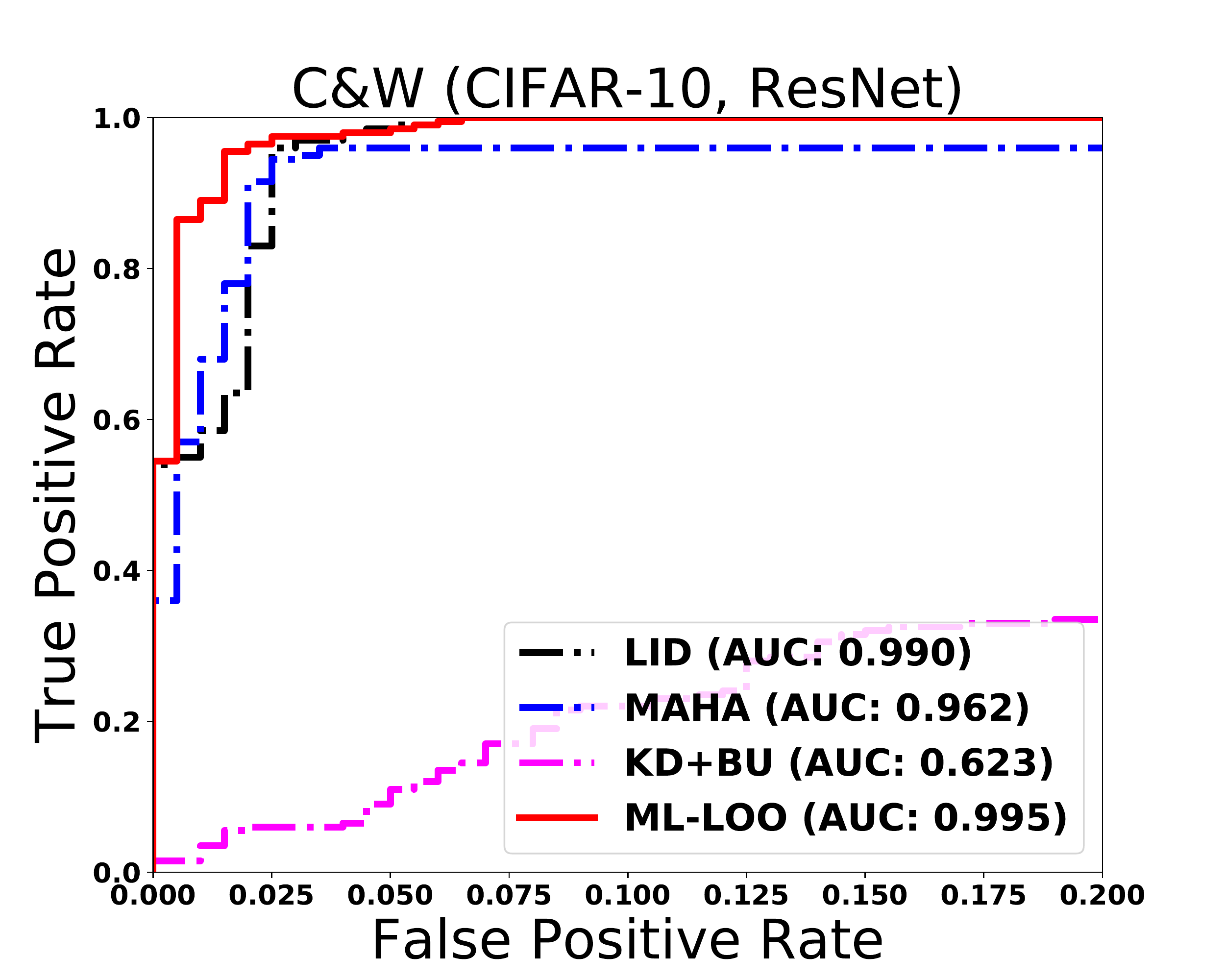} 
\includegraphics[width=0.3\linewidth]{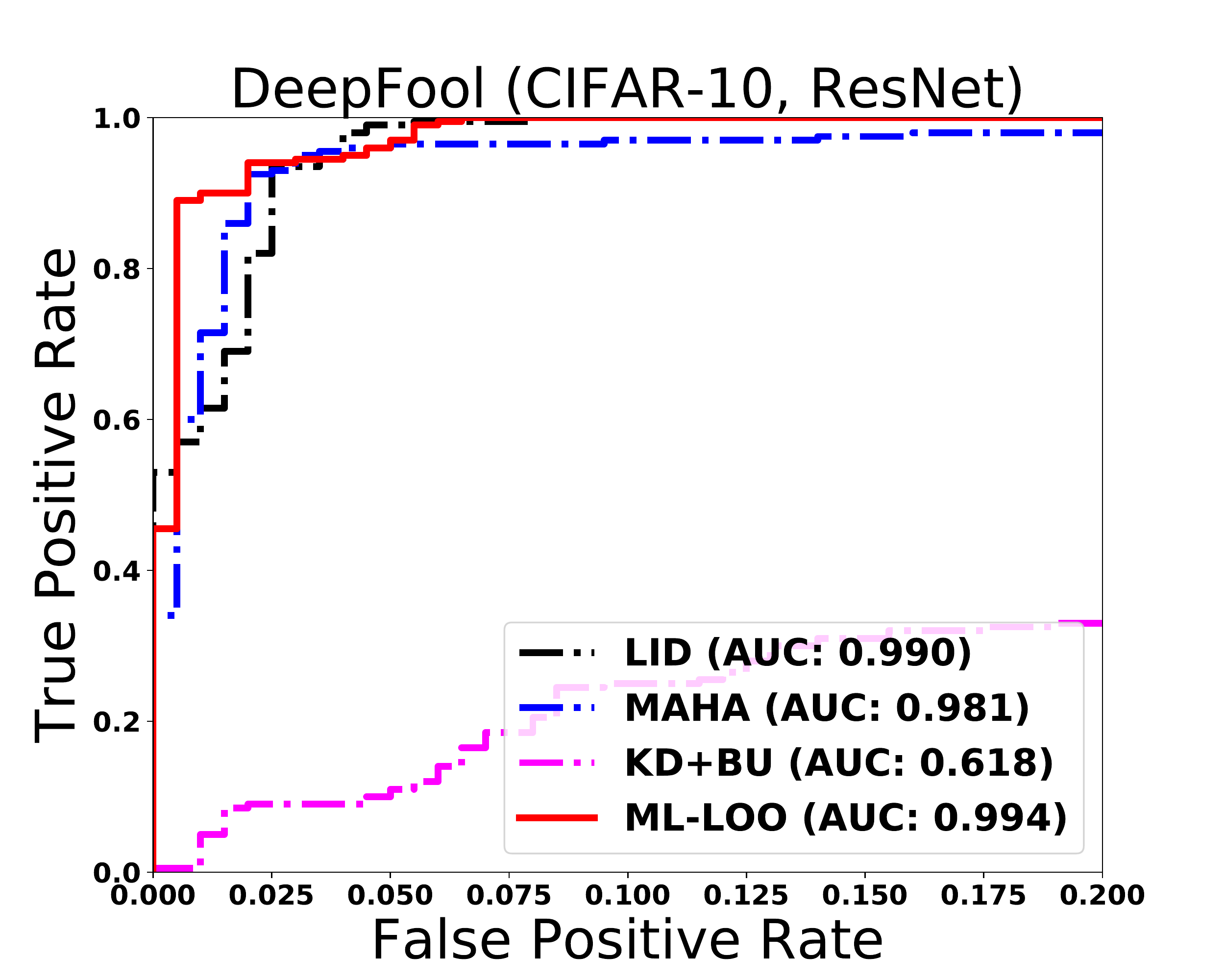} 
\includegraphics[width=0.3\linewidth]{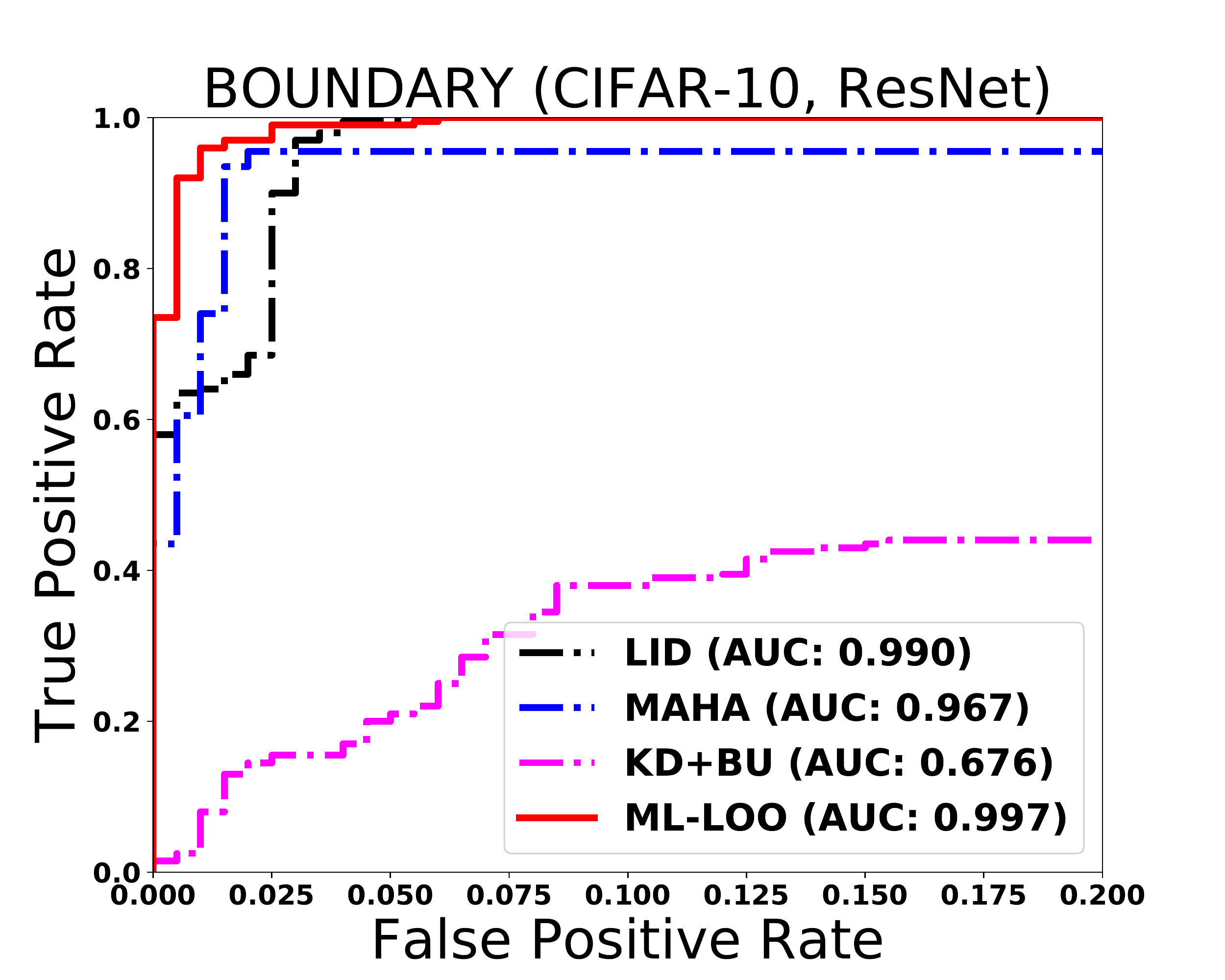} 
\includegraphics[width=0.3\linewidth]{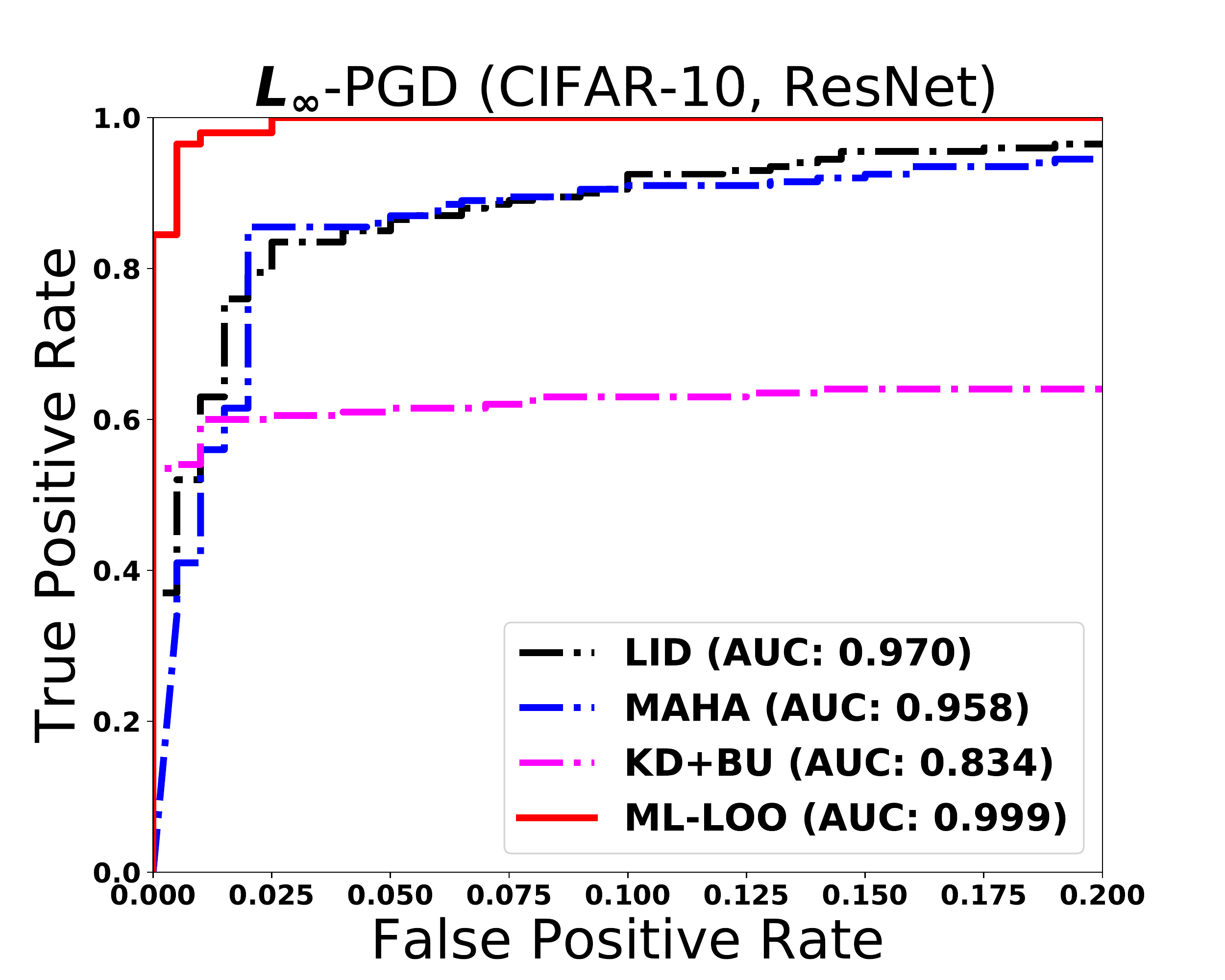}
\caption{ROC curves of detection methods on CIFAR-10 dataset with ResNet}
\label{fig:CIFAR10RESNET}
\end{figure}

\subsection{Extension to multi-layer LOO: detection of attacks with mixed confidence levels}
\citet{carlini2017towards} proposed the following objective to generate adversarial images with small $L_2$ perturbation.
\begin{equation}\label{eq:cw_hc}
\min_w \|x' - x\|_2 + \alpha \max \{ F(x)_{y_{\text{true}}} - \max_{j\neq y_{\text{true}}} F(x')_j + c,0\}, 
\end{equation}
where $x' = 0.5(\tanh (w) + 1)$, $F$ maps an image to logits, $y_{\text{true}} = \arg\max F(x)$ is the original label, and $c$ is a hyperparameter for tuning confidence. Adversarial images with high confidence can be obtained by assigning a large value to $c$. The loss can be modified to generate $L_\infty$ constrained perturbation at different confidence levels as well~\cite{madry2018towards}. Recently, \citet{lu2018on} and \citet{athalye2018obfuscated} observed that LID has a poor performance when faced with adversarial examples at various confidence scales. In our experiments, a similar phenomenon is observed for several other state-of-the-art detection methods, including KD+BU and MAHA, as is shown in Figure~\ref{fig:cw}. This suggests that characterization of adversarial examples in related work may only hold true for adversarial examples near the decision boundary. IQR of feature attribution map, unfortunately, suffers from the same problem. 

To detect adversarial images with mixed confidence levels, we generalize our method to capture dispersion of feature attributions beyond the output layer of the model. 
For an adversarial example within a small neighborhood of its original example in the pixel space but achieving a high confidence at the output layer in a different class from the original one, the feature representation deviates away from that of its original example gradually along the layers. Thus, we expect neurons of middle layers contain uncertainty that can be captured by a feature attribution map.
We denote the map from input to an arbitrary neuron $n$ of an intermediate layer of the model by $f_n: \mathbb R^d\to \mathbb R$. The feature attribution of neuron $n$ is defined as $\phi_{f_n}(x): \mathbb R^d\to \mathbb R^d$, such that the $i$th entry quantifies the contribution of feature $i$ to neuron $n$. For Leave-One-Out (LOO), we have
$$\phi_{f_n}(x)_i = f_n(x) - f_n(x_{(i)}).$$
To coordinate the scale difference between different neurons, we fit a logistic regression for the dispersion of feature attribution from different neurons on a hold-out training set to distinguish adversarial images from original images. The multi-layer extension of our method is called 'ML-LOO'.

\begin{table}[!bt]
\caption{Performance of detection methods on different data sets, models and attack methods.}
\begin{center}
\resizebox{0.96\columnwidth}{!}{%
\begin{tabular}{|c|c|c|c|c|c|c||c|c|c|c||c|c|c|c|}
 \hline
 \multirow{3}{*}{Data} & \multirow{3}{*}{Model} &\multirow{3}{*}{Metric} & \multicolumn{12}{c|}{Attacks}\\
 \cline{4-15}
 &&& \multicolumn{4}{c||}{C\&W} & \multicolumn{4}{c|}{$L_\infty$-PGD}& \multicolumn{4}{c|}{FGSM}\\
 \cline{4-15}
 &&&KD+BU & LID & MAHA & ML-LOO & KD+BU & LID & MAHA & ML-LOO & KD+BU & LID & MAHA & ML-LOO \\
 \hline
 \multirow{4}{*}{MNIST} & \multirow{4}{*}{CNN} & AUC &0.893 &1.000 &0.957 &\textbf{1.000} &0.766 &0.902 &0.736 &\textbf{1.000} &0.744 &0.780 &0.967 &\textbf{1.000}\\
\cline{3-15}
& & TPR (FPR@0.01) &0.23 &\textbf{0.99} &0.94 &0.98 &0.09 &0.32 &0.01 &\textbf{0.99} &0.01 &0.09 &0.54 &\textbf{0.99}\\
\cline{3-15}
& & TPR (FPR@0.05) &0.46 &\textbf{0.99} &0.94 &0.98 &0.28 &0.58 &0.12 &\textbf{0.99} &0.15 &0.23 &0.92 &\textbf{0.99}\\
\cline{3-15}
& & TPR (FPR@0.10) &0.55 &\textbf{0.99} &0.94 &0.98 &0.34 &0.72 &0.29 &\textbf{0.99} &0.24 &0.40 &0.94 &\textbf{0.99}\\
 \hline
 \hline
 \multirow{8}{*}{CIFAR10} & \multirow{4}{*}{ResNet} & AUC &0.623 &0.990 &0.962 &\textbf{0.995} &0.834 &0.970 &0.958 &\textbf{0.999} &0.673 &0.972 &0.770 &\textbf{0.997}\\
\cline{3-15}
& & TPR (FPR@0.01) &0.01 &0.55 &0.57 &\textbf{0.86} &0.54 &0.52 &0.41 &\textbf{0.96} &0.04 &0.29 &0.04 &\textbf{0.82}\\
\cline{3-15}
& & TPR (FPR@0.05) &0.09 &\textbf{0.98} &0.95 &\textbf{0.98} &0.61 &0.85 &0.86 &\textbf{0.98} &0.20 &0.82 &0.16 &\textbf{0.99}\\
\cline{3-15}
& & TPR (FPR@0.10) &0.22 &\textbf{0.99} &0.95 &\textbf{0.99} &0.62 &0.91 &0.91 &\textbf{0.98} &0.29 &0.93 &0.38 &\textbf{0.99}\\
 \cline{2-15}
 & \multirow{4}{*}{DenseNet} & AUC &0.679 &0.958 &0.966 &\textbf{0.977} &0.955 &0.952 &0.768 &\textbf{0.997} &0.790 &0.706 &0.829 &\textbf{1.000}\\
\cline{3-15}
& & TPR (FPR@0.01) &0.06 &0.30 &\textbf{0.48} &0.33 &0.69 &0.51 &0.03 &\textbf{0.99} &0.17 &0.04 &0.00 &\textbf{0.99}\\
\cline{3-15}
& & TPR (FPR@0.05) &0.13 &0.79 &\textbf{0.91} &0.84 &0.74 &0.84 &0.23 &\textbf{0.99} &0.28 &0.12 &0.29 &\textbf{0.99}\\
\cline{3-15}
& & TPR (FPR@0.10) &0.22 &0.91 &0.94 &\textbf{0.98} &0.80 &0.88 &0.31 &\textbf{0.99} &0.41 &0.23 &0.51 &\textbf{0.99}\\
 \hline
 \hline
 \multirow{8}{*}{CIFAR100} & \multirow{4}{*}{ResNet} & AUC &0.637 &0.717 &0.945 &\textbf{0.967} &0.855 &0.984 &0.966 &\textbf{0.999} &0.773 &0.985 &0.875 &\textbf{1.000}\\
\cline{3-15}
& & TPR (FPR@0.01) &0.07 &0.00 &0.00 &\textbf{0.33} &0.59 &0.69 &0.48 &\textbf{0.94} &0.39 &0.48 &0.12 &\textbf{0.99}\\
\cline{3-15}
& & TPR (FPR@0.05) &0.16 &0.01 &0.52 &\textbf{0.70} &0.61 &0.94 &0.82 &\textbf{0.99} &0.49 &0.89 &0.43 &\textbf{0.99}\\
\cline{3-15}
& & TPR (FPR@0.10) &0.29 &0.01 &0.80 &\textbf{0.92} &0.64 &0.96 &0.92 &\textbf{0.99} &0.56 &0.99 &0.57 &\textbf{0.99}\\
\cline{2-15}
 & \multirow{4}{*}{DenseNet} &AUC &0.567 &0.727 &0.916 &\textbf{0.958} &0.549 &0.732 &0.947 &\textbf{0.971} &0.577 &0.751 &0.951 &\textbf{0.974}\\
\cline{3-15}
& & TPR (FPR@0.01) &0.02 &\textbf{0.07} &0.00 &\textbf{0.07} &0.01 &0.00 &0.00 &\textbf{0.21} &0.01 &0.01 &0.00 &\textbf{0.31}\\
\cline{3-15}
& & TPR (FPR@0.05) &0.17 &0.15 &0.61 &\textbf{0.66} &0.14 &0.01 &0.70 &\textbf{0.75} &0.17 &0.06 &0.77 &\textbf{0.81}\\
\cline{3-15}
& & TPR (FPR@0.10) &0.22 &0.26 &0.84 &\textbf{0.88} &0.20 &0.04 &0.91 &\textbf{0.96} &0.23 &0.18 &0.93 &\textbf{0.94}\\
 \hline
\end{tabular}
}
\resizebox{0.96\columnwidth}{!}{%
\begin{tabular}{|c|c|c|c|c|c|c||c|c|c|c||c|c|c|c|}
 \hline
 \multirow{3}{*}{Data} & \multirow{3}{*}{Model} &\multirow{3}{*}{Metric} & \multicolumn{12}{c|}{Attacks}\\
 \cline{4-15}
 &&& \multicolumn{4}{c||}{JSMA} & \multicolumn{4}{c|}{DeepFool}& \multicolumn{4}{c|}{Boundary}\\
 \cline{4-15}
 &&&KD+BU & LID & MAHA & ML-LOO & KD+BU & LID & MAHA & ML-LOO & KD+BU & LID & MAHA & ML-LOO \\
 \hline
 \multirow{4}{*}{MNIST} & \multirow{4}{*}{CNN} &AUC &0.886 &\textbf{1.000} &0.976 &\textbf{1.000} &0.901 &\textbf{1.000} &0.869 &\textbf{1.000} &0.905 &\textbf{1.000} &0.991 &\textbf{1.000}\\
\cline{3-15}
& & TPR (FPR@0.01) &0.30 &\textbf{1.00} &0.87 &0.99 &0.32 &\textbf{1.00} &0.04 &\textbf{1.00} &0.32 &\textbf{1.00} &0.79 &\textbf{1.00}\\
\cline{3-15}
& & TPR (FPR@0.05) &0.46 &\textbf{1.00} &0.94 &\textbf{1.00} &0.43 &\textbf{1.00} &0.36 &\textbf{1.00} &0.45 &\textbf{1.00} &0.98 &\textbf{1.00}\\
\cline{3-15}
& & TPR (FPR@0.10) &0.51 &\textbf{1.00} &0.95 &\textbf{1.00} &0.57 &\textbf{1.00} &0.59&\textbf{1.00} &0.55 &\textbf{1.00} &0.98 &\textbf{1.00}\\
 \hline
 \hline
 \multirow{8}{*}{CIFAR10} & \multirow{4}{*}{ResNet} & AUC &0.614 &\textbf{0.986} &0.941 &0.981 &0.618 &0.990 &0.981 &\textbf{0.994} &0.676 &0.990 &0.967 &\textbf{0.997}\\
\cline{3-15}
& & TPR (FPR@0.01) &0.01 &\textbf{0.49} &0.45 &0.46 &0.01 &0.57 &0.60 &\textbf{0.89} &0.03 &0.64 &0.60 &\textbf{0.92}\\
\cline{3-15}
& & TPR (FPR@0.05) &0.10 &\textbf{0.98} &0.87 &0.82 &0.10 &\textbf{0.99} &0.96 &0.96 &0.20 &\textbf{0.99} &0.94 &\textbf{0.99}\\
\cline{3-15}
& & TPR (FPR@0.10) &0.21 &\textbf{0.99} &0.90 &\textbf{0.99} &0.24 &\textbf{0.99} &0.96 &\textbf{0.99} &0.38 &\textbf{0.99} &0.94 &\textbf{0.99}\\
 \cline{2-15}
 & \multirow{4}{*}{DenseNet} & AUC &0.645 &0.937 &0.947 &\textbf{0.964} &0.646 &0.976 &\textbf{0.977} &0.976 &0.700 &\textbf{0.983} &0.981 &0.980\\
\cline{3-15}
& & TPR (FPR@0.01) &0.04 &0.14 &\textbf{0.41} &0.12 &0.03 &0.34 &\textbf{0.51} &0.24 &0.05 &0.58 &\textbf{0.62} &0.31\\
\cline{3-15}
& & TPR (FPR@0.05) &0.10 &0.67 &0.68 &\textbf{0.72} &0.09 &0.90 &\textbf{0.95} &0.82 &0.12 &\textbf{0.93} &0.91 &0.89\\
\cline{3-15}
& & TPR (FPR@0.10) &0.18 &0.86 &0.88 &\textbf{0.96} &0.17 &\textbf{0.98} &0.97 &\textbf{0.98} &0.23 &\textbf{0.98} &0.96 &\textbf{0.98}\\

 \hline
 \hline
 \multirow{8}{*}{CIFAR100} & \multirow{4}{*}{ResNet} & AUC &0.600 &0.740 &0.907 &\textbf{0.964} &0.610 &0.714 &0.953 &\textbf{0.970} &0.635 &0.732 &0.956 &\textbf{0.972}\\
\cline{3-15}
& & TPR (FPR@0.01) &0.00 &0.01 &0.00 &\textbf{0.42} &0.06 &0.00 &0.00 &\textbf{0.41} &0.07 &0.01 &0.00 &\textbf{0.49}\\
\cline{3-15}
& & TPR (FPR@0.05) &0.12 &0.14 &0.49 &\textbf{0.70} &0.14 &0.01 &0.56 &\textbf{0.74} &0.16 &0.07 &0.61 &\textbf{0.78}\\
\cline{3-15}
& & TPR (FPR@0.10) &0.27 &0.24 &0.77 &\textbf{0.91} &0.29 &0.01 &0.87 &\textbf{0.94} &0.30 &0.15 &\textbf{0.94} &0.93\\
 \cline{2-15}
 & \multirow{4}{*}{DenseNet} & AUC &0.567 &0.727 &0.916 &\textbf{0.958} &0.549 &0.732 &0.947 &\textbf{0.971} &0.577 &0.751 &0.951 &\textbf{0.974}\\
\cline{3-15}
& & TPR (FPR@0.01) &0.02 &\textbf{0.07} &0.00 &\textbf{0.07} &0.01 &0.00 &0.00 &\textbf{0.21} &0.01 &0.01 &0.00 &\textbf{0.31}\\
\cline{3-15}
& & TPR (FPR@0.05) &0.17 &0.15 &0.61 &\textbf{0.66} &0.14 &0.01 &0.70 &\textbf{0.75} &0.17 &0.06 &0.77 &\textbf{0.81}\\
\cline{3-15}
& & TPR (FPR@0.10) &0.22 &0.26 &0.84 &\textbf{0.88} &0.20 &0.04 &0.91 &\textbf{0.96} &0.23 &0.18 &0.93 &\textbf{0.94}\\
 \hline
 \hline
\end{tabular}
}
\end{center}
\label{table:auc_table2}
\end{table}
\section{Experiments}
We present an experimental evaluation of ML-LOO, and compare our method with several state-of-the-art detection methods. Then we consider the setting where attacks have different confidence levels. We further evaluate the transferability of various detection methods on an unknown attack. 
\subsection{Known Attacks}
\label{sec:known}
We compare our method with state-of-the-art detection algorithms including LID~\cite{ma2018characterizing}, Mahalanobis (MAHA)~\cite{lee2018simple}, and KD+BU~\cite{feinman2017detecting}, on three data sets: MNIST, CIFAR-10 and CIFAR-100, with the standard train/test split~\cite{chollet2015keras}. We used a convolutional network composed of 32-filter convolutional layers followed by a hidden dense layer with 1024 units for MNIST. Each convolutional layer was followed by a max-pooling layer. For both CIFAR-10 and CIFAR-100, we trained a 20-layer ResNet~\cite{he2016identity} and 121-layer DenseNet~\cite{huang2017densely} respectively. For each data set, we generated 2,000 adversarial examples from correctly classified test images by each attacking method. 
Among them, 1,000 adversarial images with the corresponding 1,000 natural images were used for the training process of LID, Mahalanobis and our method. Results are reported for the other 1,000 adversarial images with the corresponding natural images.
We consider the following attacking methods, grouped by the norms they are optimized for:
\vspace{-0.4cm}
\begin{itemize}[leftmargin=*]\setlength{\itemsep}{0pt}
\item $L_{\infty}$: FGSM~\cite{goodfellow2014explaining}, $L_{\infty}$-PGD~\cite{kurakin2016adversarial, madry2018towards}.
\item $L_2$: C\&W~\cite{carlini2017towards}, DeepFool~\cite{moosavi2016deepfool}, Boundary Attack~\cite{brendel2018decisionbased}.
\item  $L_0$: JSMA~\cite{papernot2016limitations}.
\end{itemize}
\vspace{-0.3cm}
Let true positive rate (TPR) be the proportion of adversarial images classified as adversarial, and false positive rate (FPR) be the proportion of natural images classified as adversarial. We report area under the curve (AUC) of the ROC curve as the performance evaluation as well as the true positive rates by thresholding FPR at 0.01,0.05 and 0.1, as it is practical to keep misclassified natural images at a low proportion.

The results are reported in Table~\ref{table:auc_table2}, and the ROC curves on CIFAR-10 with ResNet are shown in Figure~\ref{fig:CIFAR10RESNET}. The rest of the plots can be found in Appendix~\ref{sec:roc_plots2} and Appendix~\ref{sec:roc_plots}. ML-LOO shows superior performance over the other three detection methods across different data sets, models for all attacks optimized for $L_2$ and $L_\infty$ distances. By controlling FPR at 0.1, our method is able to find over 95\% adversarial examples generated by most existing attacks. 

\begin{table}[!bt]
\caption{
Top: Performance of detection methods trained with C\&W-MIX and tested on C\&W-LC, C\&W-HC and C\&W-MIX. Bottom: Performance of detection methods trained with $L_\infty$-PGD-MIX and tested on $L_\infty$-PGD-LC, $L_\infty$-PGD-HC and $L_\infty$-PGD-MIX.}
\begin{center}
\resizebox{0.96\columnwidth}{!}{%
\begin{tabular}{|c|c|c|c|c|c|c||c|c|c|c||c|c|c|c|}
 \hline
 \multirow{3}{*}{Data} & \multirow{3}{*}{Model} &\multirow{3}{*}{Metric} & \multicolumn{12}{c|}{Attacks}\\
 \cline{4-15}
 &&& \multicolumn{4}{c||}{C\&W MIX} & \multicolumn{4}{c||}{C\&W LC}& \multicolumn{4}{c|}{C\&W HC}\\
 \cline{4-15}
 &&&KD+BU & LID & MAHA & ML-LOO & KD+BU & LID & MAHA & ML-LOO & KD+BU & LID & MAHA & ML-LOO \\
 \hline
 \multirow{4}{*}{CIFAR10} & \multirow{4}{*}{ResNet} & AUC &0.620 &0.649 &0.640 &\textbf{0.840} &0.623 &0.445 &0.641 &\textbf{0.711} &0.829 &0.816 &0.966 &\textbf{0.988}\\
\cline{3-15}
& & TPR (FPR@0.01) &0.04 &0.01 &0.03 &\textbf{0.25} &0.01 &0.00 &0.01 &\textbf{0.12} &0.52 &0.23 &0.51 &\textbf{0.87}\\
\cline{3-15}
& & TPR (FPR@0.05) &0.17 &0.06 &0.14 &\textbf{0.42} &0.09 &0.06 &0.10 &\textbf{0.21} &0.59 &0.43 &0.90 &\textbf{0.94}\\
\cline{3-15}
& & TPR (FPR@0.10) &0.28 &0.19 &0.21 &\textbf{0.59} &0.22 &0.11 &0.16 &\textbf{0.34} &0.60 &0.62 &0.93 &\textbf{0.97}\\
 \hline
\end{tabular}
}
\resizebox{0.96\columnwidth}{!}{%
\begin{tabular}{|c|c|c|c|c|c|c||c|c|c|c||c|c|c|c|}
 \hline
 \multirow{3}{*}{Data} & \multirow{3}{*}{Model} &\multirow{3}{*}{Metric} & \multicolumn{12}{c|}{Attacks}\\
 \cline{4-15}
 &&& \multicolumn{4}{c||}{$L_\infty$-PGD-MIX} & \multicolumn{4}{c||}{$L_\infty$-PGD-LC}& \multicolumn{4}{c|}{$L_\infty$-PGD-HC}\\
 \cline{4-15}
 &&&KD+BU & LID & MAHA & ML-LOO & KD+BU & LID & MAHA & ML-LOO & KD+BU & LID & MAHA & ML-LOO \\
 \hline
 \multirow{4}{*}{CIFAR10} & \multirow{4}{*}{ResNet} & AUC &0.753 &0.812 &0.813 &\textbf{0.953} &0.606 &0.578 &0.578 &\textbf{0.767} &0.834 &0.935 &0.962 &\textbf{0.996}\\
\cline{3-15}
& & TPR (FPR@0.01) &0.20 &0.10 &0.11 &\textbf{0.60} &0.01 &0.01 &0.01 &\textbf{0.09} &0.54 &0.26 &0.46 &\textbf{0.89}\\
\cline{3-15}
& & TPR (FPR@0.05) &0.37 &0.36 &0.45 &\textbf{0.77} &0.12 &0.07 &0.04 &\textbf{0.23} &0.61 &0.67 &0.89 &\textbf{0.98}\\
\cline{3-15}
& & TPR (FPR@0.10) &0.46 &0.41 &0.56 &\textbf{0.84} &0.25 &0.17 &0.12 &\textbf{0.33} &0.62 &0.85 &0.91 &\textbf{0.99}\\
 \hline
 \hline
\end{tabular}
}
\end{center}
\label{table:auc_table_pgd_mix}
\end{table}
\begin{figure}[!bt]
\vspace{-0.5cm}
\centering 
\includegraphics[width=0.24\linewidth]{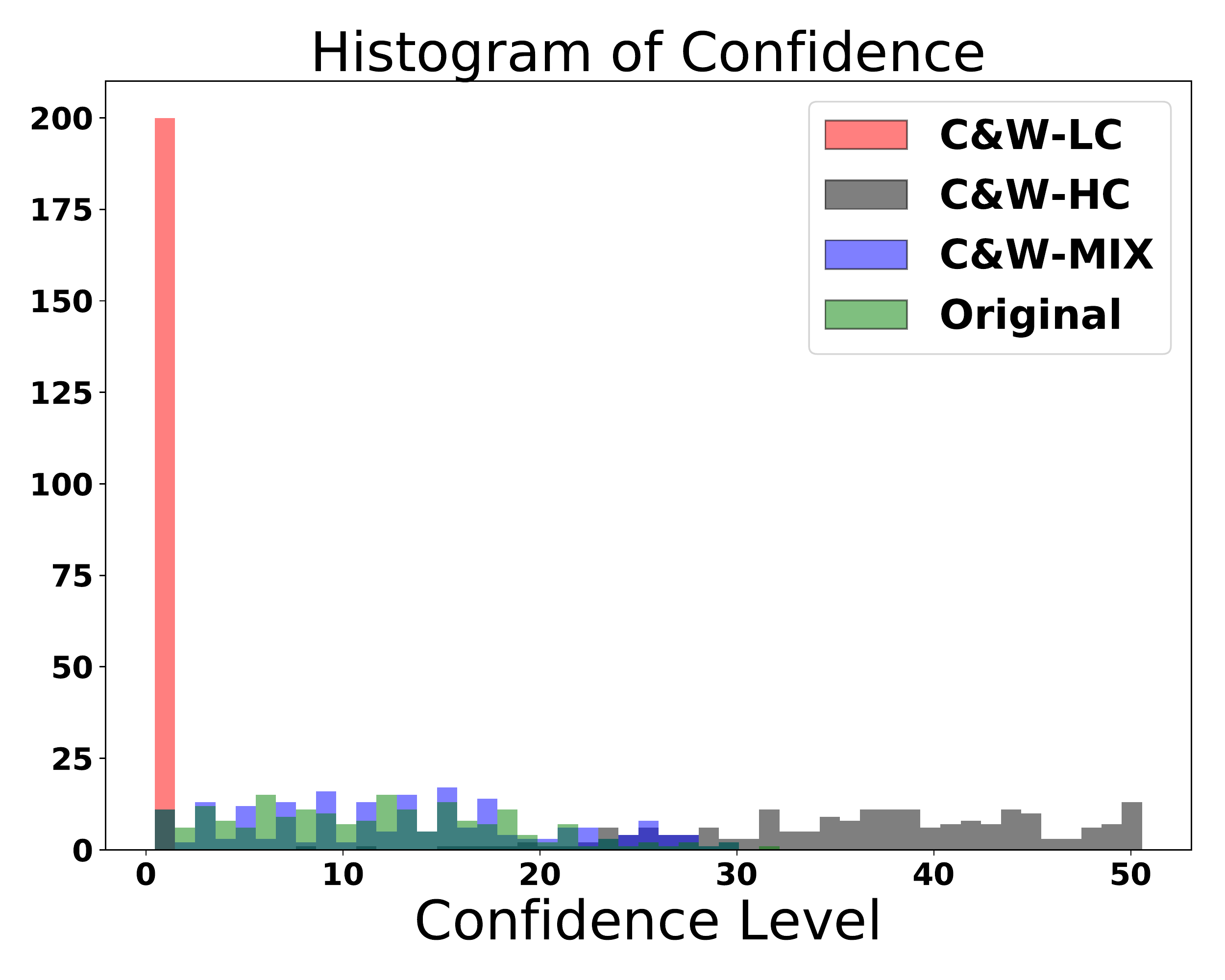} 
\includegraphics[width=0.24\linewidth]{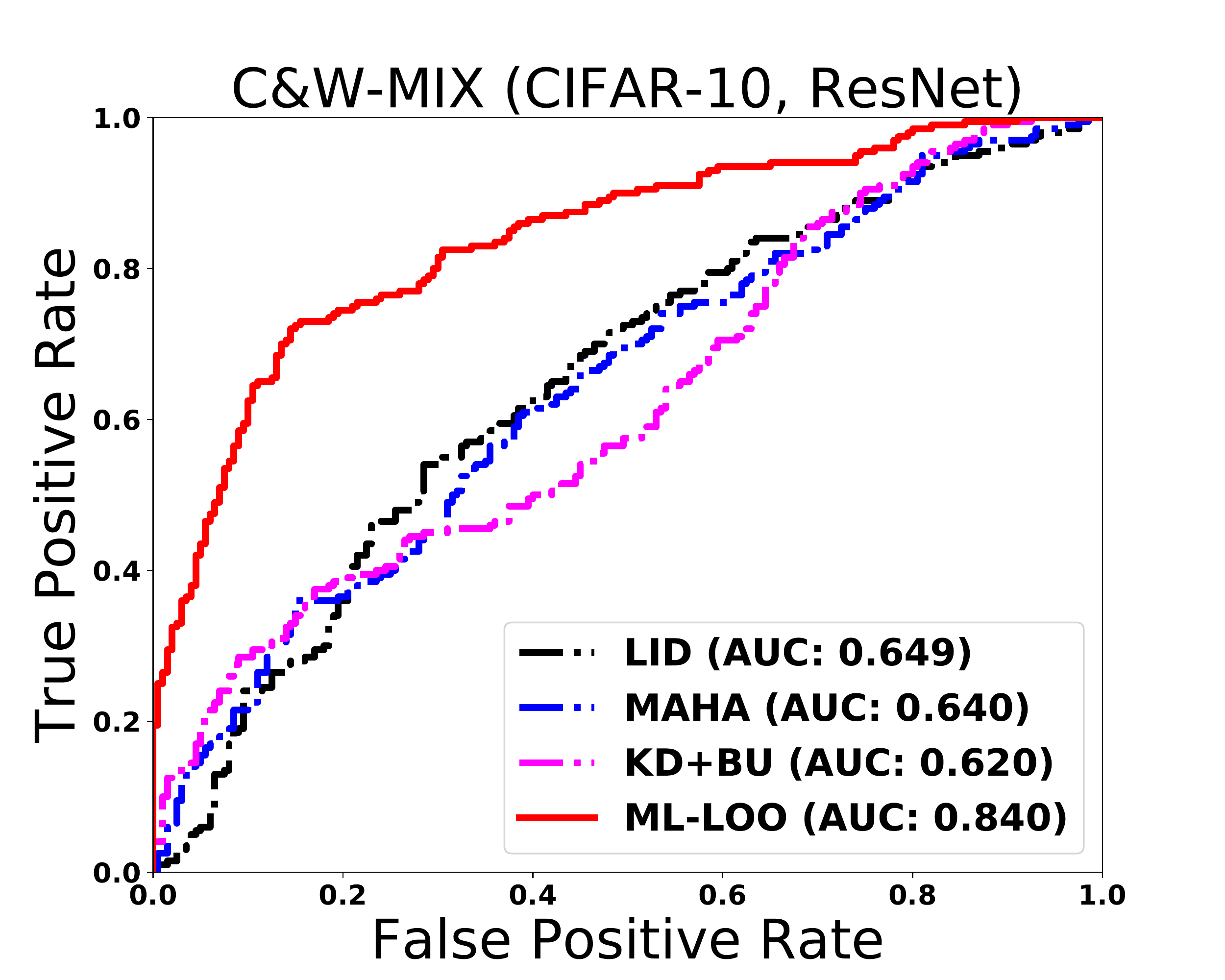}
\includegraphics[width=0.24\linewidth]{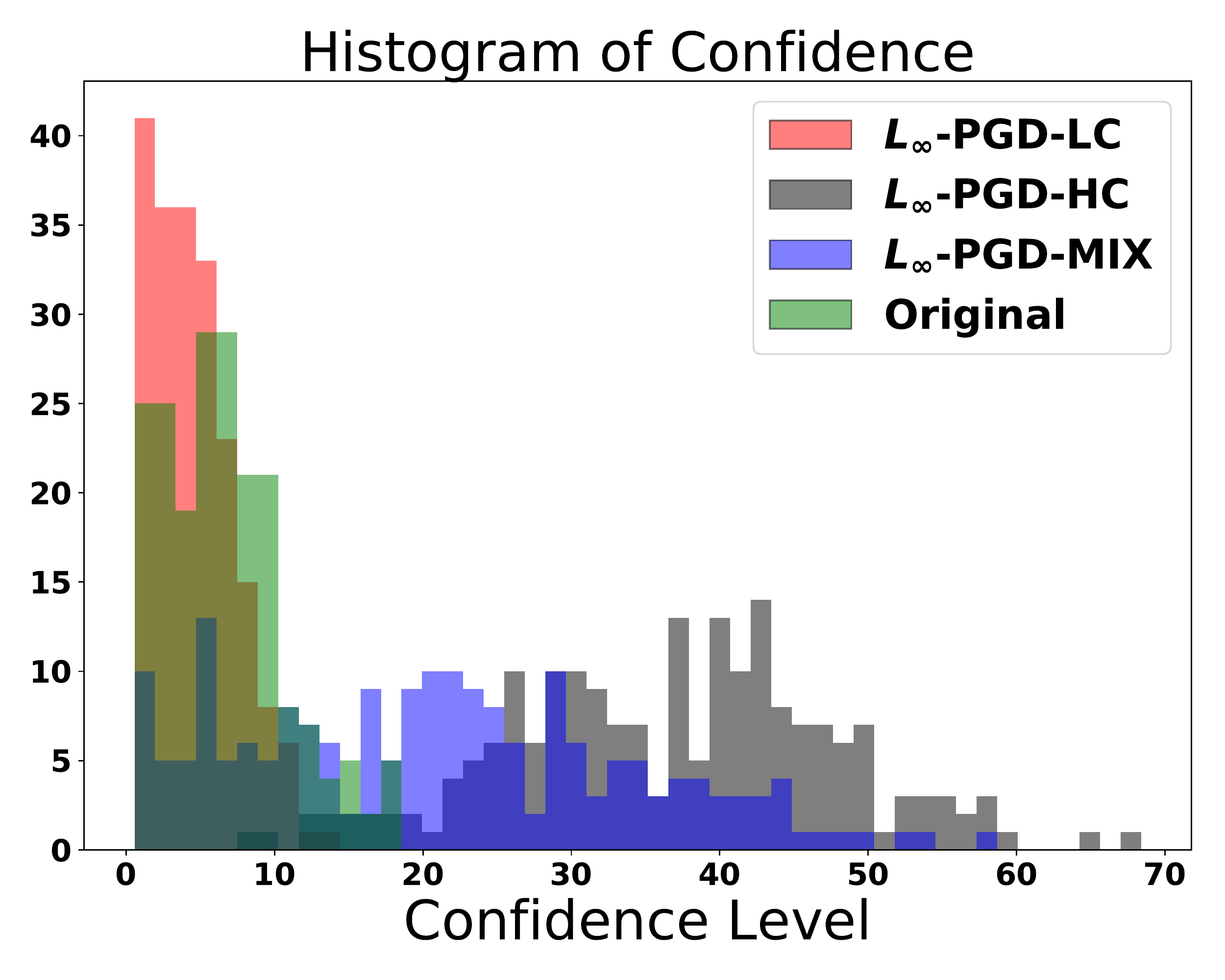}
\includegraphics[width=0.24\linewidth]{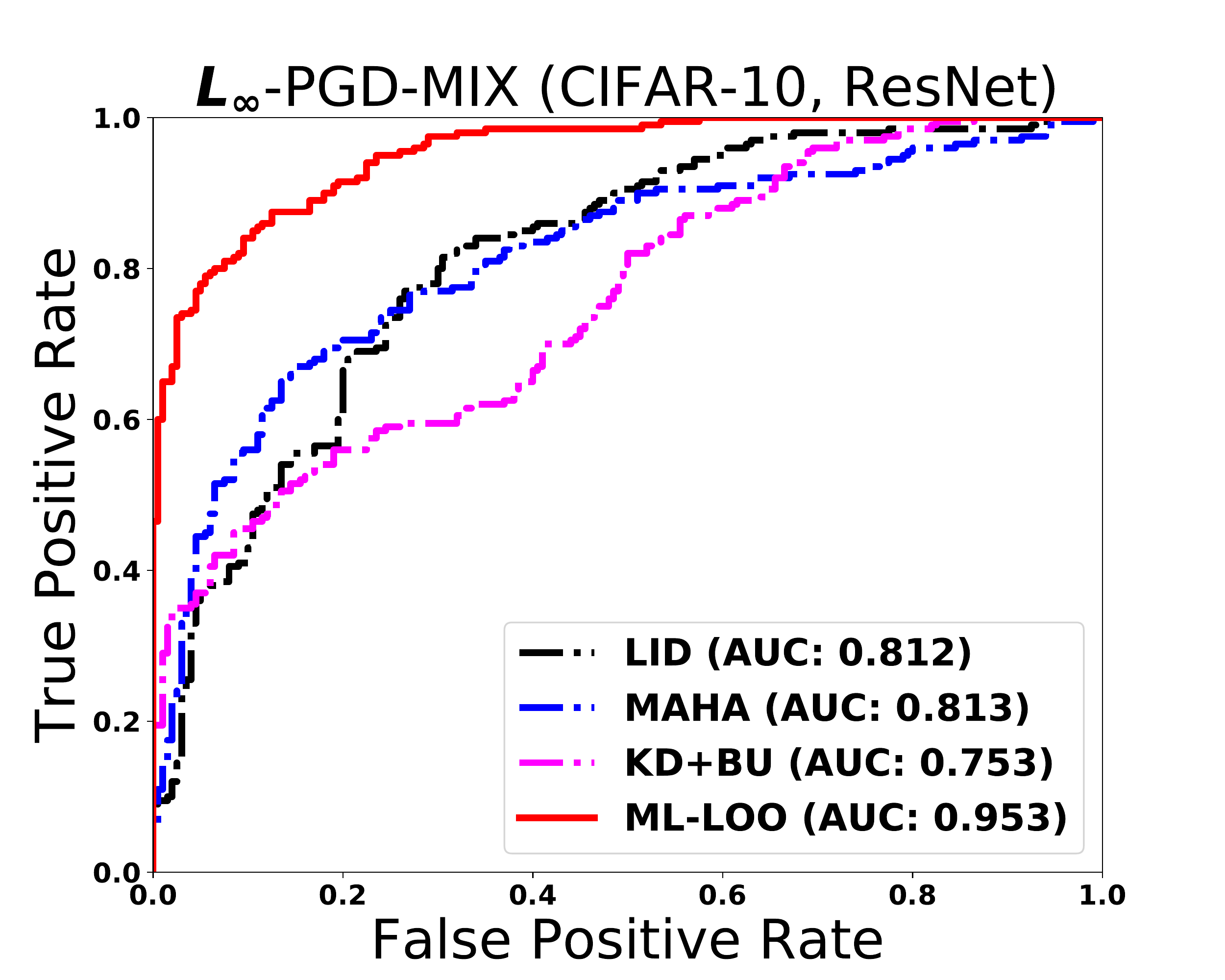} 
\caption{The left two figures plot the histogram of confidence levels of C\&W-LC, C\&W-HC, and C\&W-MIX, and the ROC curves of detection methods under C\&W-MIX attack. The right two figures plot the histogram of confidence levels of $L_\infty$-PGD-LC, $L_\infty$-PGD-HC, and $L_\infty$-PGD-MIX, and the ROC curves of detection methods under $L_\infty$-PGD-MIX attack.}
\label{fig:cw}
\end{figure} 
\subsection{Attacks with varied confidence levels}
\citet{lu2018on} and \citet{athalye2018obfuscated} observed that LID fails when the confidence level of adversarial examples generated from C\&W attack varies. We consider adversarial images with varied confidence levels for both $L_2$ and $L_\infty$ attacks. We use C\&W attack for optimizing $L_2$ distance, and adjust the confidence hyperparameter $c$ in Equation~\eqref{eq:cw_hc} to achieve mixed confidence levels. To achieve adversarial examples optimized for $L_\infty$ distance, we use $L_\infty$-PGD for optimizing $L_\infty$ distance, and vary the constraint $\varepsilon$ for different confidence levels.

\paragraph{C\&W Attack for optimizing $L_2$ distance}
We consider three settings for C\&W attack, low-confidence (C\&W-LC), mixed-confidence (C\&W-MIX) and high-confidence (C\&W-HC). 
We set the confidence parameter $c=0$ for C\&W-LC and $c=50$ for C\&W-HC. For mixed-confidence C\&W attack, we generate adversarial images from C\&W attack with the confidence parameter in Equation~\eqref{eq:cw_hc} randomly selected from $\{1, 3, 5, \cdots, 29\}$ when generating an adversarial image, so that the distribution of confidence levels for adversarial images is comparable with that of original images.
The confidence levels of images under the three settings, along with confidence levels of original images are shown in Figure~\ref{fig:cw}. The confidence level in Figure~\ref{fig:cw} is defined as $-\log (1-p)$, where $p$ is the probability score of the predicted class. 

We carried out the experiments on ResNet trained on CIFAR-10 using $1,000$ adversarial images generated from the mixed-confidence C\&W attack, together with the corresponding original images, as the training data for LID, Mahalanobis, KD+BU, and our method. We test the detection methods on a different set of original and adversarial images generated from three versions: low-confidence C\&W attack ($c=0$), high-confidence C\&W attack ($c=50$), and the mixed-confidence C\&W attack. 
Table~\ref{table:auc_table_pgd_mix} (Top) and Figure~\ref{fig:cw} (Left) show TPRs at different FPR thresholds, AUC, and the ROC curve. Mahalanobis, LID and KD+BU fail to detect adversarial examples of mixed-confidence effectively, while our method performs consistently better for adversarial images across the three settings.

\begin{table}[!bt]
\caption{Performance of detection methods trained with C\&W and transferred to $L_\infty$-PGD, FGSM, JSMA, Boundary and DeepFool.}
\begin{center}
\resizebox{\columnwidth}{!}{%
\begin{tabular}{|c|c|c|c|c|c|c||c|c|c|c||c|c|c|c||c|c|c|c||c|c|c|c||}
 \hline
 \multirow{3}{*}{Data} & \multirow{3}{*}{Model} &\multirow{3}{*}{Metric} & \multicolumn{20}{c|}{Attacks}\\
 \cline{4-23}
 &&& \multicolumn{4}{c||}{$L_\infty$-PGD} & \multicolumn{4}{c||}{DeepFool}& \multicolumn{4}{c||}{FGSM}& \multicolumn{4}{c||}{JSAM}& \multicolumn{4}{c||}{Boundary}\\
 \cline{4-23}
 &&&KD+BU & LID & MAHA & ML-LOO & KD+BU & LID & MAHA & ML-LOO & KD+BU & LID & MAHA & ML-LOO &KD+BU & LID & MAHA & ML-LOO &KD+BU & LID & MAHA & ML-LOO\\
 \hline
 \multirow{4}{*}{CIFAR10} & \multirow{4}{*}{ResNet} & AUC &0.753 &0.763 &0.818 &\textbf{0.879} &0.618 &0.990 &0.962 &\textbf{0.992} &0.673 &0.610 &0.730 &\textbf{0.796} &0.614 &0.984 &0.957 &\textbf{0.984} &0.676 &0.991 &0.964 &\textbf{0.994}\\
\cline{3-23}
& & TPR (FPR@0.01) &0.20 &0.08 &0.14 &\textbf{0.21} &0.01 &0.56 &0.61 &\textbf{0.72} &0.04 &\textbf{0.07} &0.06 &0.04 &0.01 &0.43 &0.44 &\textbf{0.45} &0.03 &0.56 &0.60 &\textbf{0.82}\\
\cline{3-23}
& & TPR (FPR@0.05) &0.37 &0.35 &0.45 &\textbf{0.48} &0.10 &0.96 &0.94 &\textbf{0.96} &0.20 &0.17 &\textbf{0.22} &0.14 &0.10 &\textbf{0.93} &0.91 &0.91 &0.20 &\textbf{0.99} &0.95 &0.97\\
\cline{3-23}
& & TPR (FPR@0.10) &0.46 &0.45 &0.60 &\textbf{0.65} &0.24 &0.98 &0.94 &\textbf{0.99} &0.29 &0.23 &0.34 &\textbf{0.37} &0.21 &0.98 &0.94 &\textbf{0.99} &0.38 &\textbf{0.99} &0.95 &\textbf{0.99}\\
 \hline
 \hline
\end{tabular}
}
\end{center}
\label{table:auc_table_transfer}
\end{table}
\begin{figure}[!bt]
\vspace{-0.5cm}
\centering 
\includegraphics[width=0.2\linewidth]{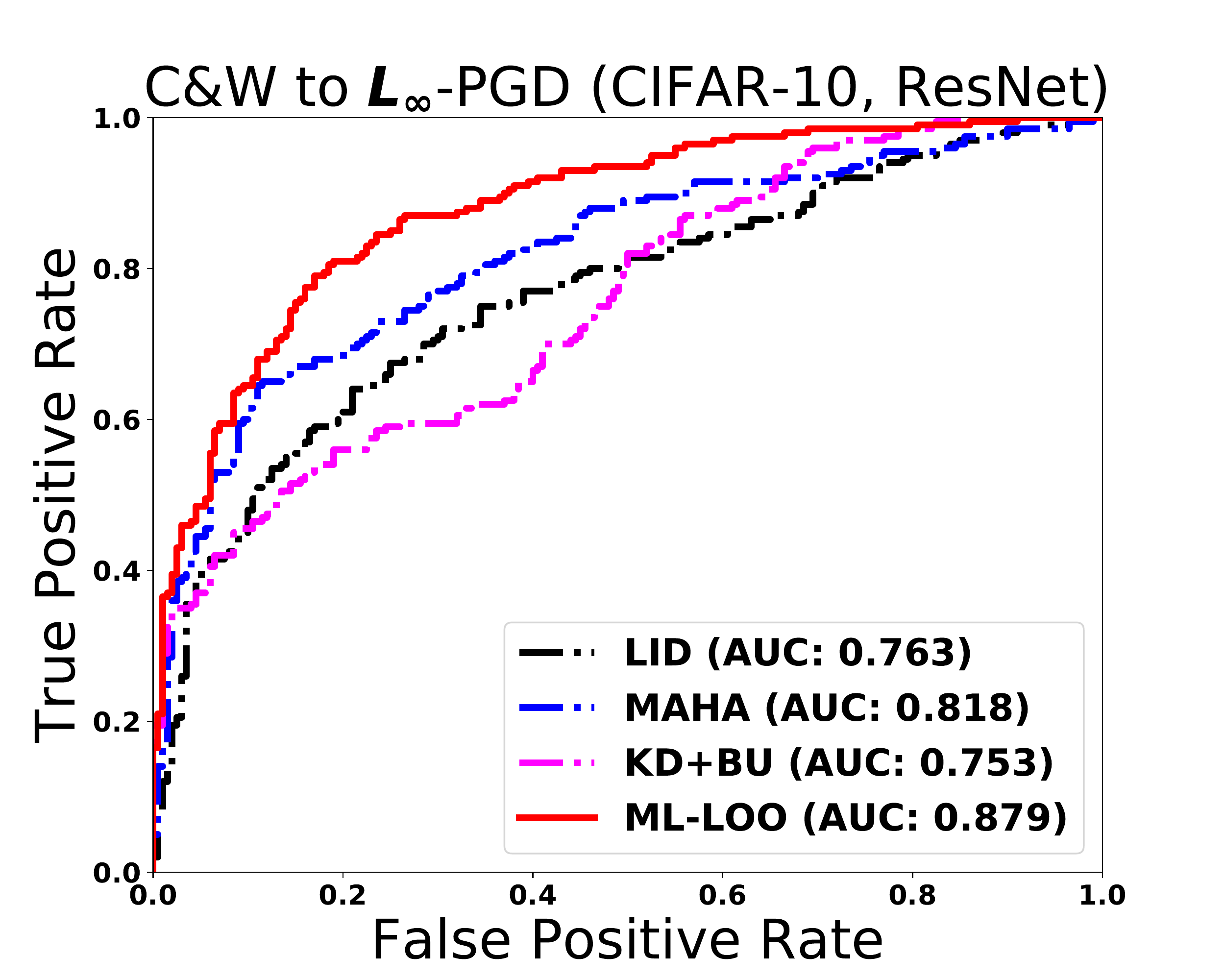}
\hspace{-3mm}
\includegraphics[width=0.2\linewidth]{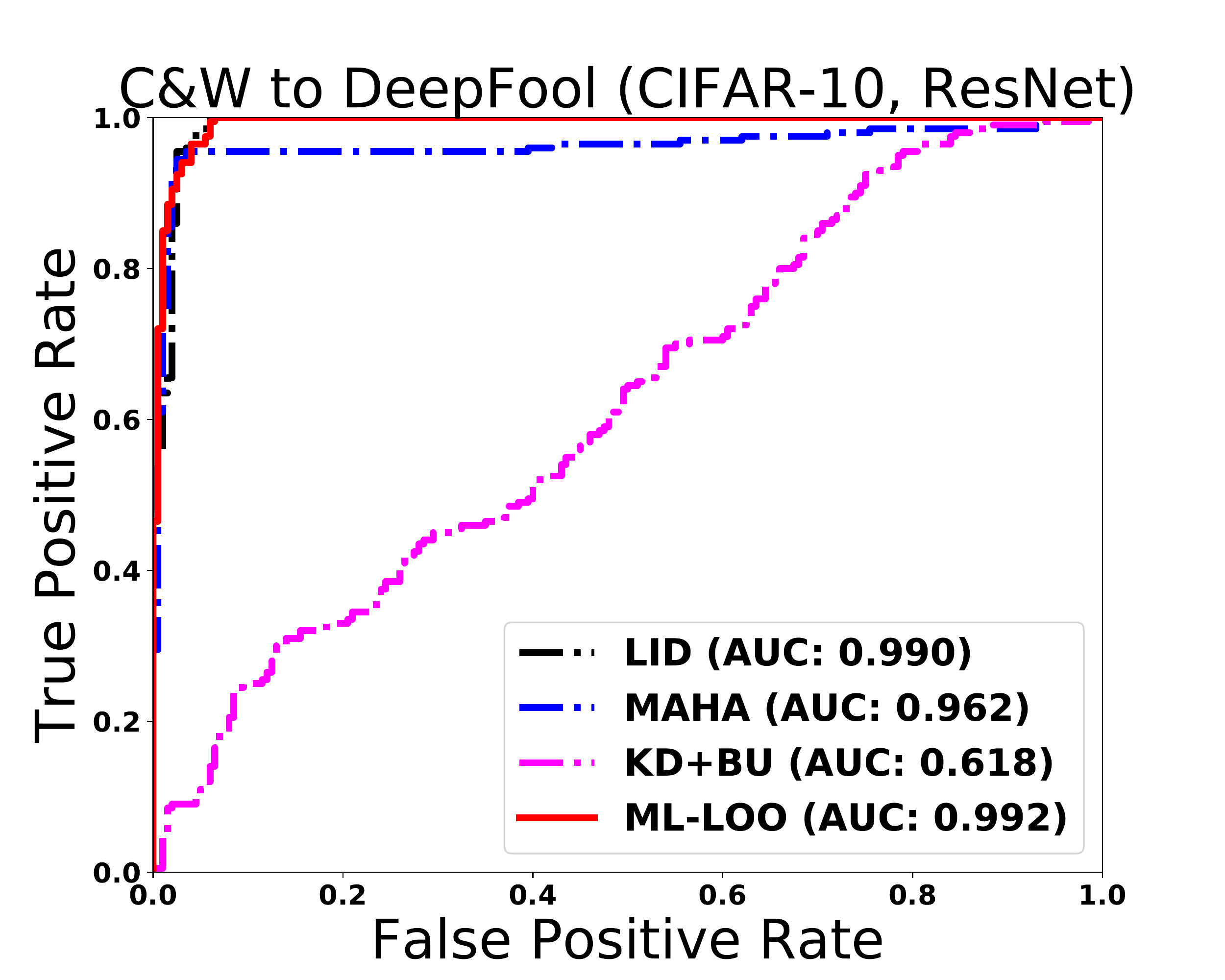}
\hspace{-3mm}
\includegraphics[width=0.2\linewidth]{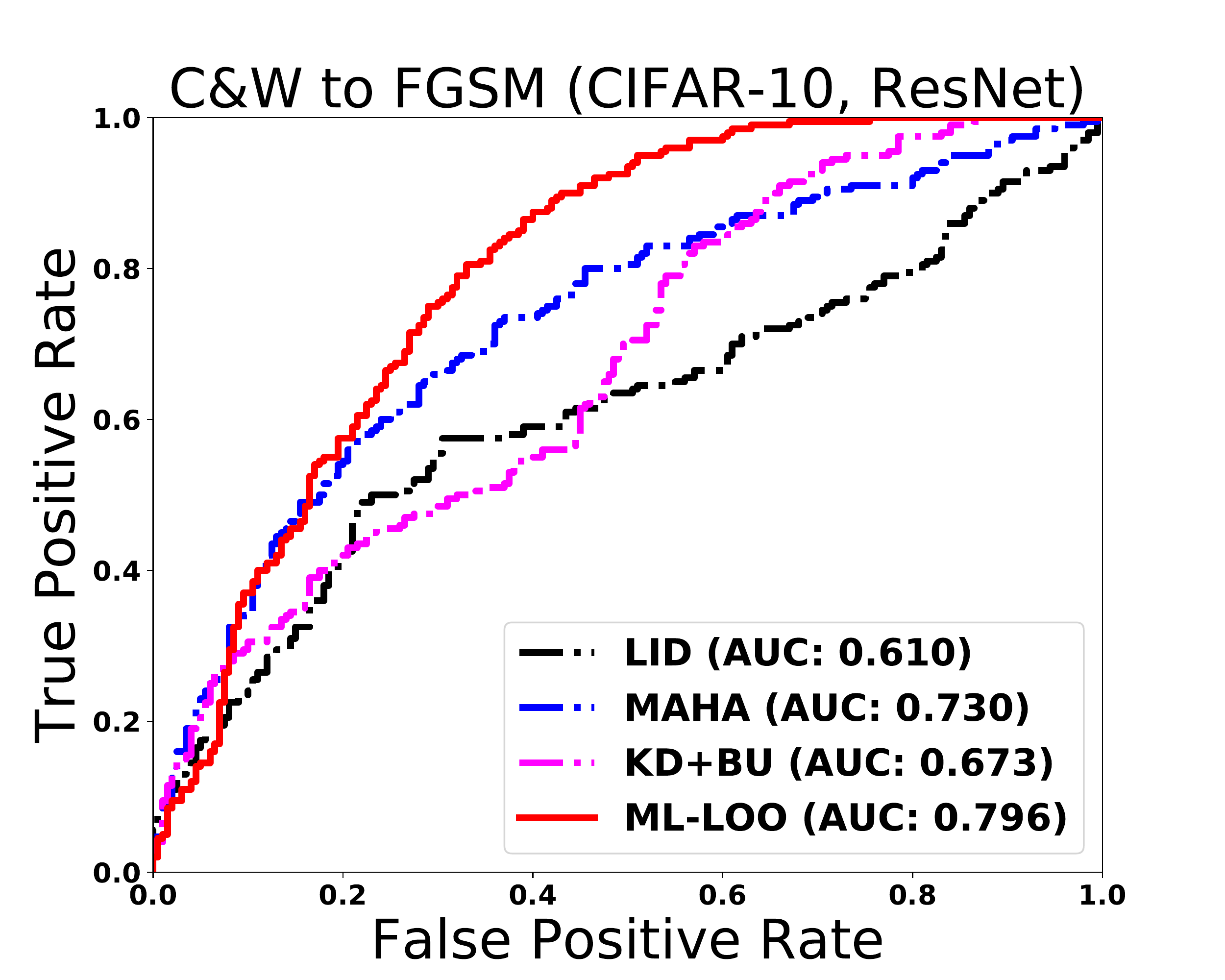} 
\hspace{-3mm}
\includegraphics[width=0.2\linewidth]{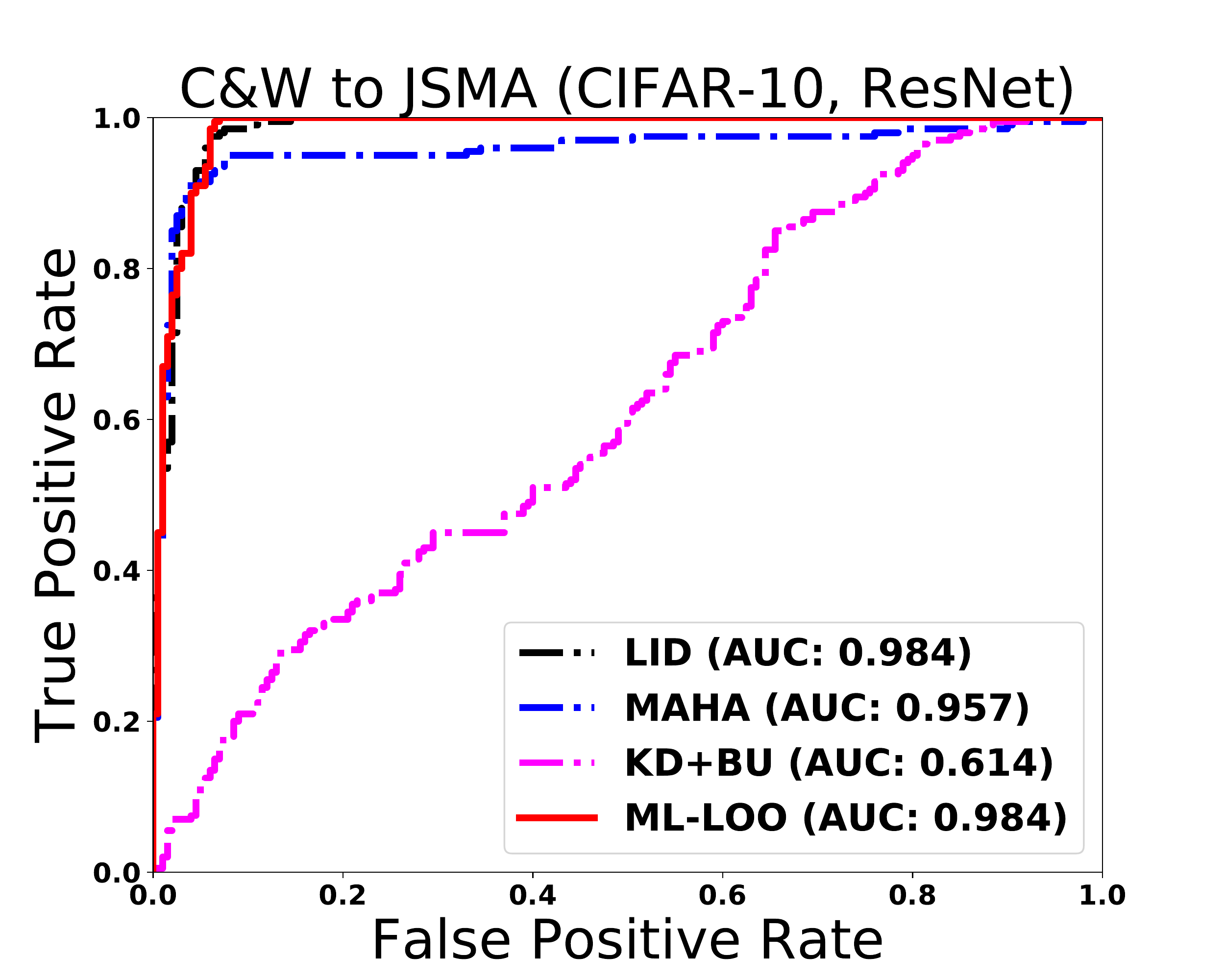}
\hspace{-3mm}
\includegraphics[width=0.2\linewidth]{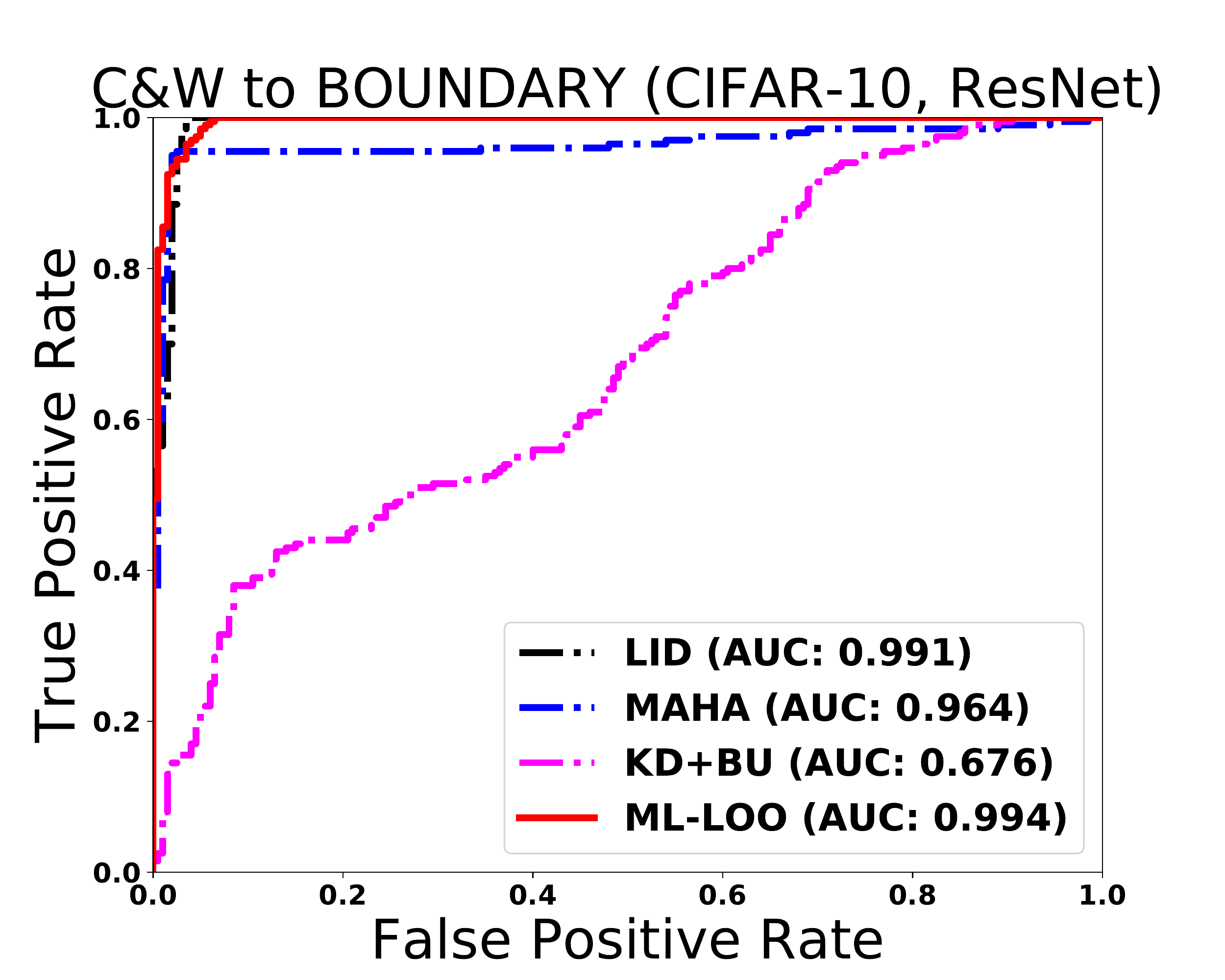}
\caption{Transferability of detection methods trained with C\&W attack and tested on $L_\infty$-PGD, FGSM, JSMA, Boundary and DeepFool.}
\label{fig:transfer_cw}
\end{figure}

\paragraph{$L_{\infty}$-PGD for optimizing $L_\infty$ distance} 
$L_{\infty}$-PGD~\cite{madry2018towards}, also named as BIM~\cite{kurakin2016adversarial}, searches for adversarial examples by iteratively updating the original image with the following:
\begin{equation}\label{eq:linf-pgd}
x_{N+1} = \text{Clip}_{x, \varepsilon}\{x_N + \alpha \text{sign}(\nabla_X J(x_N, y_{\text{true}}))\},
\end{equation}
where $y_{\text{true}}$ is the original class, $J$ is the cross-entropy loss, and Clip operator clips an image elementwise to an $\varepsilon$-neighborhood. 
For mixed-confidence $L_\infty$-PGD attack, we generated adversarial images from $L_\infty$-PGD with different confidence levels by randomly selecting the constraint $\varepsilon$ in Equation~\eqref{eq:linf-pgd} from $\{1,2,3,4,5,6,7,8\}/255$. The confidence levels of images from mixed-confidence $L_\infty$-PGD attack are shown in Figure~\ref{fig:cw}. 

We used $1,000$ adversarial images generated from the mixed-confidence $L_\infty$-PGD, together with their corresponding original images, as the training data for all detection methods. We report the results on adversarial images generated from three versions: high-confidence $L_\infty$-PGD ($\varepsilon=0.03$), low-confidence $L_\infty$-PGD ($\varepsilon=0.005$), and the mixed-confidence $L_\infty$-PGD that is used to generate the training data. The corresponding original images are different from the training images. Table~\ref{table:auc_table_pgd_mix} (Bottom) and Figure~\ref{fig:cw} (Right) show TPRs at different FPR thresholds, AUC, and the ROC curve. Mahalanobis, LID and KD+BU fail to detect adversarial examples of mixed-confidence effectively, while our method performs significantly better across the three settings.

\subsection{Transferability}

In this experiment, we evaluate the transferability of different methods by training detection methods on adversarial examples generated from one attacking method and carry out the evaluation on adversarial examples generated from different attacking methods. We trained all methods on adversarial examples generated by C\&W attack 
and carried out the evaluation on adversarial examples generated by the rest of the attacking methods. 

Experiments are carried out on MNIST, CIFAR-10, and CIFAR-100 datasets. AUC and TPRs at different FPR thresholds are reported in Table~\ref{table:auc_table_transfer}. All methods trained on C\&W attack are capable of detecting adversarial examples generated from an unknown attack, even when the optimized distance is $L_\infty$, or the attack is not gradient-based. The same phenomenon has been observed in \citet{lee2018simple} as well. This indicates attacks might share some common features. Our method yields a slightly higher AUC consistently, and has a significantly higher TPR when FPRs are controlled to be small.
\section{Discussion}
In this paper, we have introduced a new framework to detect adversarial examples with multi-layer feature attribution, by capturing the scaling difference of feature attribution scores between original and adversarial examples. We show that our detection method outperforms other state-of-the-art detection methods in detecting various kinds of attacks. In particular, we show our method is able to detect adversarial examples of various confidence levels, and transfers between different attacks.

\bibliographystyle{unsrtnat}

{
\bibliography{detect_arxiv}
}
\newpage

\section{appendix}

\subsection{Performance of Integrated Gradients}\label{app:ig}

In this section, we evaluate the detection of adversarial examples by thresholding the IQR of another popular feature attribution method Integrated Gradients (IG), and compare it with KD+BU, LID, MAHA, and ML-LOO. We consider three attacks FGSM, C\&W and JSMA, which are optimized for $L_\infty,L_2$ and $L_0$ distances respectively, on CIFAR-10 dataset with ResNet. We can see that IQR of IG achieves competitive performance in detecting adversarial examples, but not as powerful as the detection methods which incorporated multi-layer information like LID, MAHA and our proposed method ML-LOO. The IG feature is also not as effective as the LOO feature (whose performance is shown in Figure~\ref{fig:hist_stat2}). 

\label{sec:ig_plots}
\begin{figure}[H]
\centering 
\includegraphics[width=0.3\linewidth]{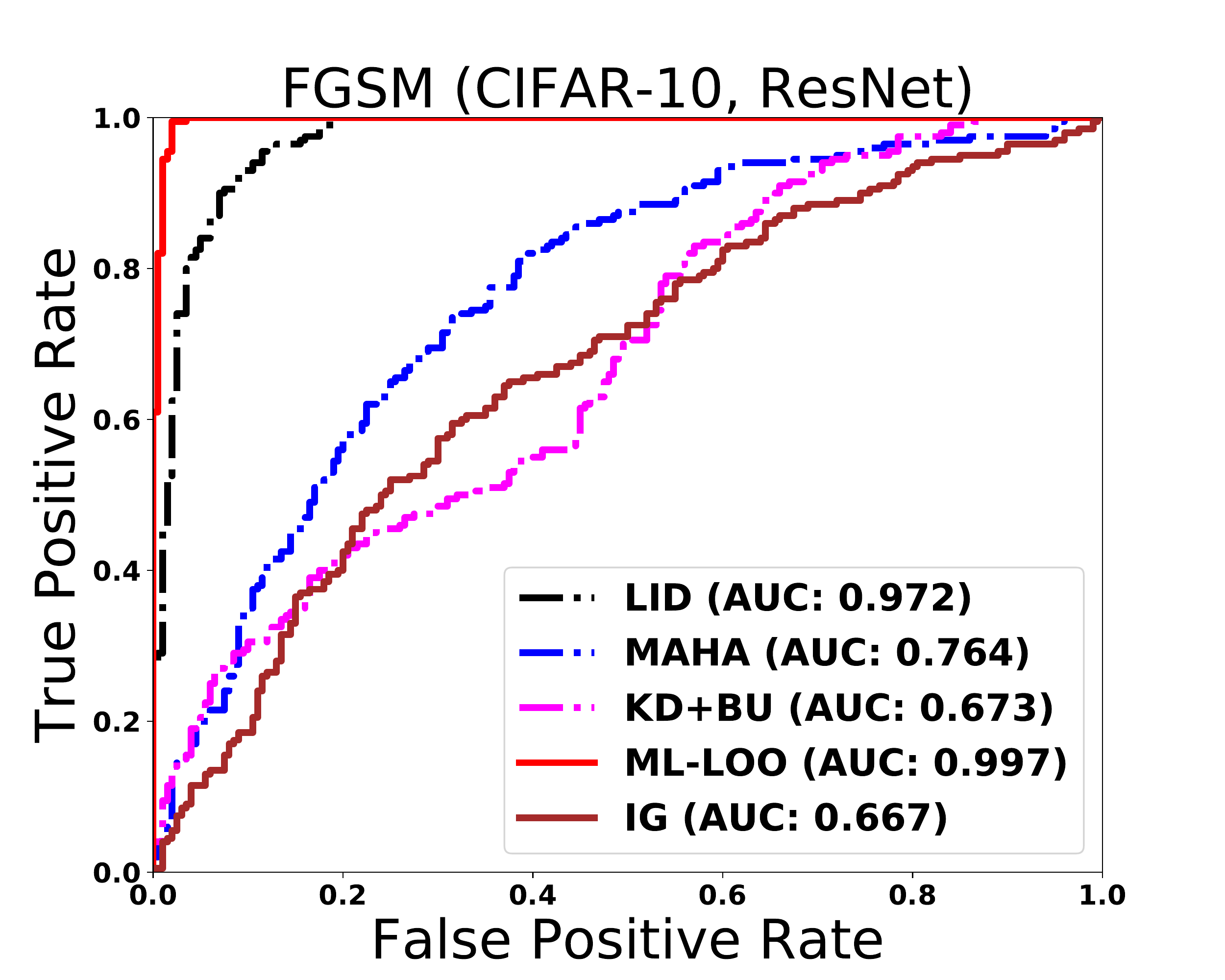} 
\includegraphics[width=0.3\linewidth]{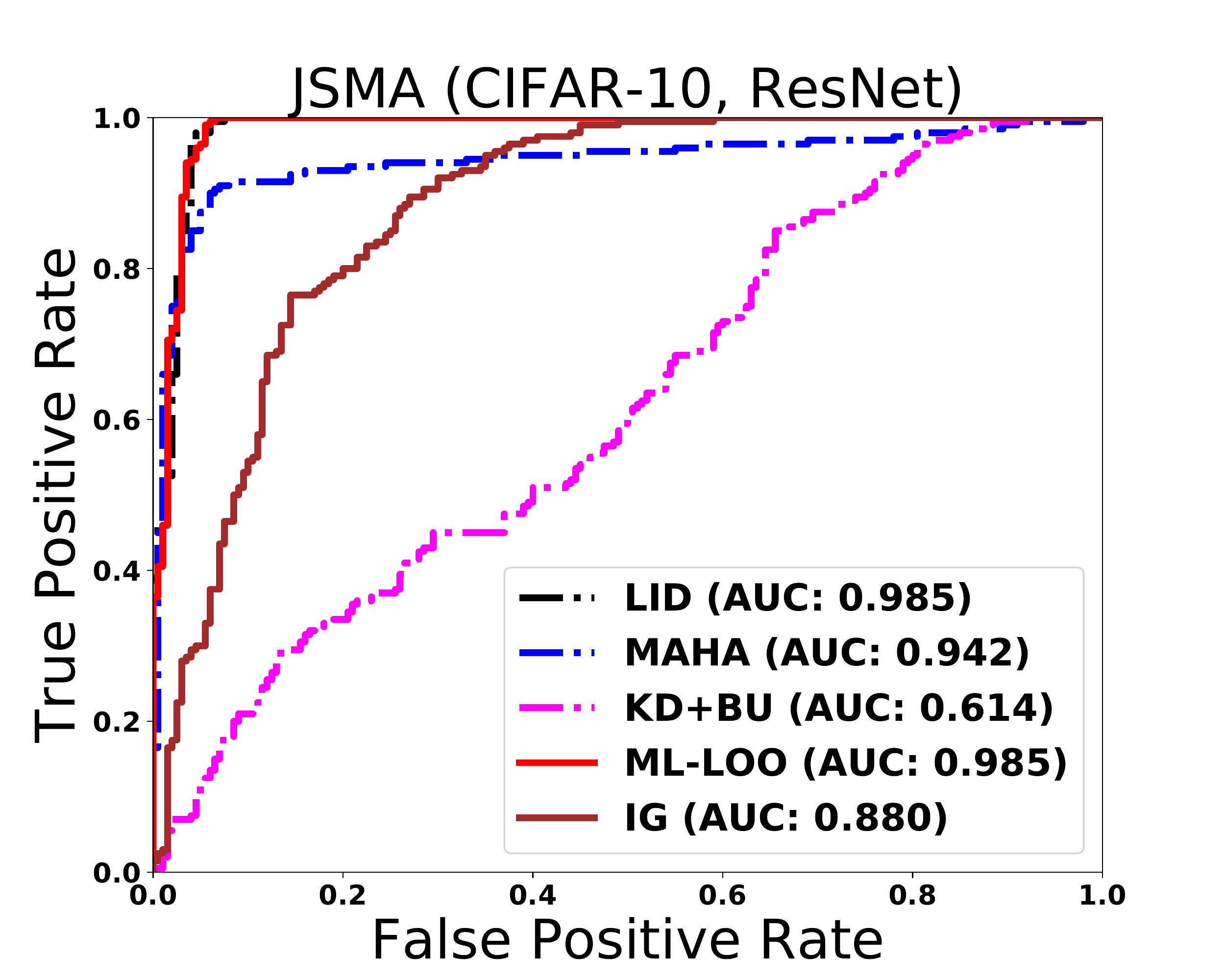} 
\includegraphics[width=0.3\linewidth]{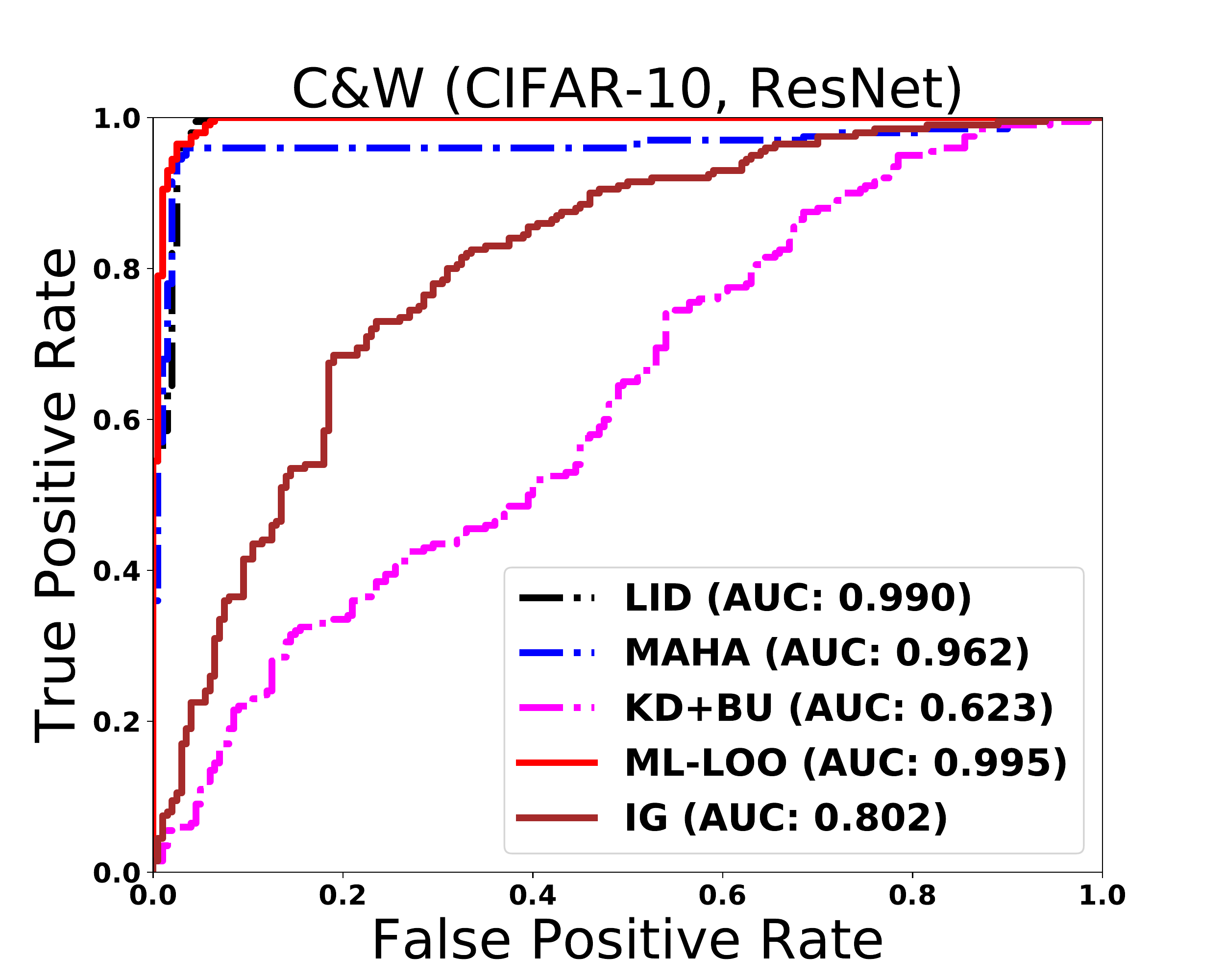}
\caption{ROC curves of detection methods on CIFAR-10 dataset with ResNet. We restrict FPR between 0 and 0.2, which is meaningful in practice. See Appendix~\ref{sec:roc_plots} for full plots.}
\label{fig:CIAFR10RESNETIG}
\end{figure} 

\newpage
\subsection{Comparison Based on Dispersion Measures}

\label{sec:roc_stat}

In this section, we compare performance of detection using three different dispersion measures of feature attributions: IQR, STD and MAD.

\begin{figure}[H]
\centering 
\includegraphics[width=0.3\linewidth]{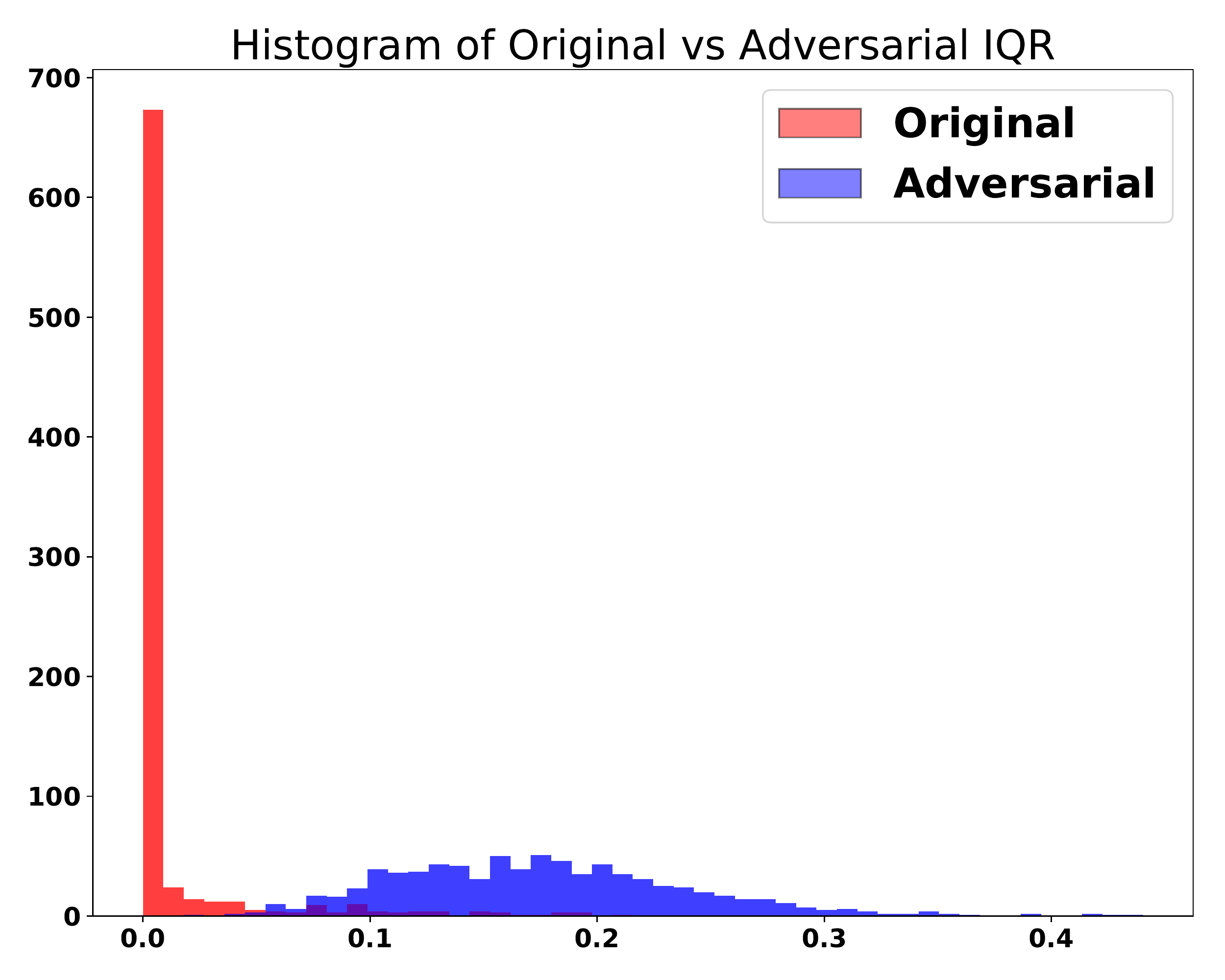} 
\includegraphics[width=0.3\linewidth]{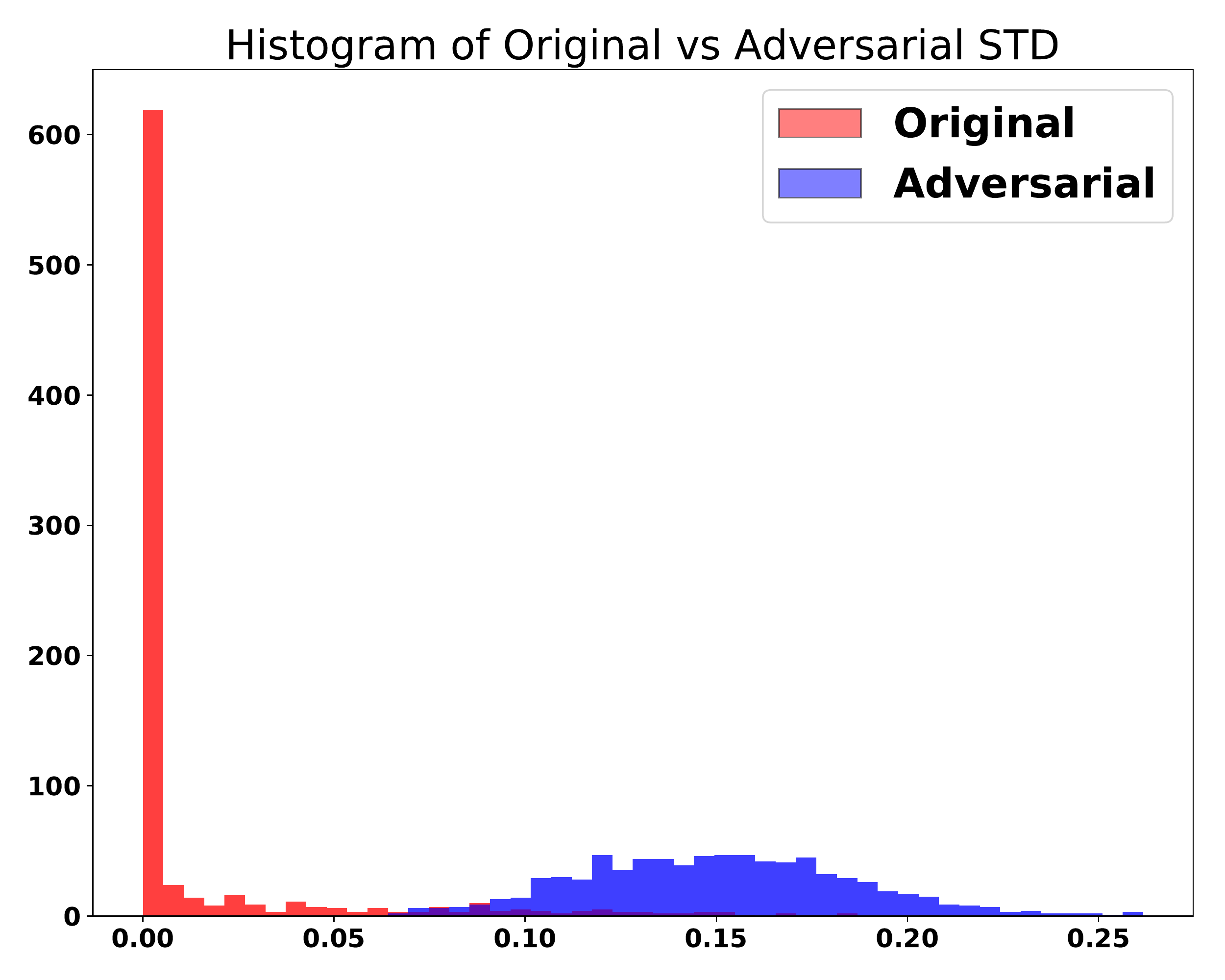} 
\includegraphics[width=0.3\linewidth]{cifar10resnet/MAD_Histogram.pdf} 
\caption{Histogram of Statistics}
\label{fig:hist_stat}
\end{figure}

Figure~\ref{fig:hist_stat} shows the histograms of these three dispersion measures of feature attributions for ResNet on natural test images from CIFAR-10 with those on adversarially perturbed images, where the adversarial perturbation is carried out by C\&W Attack. We can see there is a significant difference in the distributions of the dispersion measures between natural and adversarial images.

\begin{figure}[H]
\centering 
\includegraphics[width=0.3\linewidth]{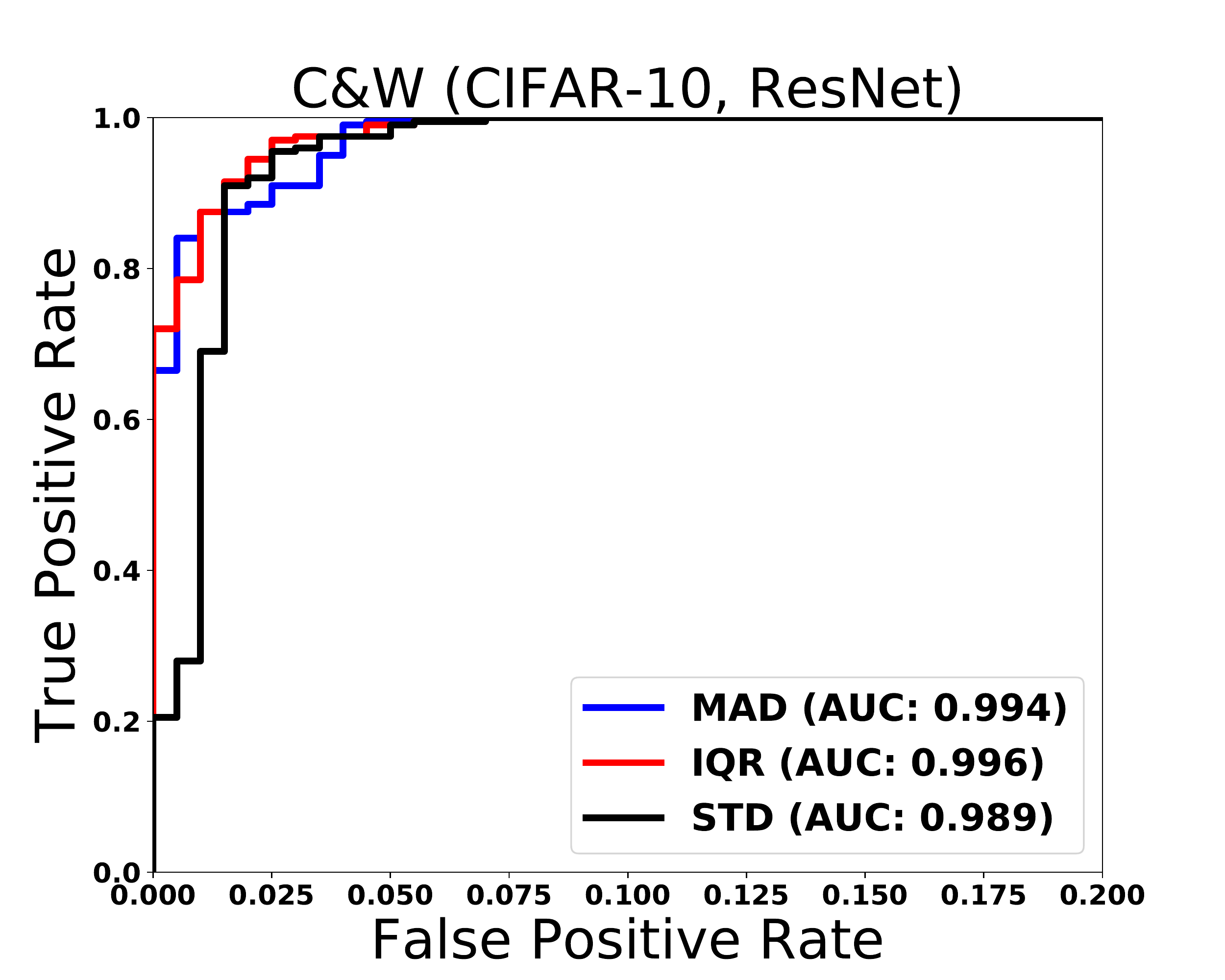} 
\includegraphics[width=0.3\linewidth]{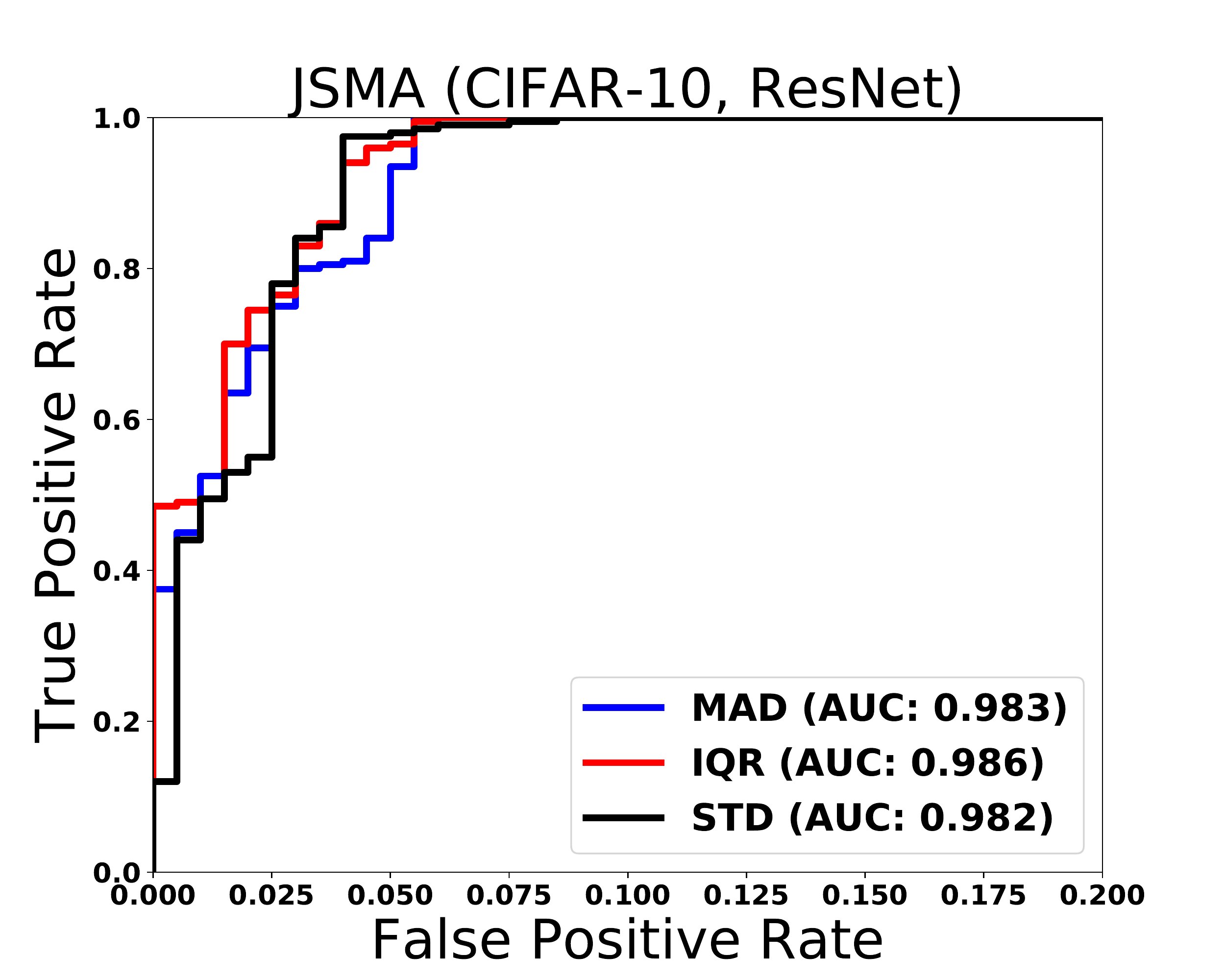}
\includegraphics[width=0.3\linewidth]{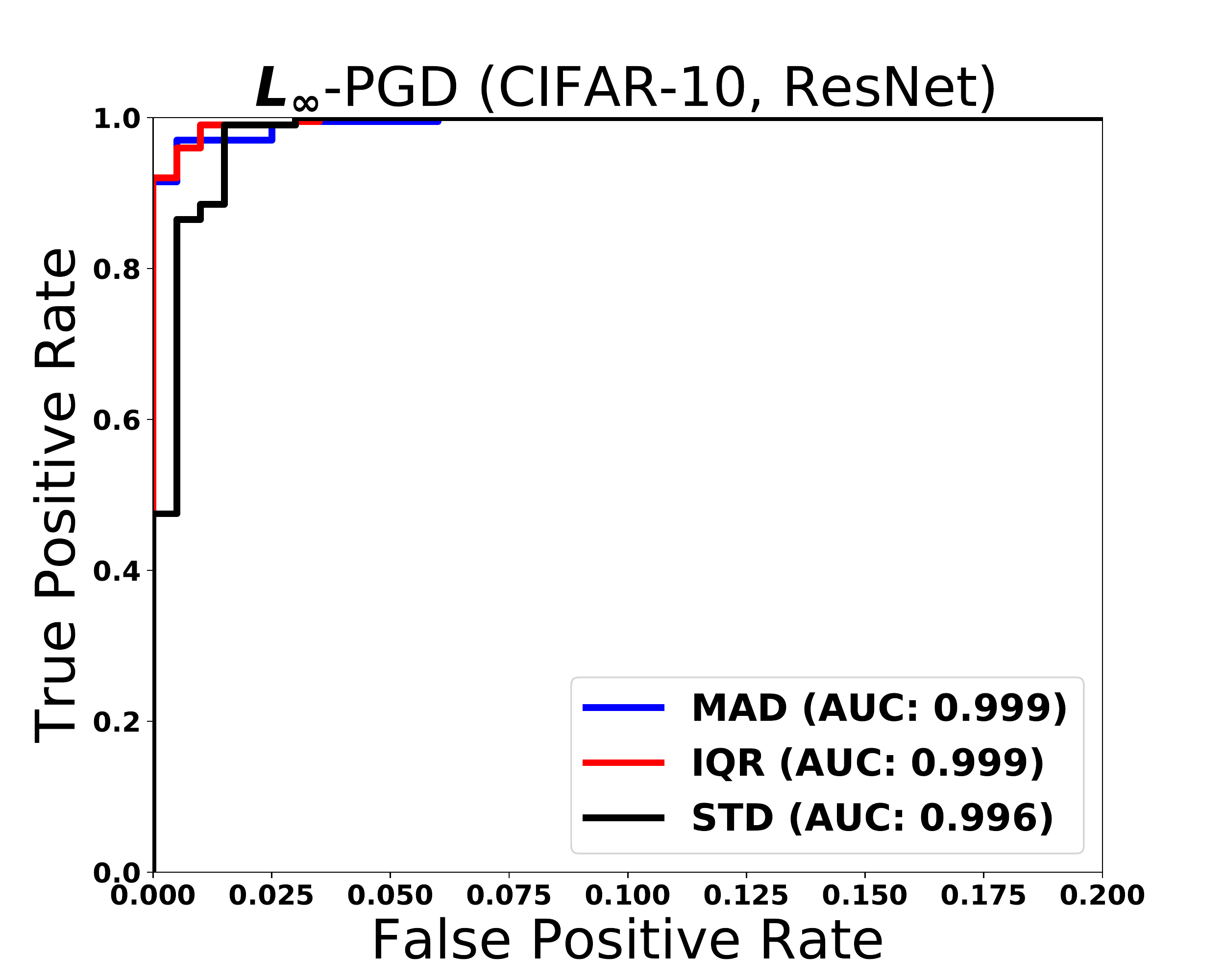} 
\caption{ROC curves of different statistics on CIFAR-10 dataset with ResNet}
\label{fig:hist_stat2}
\end{figure}

Figure~\ref{fig:hist_stat2} shows the ROC curves of the three dispersion statistics on CIFAR-10 with ResNet. We can see that all three dispersion measures achieve competitive performance in detecting adversarial examples generated by three attacks C\&W, JSMA and $L_\infty$-PGD, but IQR achieves the largest AUC values across all attacking methods.

\newpage
\subsection{ROC curves of detection methods on CIFAR-10, MNIST and CIFAR-100 data sets with FPR from 0.0 to 0.2}
\label{sec:roc_plots2}

\begin{figure}[H]
\centering 
\includegraphics[width=0.3\linewidth]{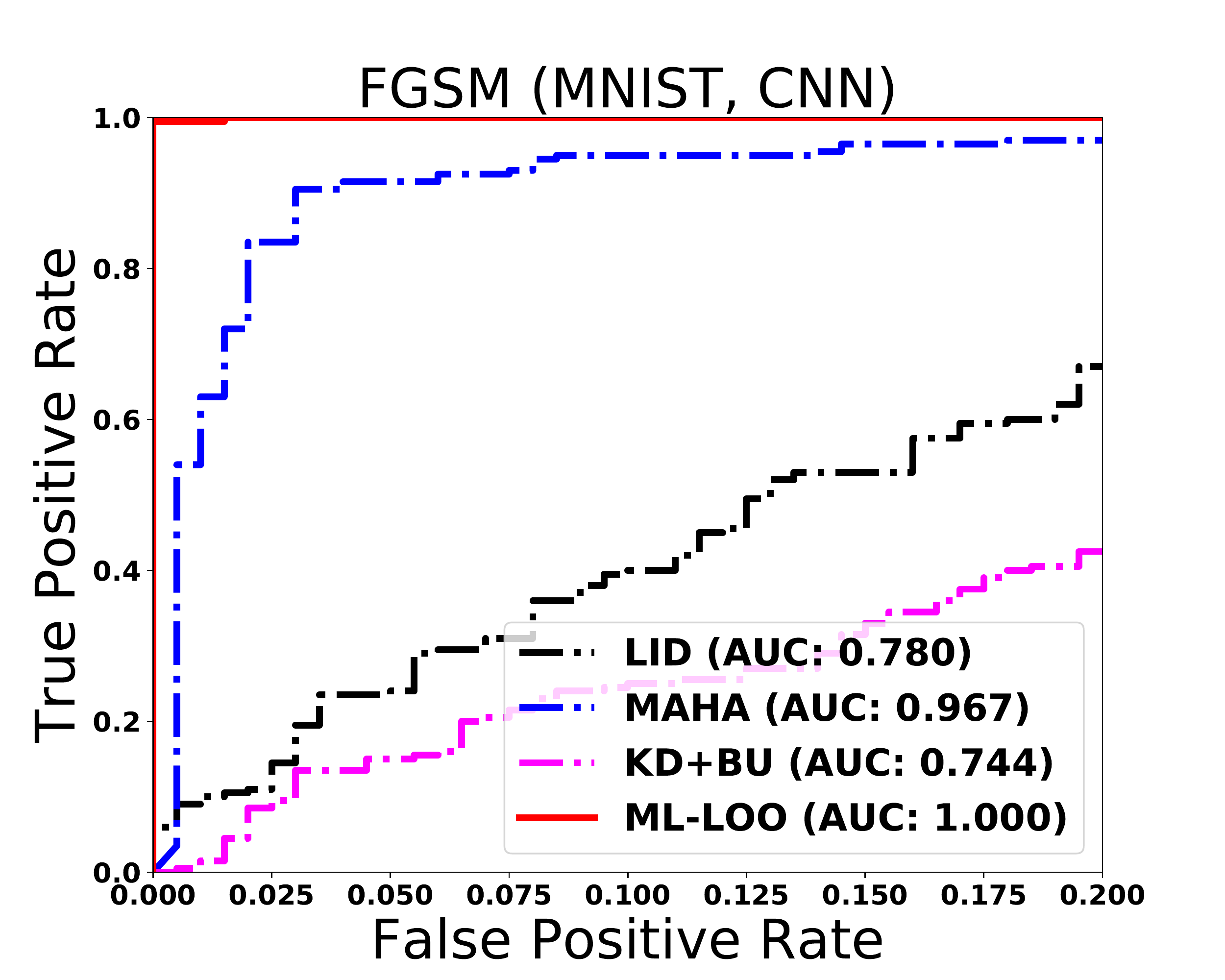} 
\includegraphics[width=0.3\linewidth]{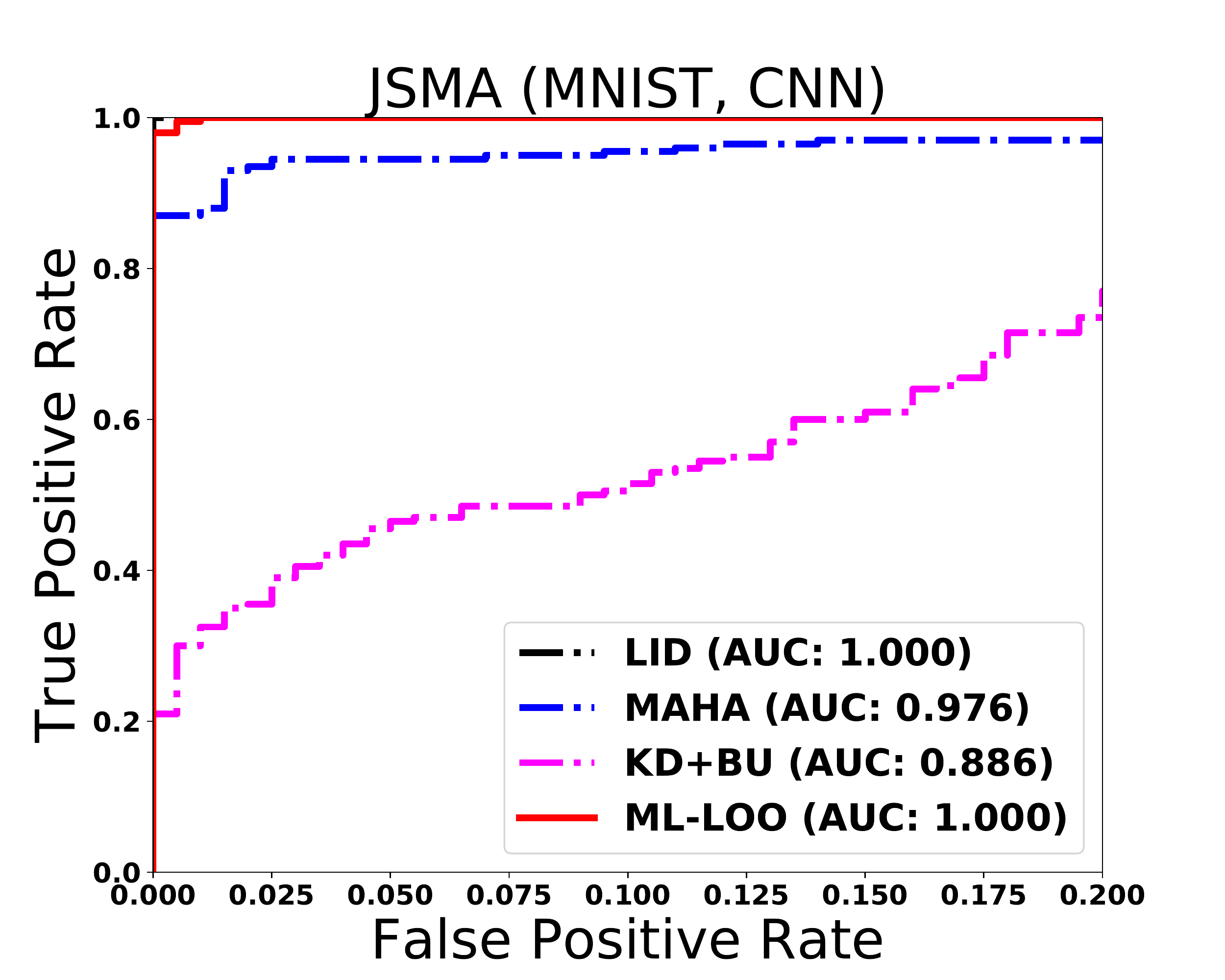} 
\includegraphics[width=0.3\linewidth]{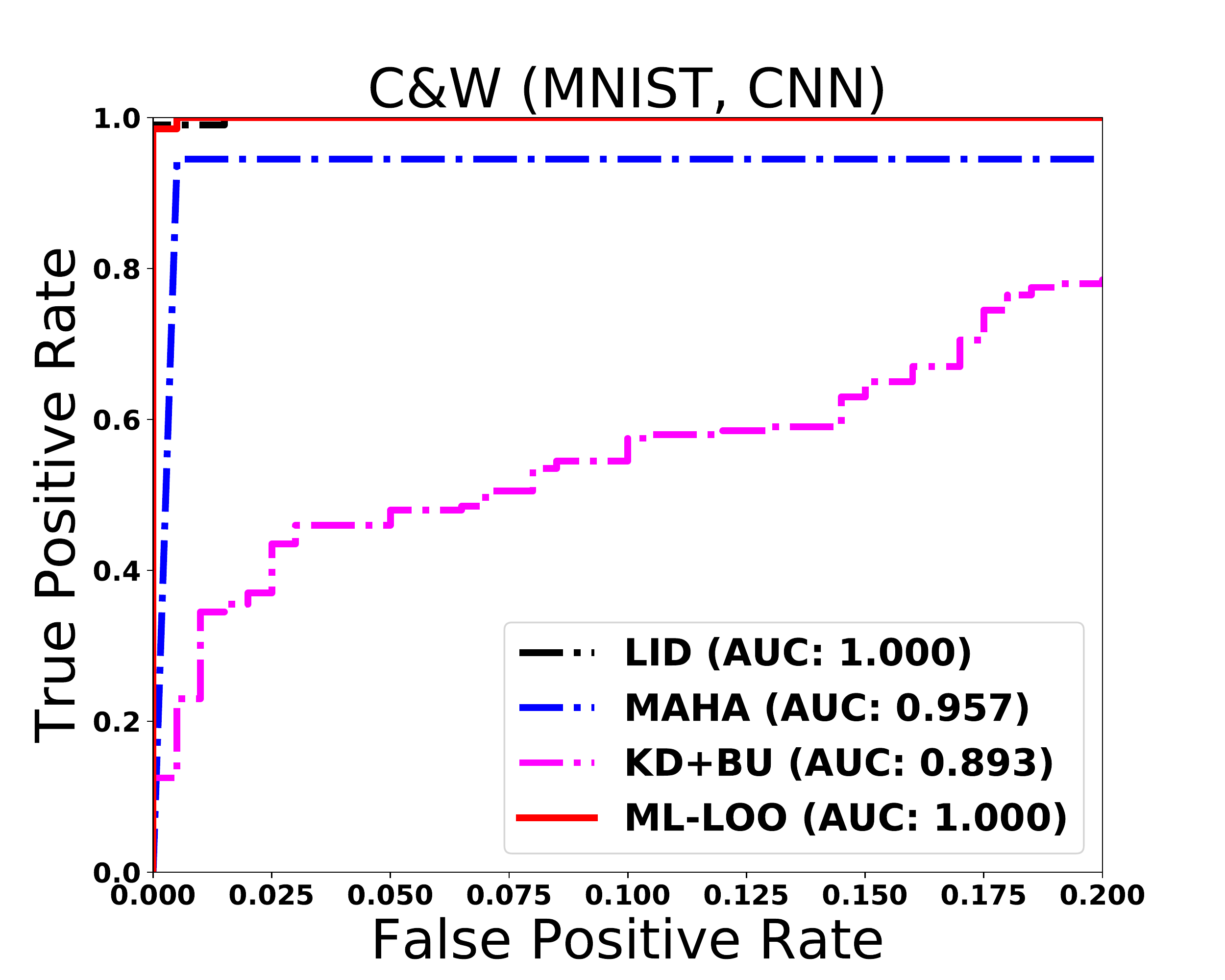} 
\includegraphics[width=0.3\linewidth]{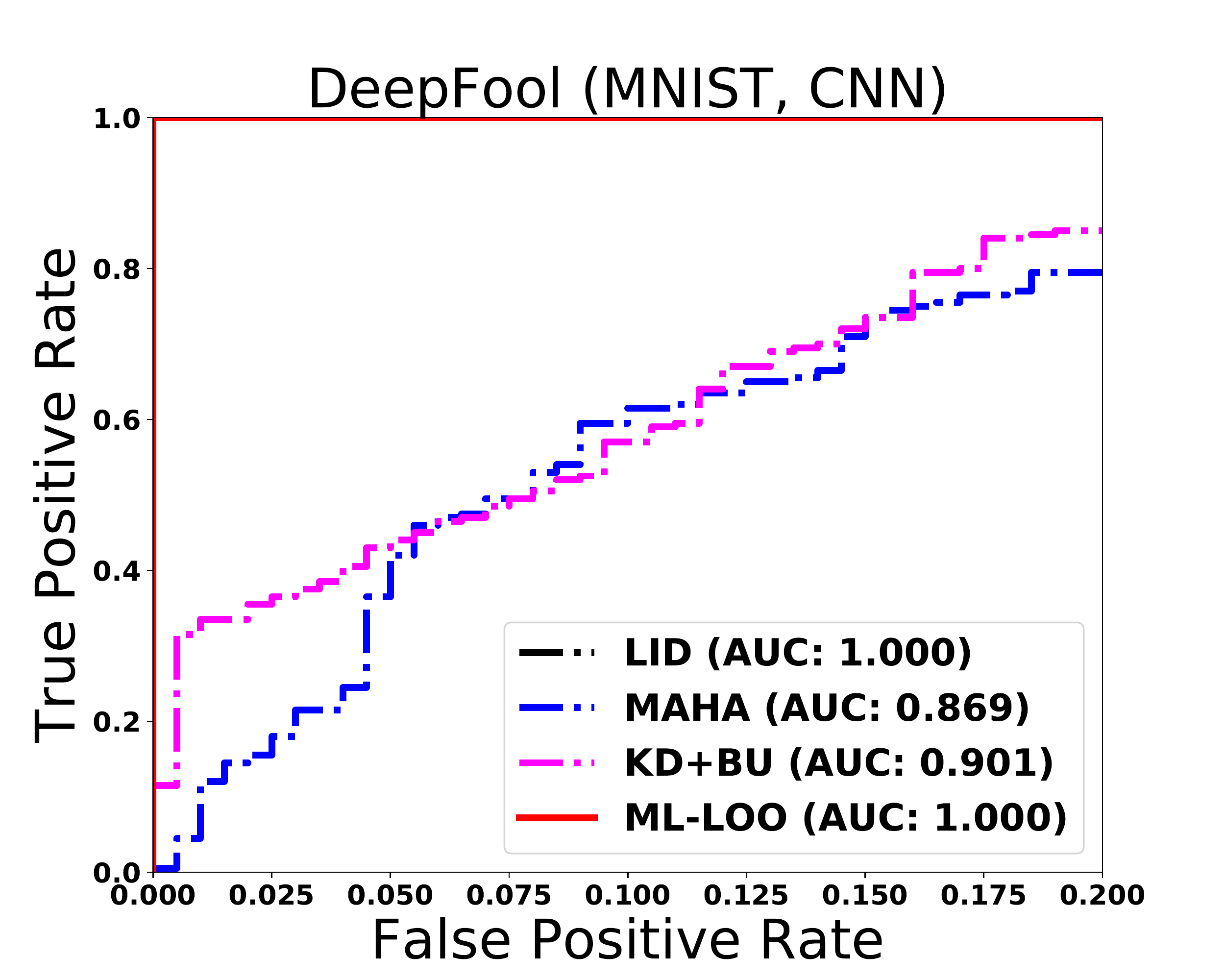} 
\includegraphics[width=0.3\linewidth]{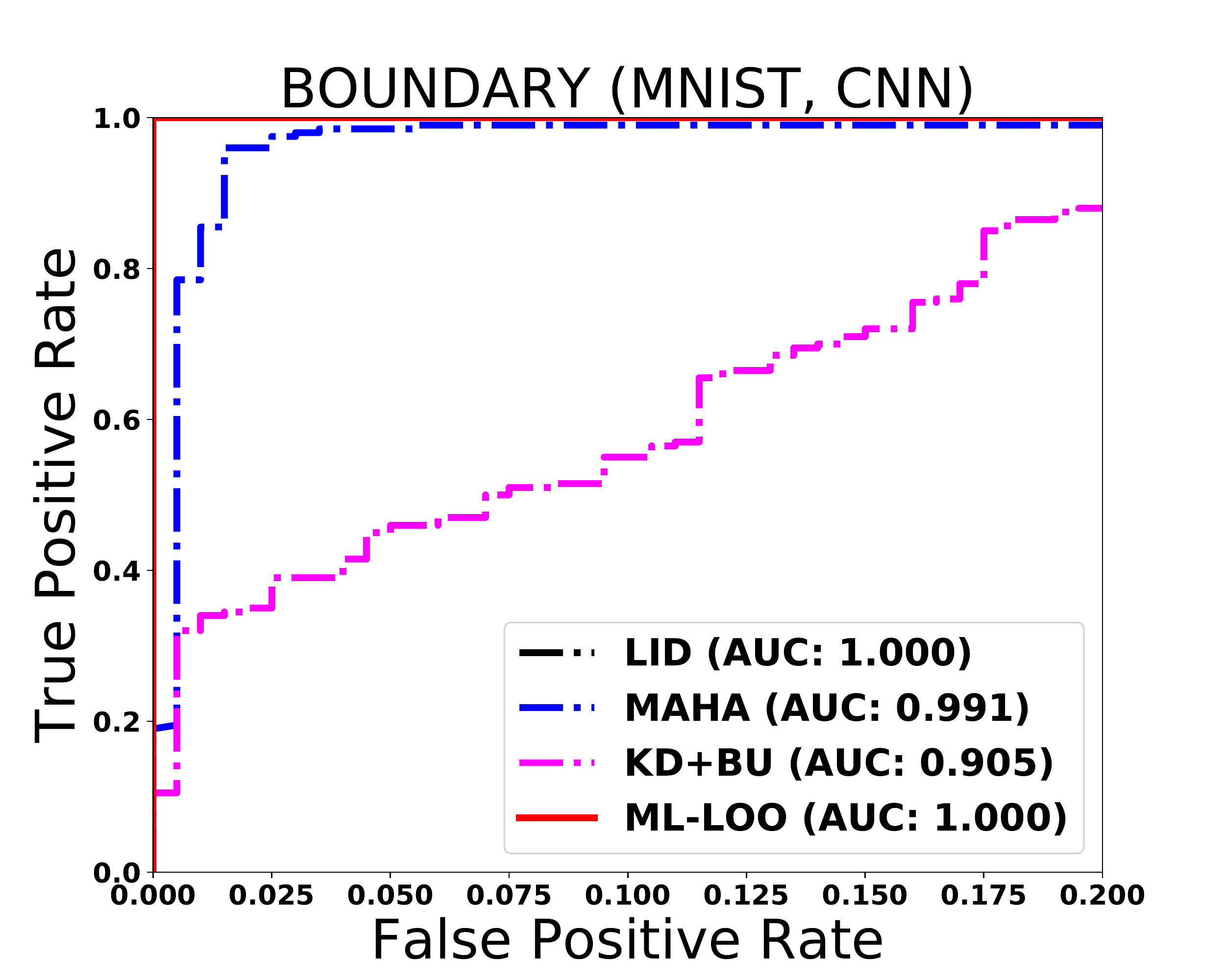} 
\includegraphics[width=0.3\linewidth]{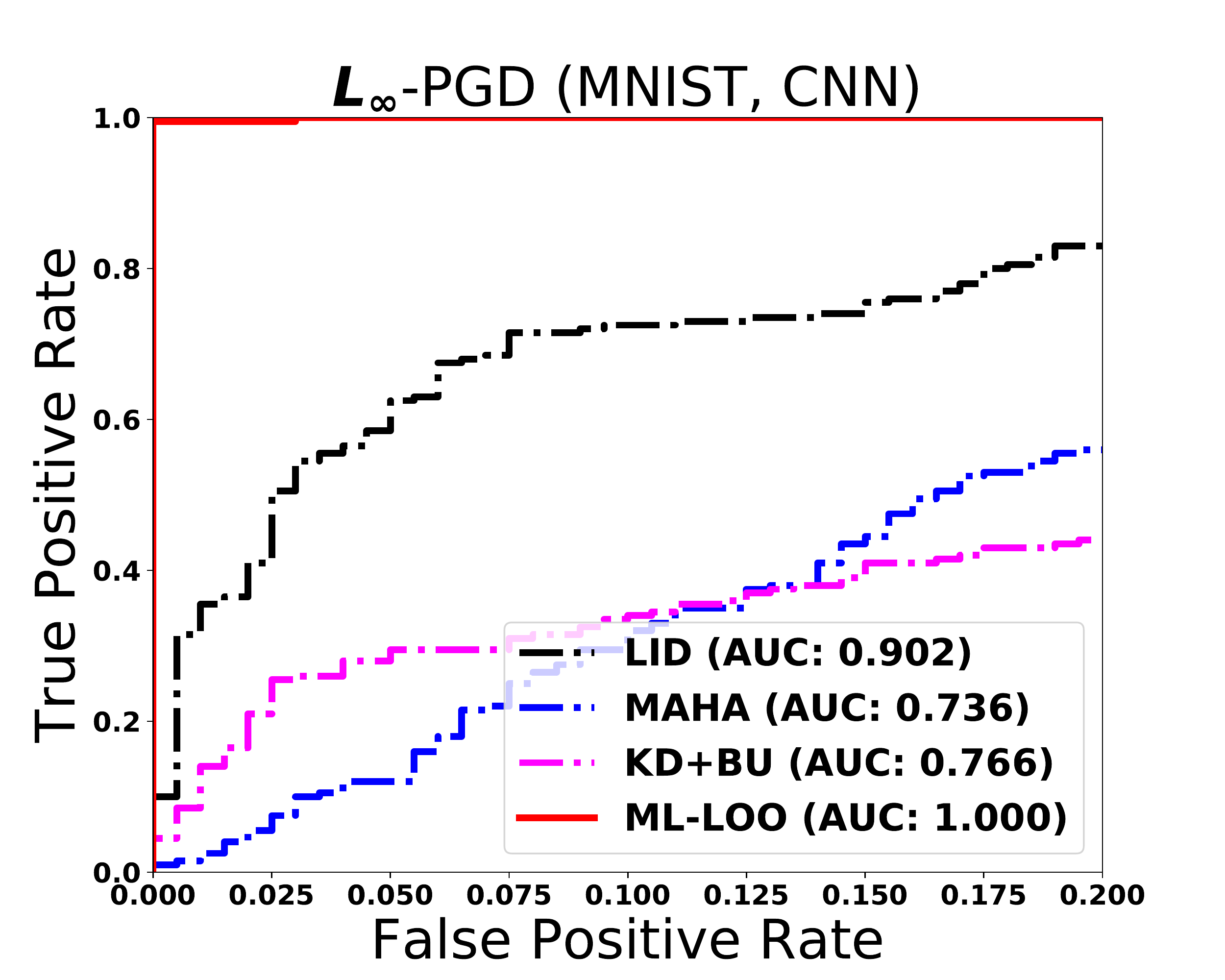}
\caption{ROC curves of detection methods on MNIST dataset with CNN}
\label{fig:MNISTSCNN}
\end{figure}

\begin{figure}[H]
\centering 
\includegraphics[width=0.3\linewidth]{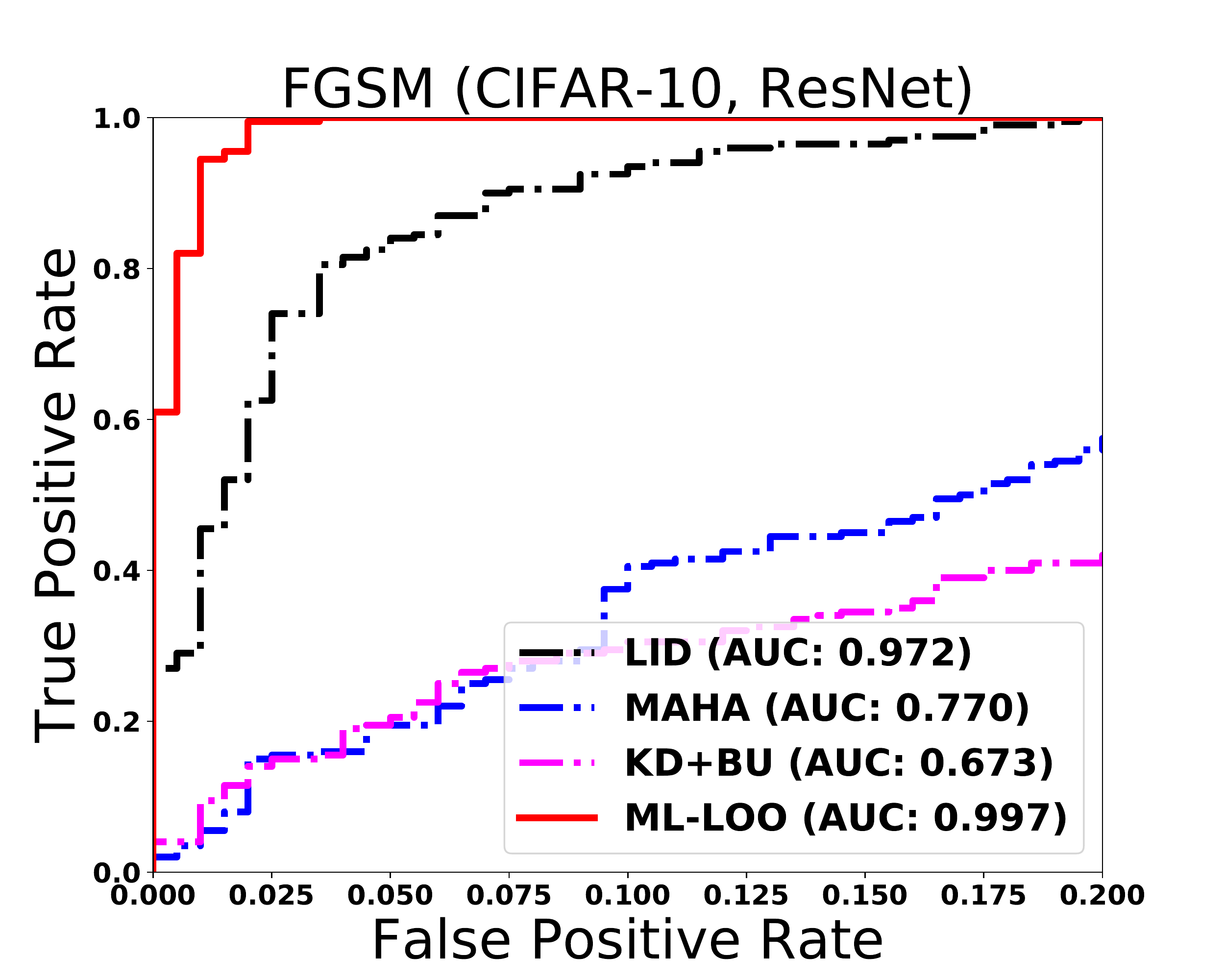} 
\includegraphics[width=0.3\linewidth]{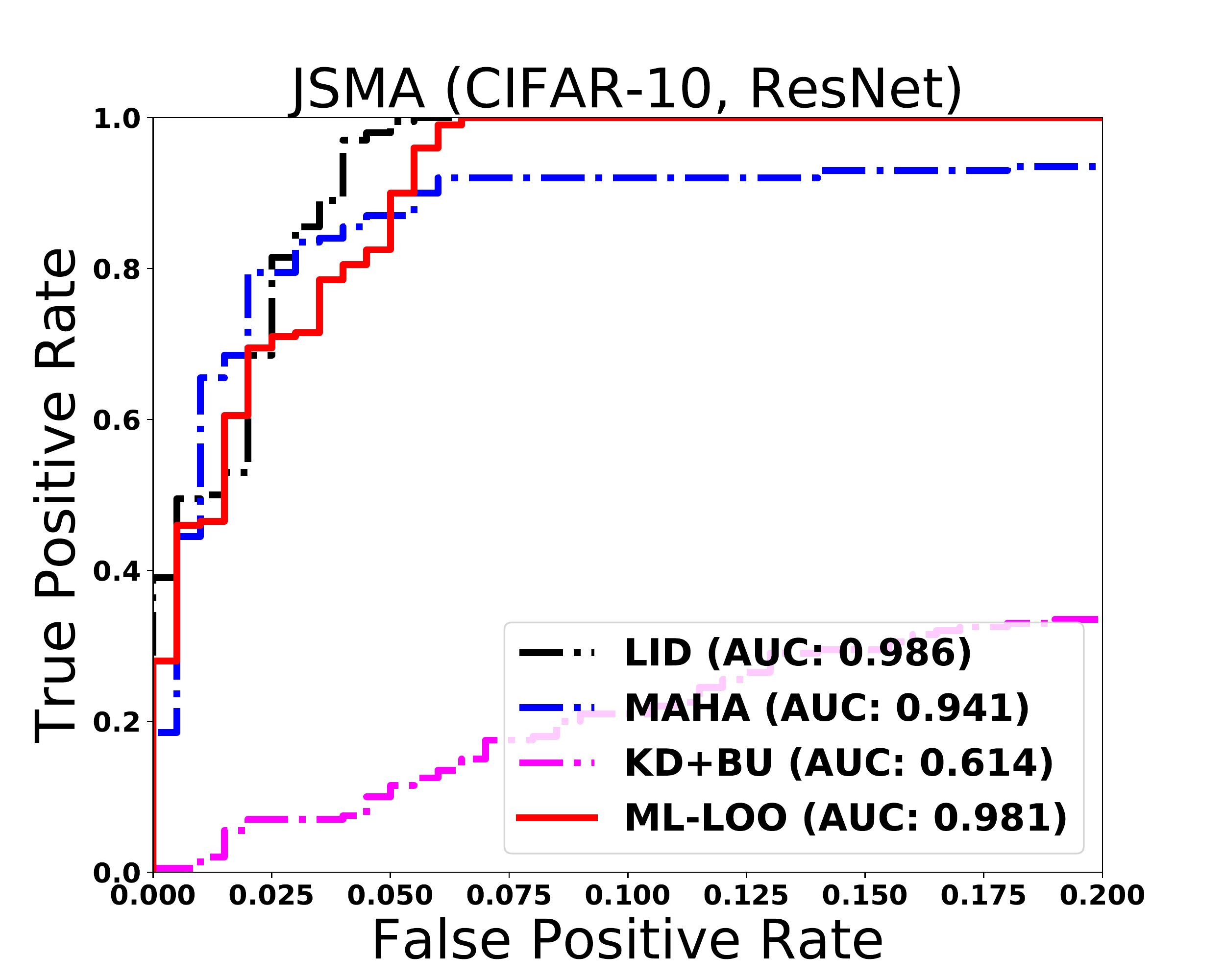} 
\includegraphics[width=0.3\linewidth]{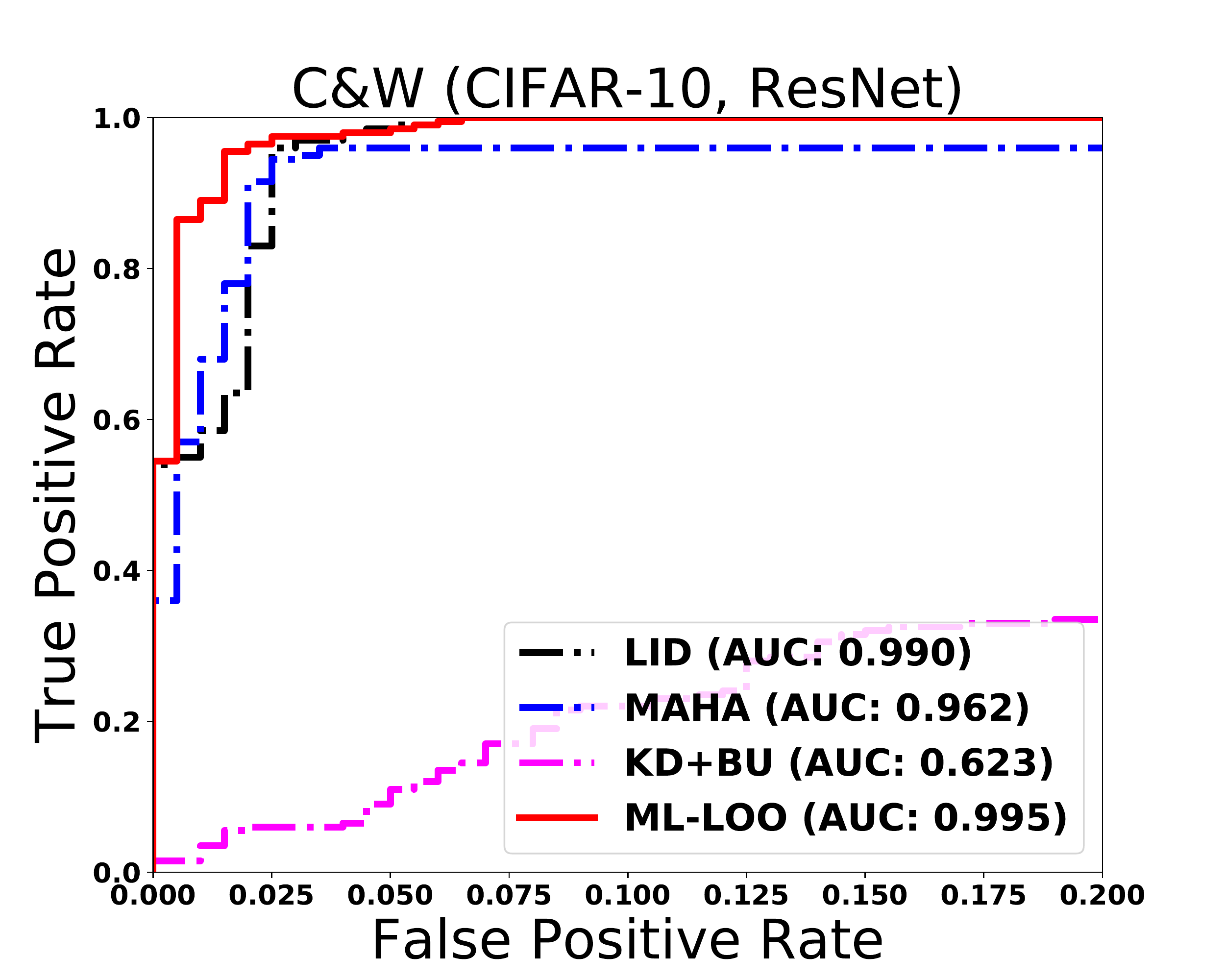} 
\includegraphics[width=0.3\linewidth]{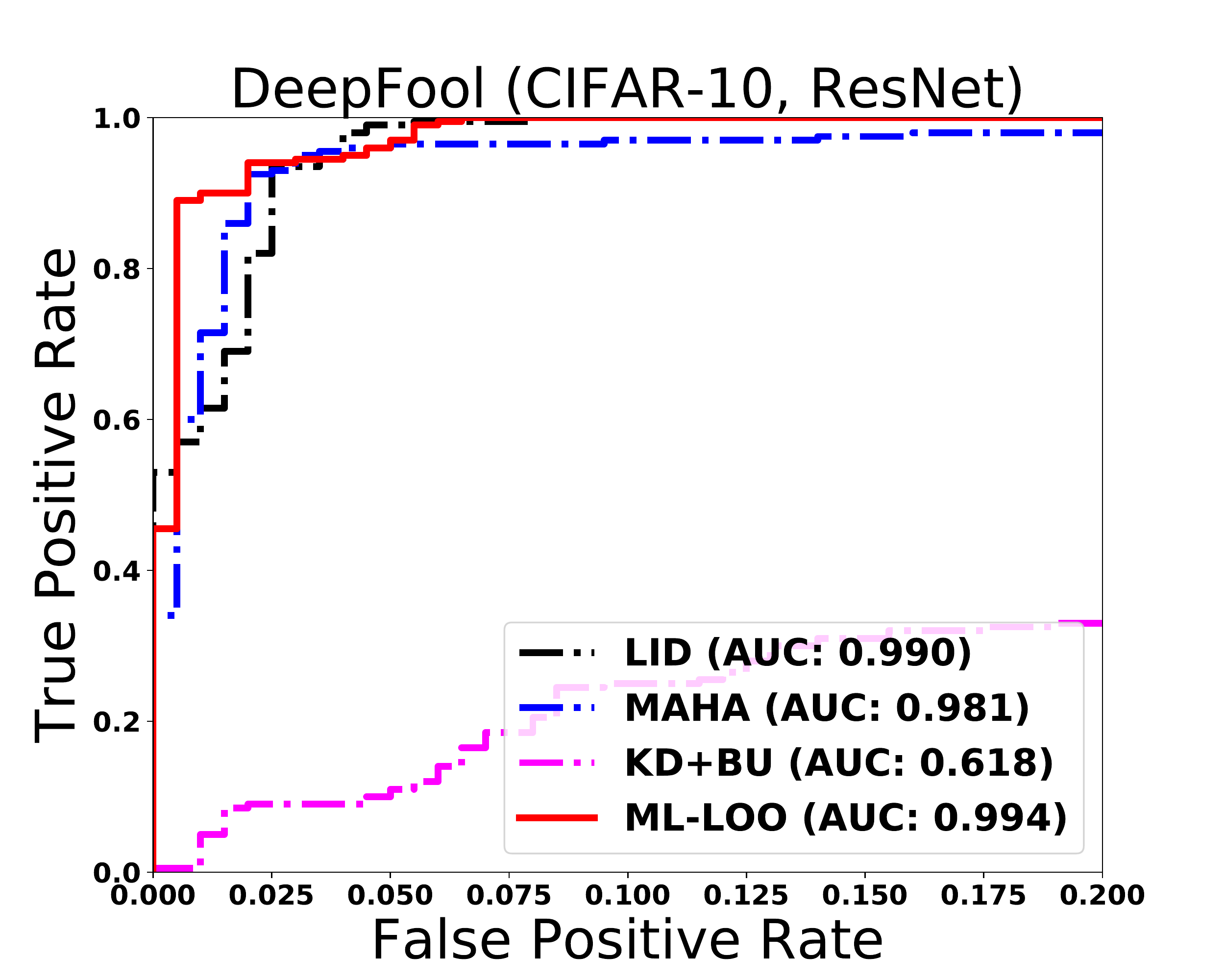} 
\includegraphics[width=0.3\linewidth]{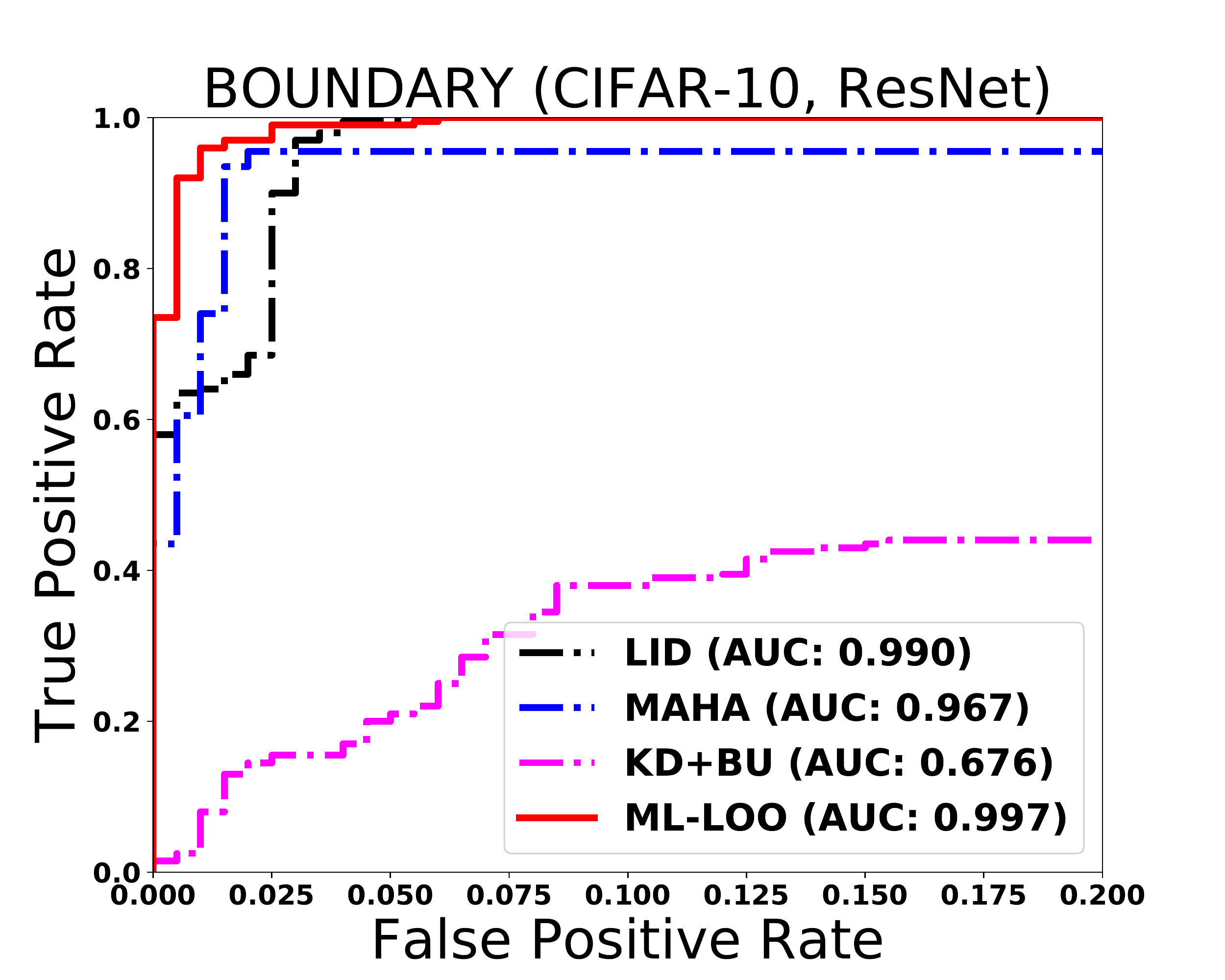} 
\includegraphics[width=0.3\linewidth]{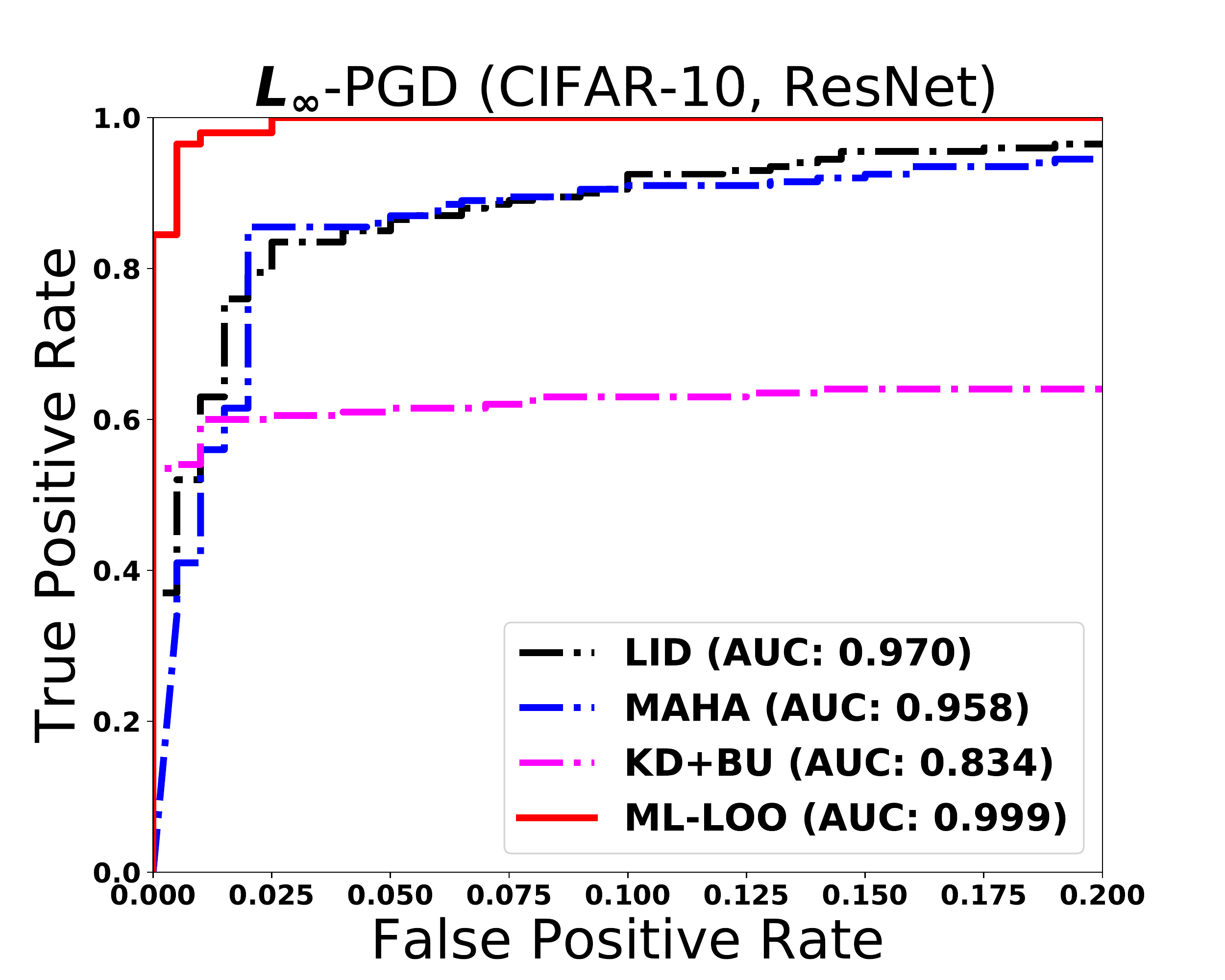}
\caption{ROC curves of detection methods on CIFAR-10 dataset with ResNet}
\label{fig:CIFAR10RESNET2}
\end{figure}

\begin{figure}[H]
\centering 
\includegraphics[width=0.3\linewidth]{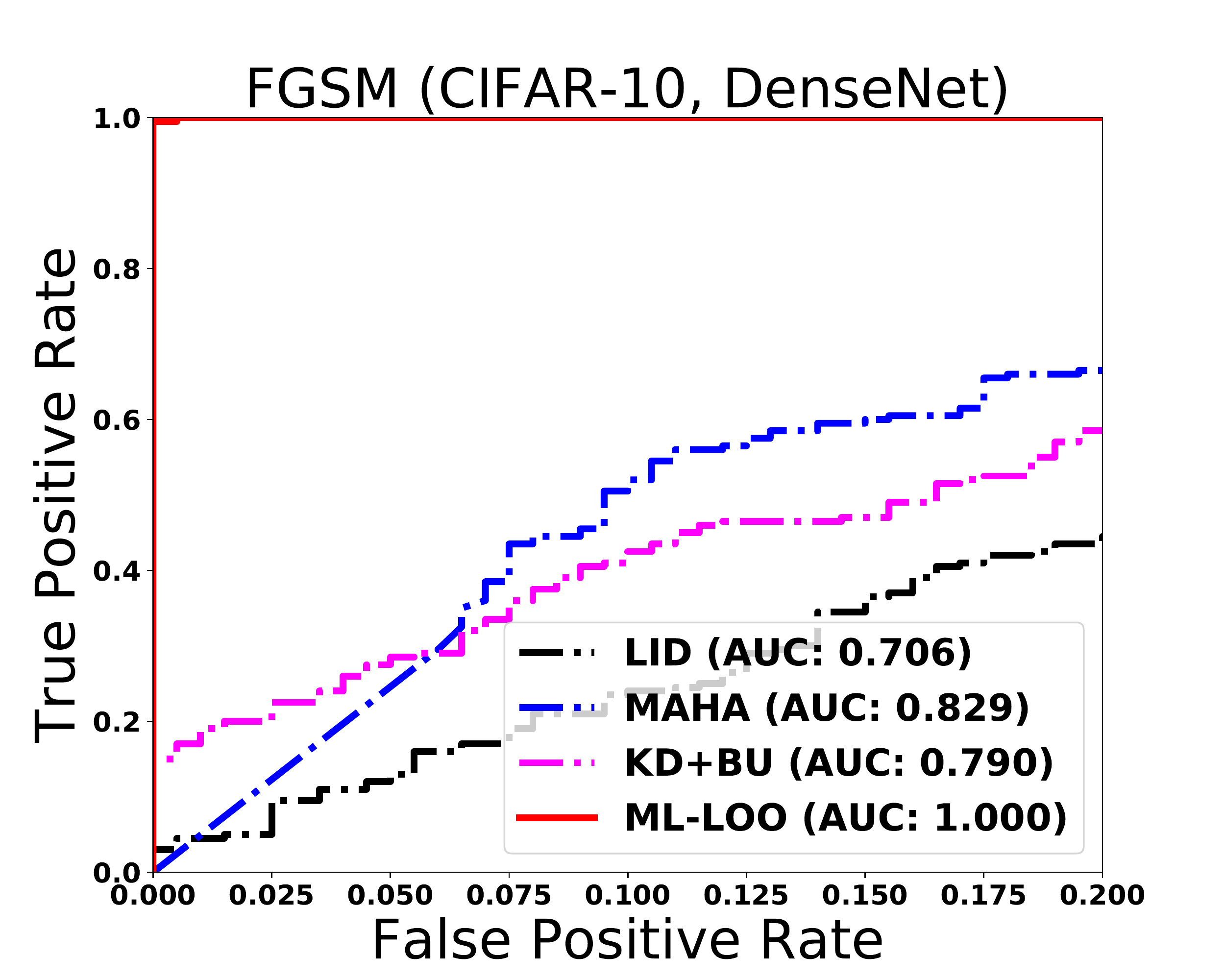} 
\includegraphics[width=0.3\linewidth]{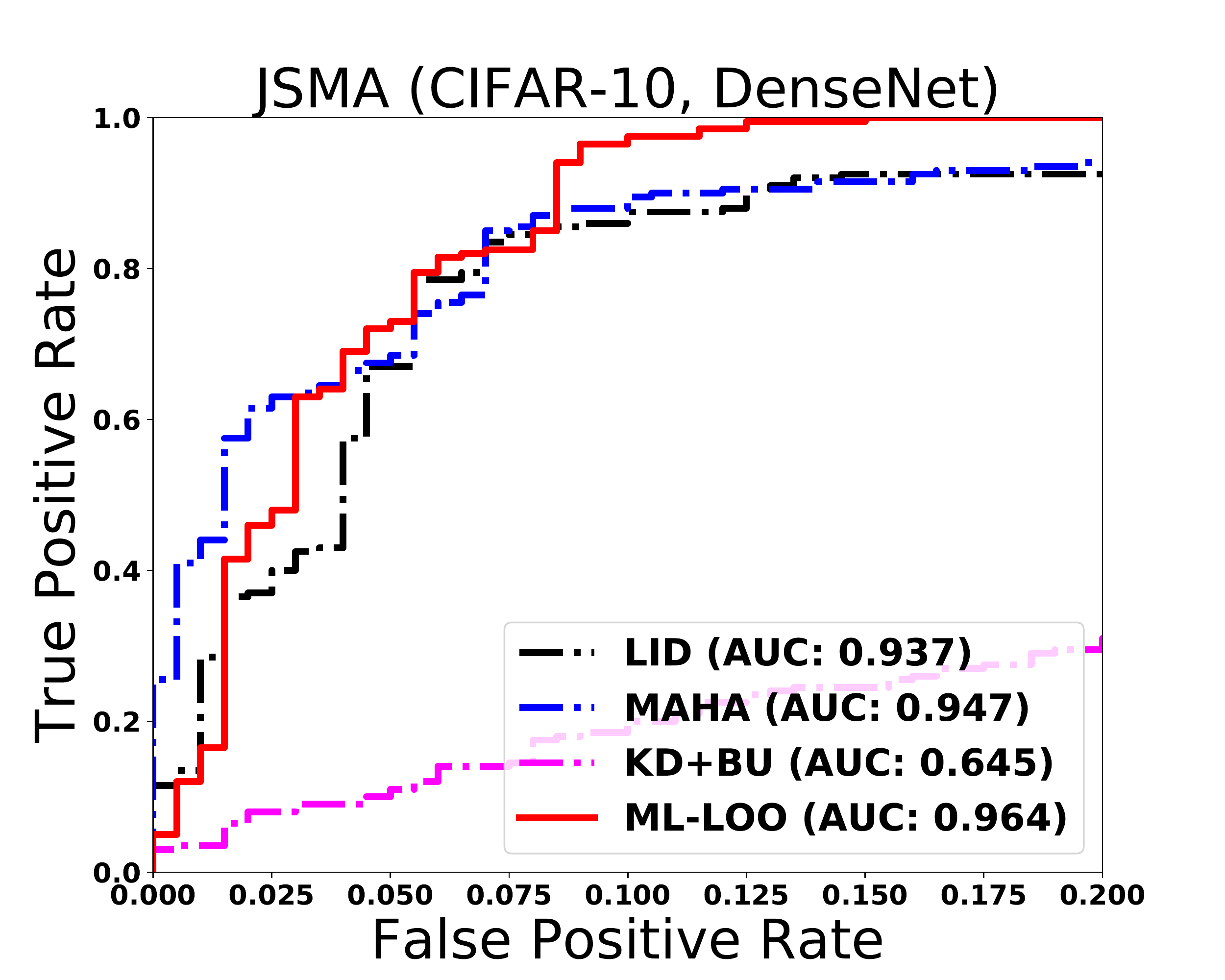} 
\includegraphics[width=0.3\linewidth]{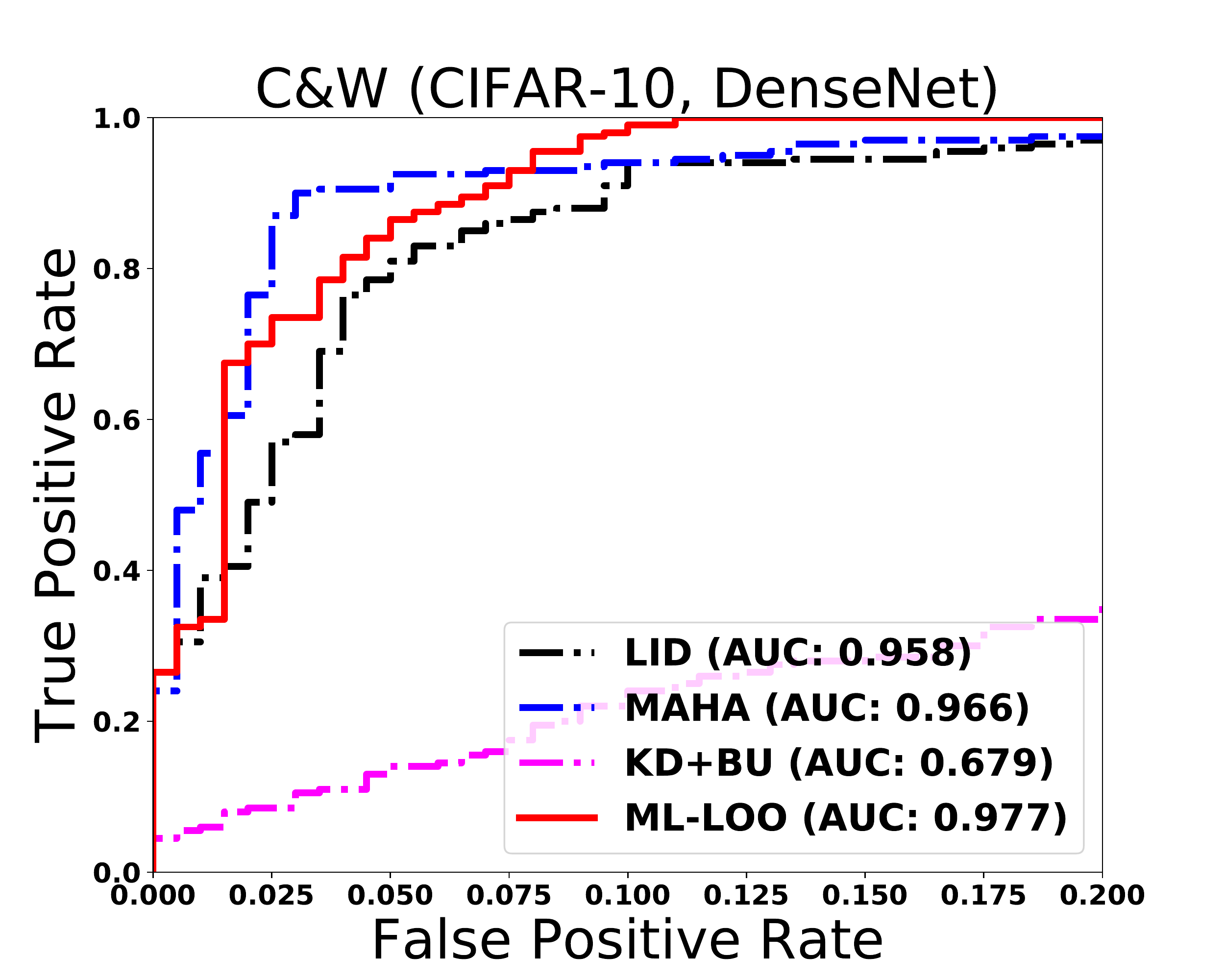} 
\includegraphics[width=0.3\linewidth]{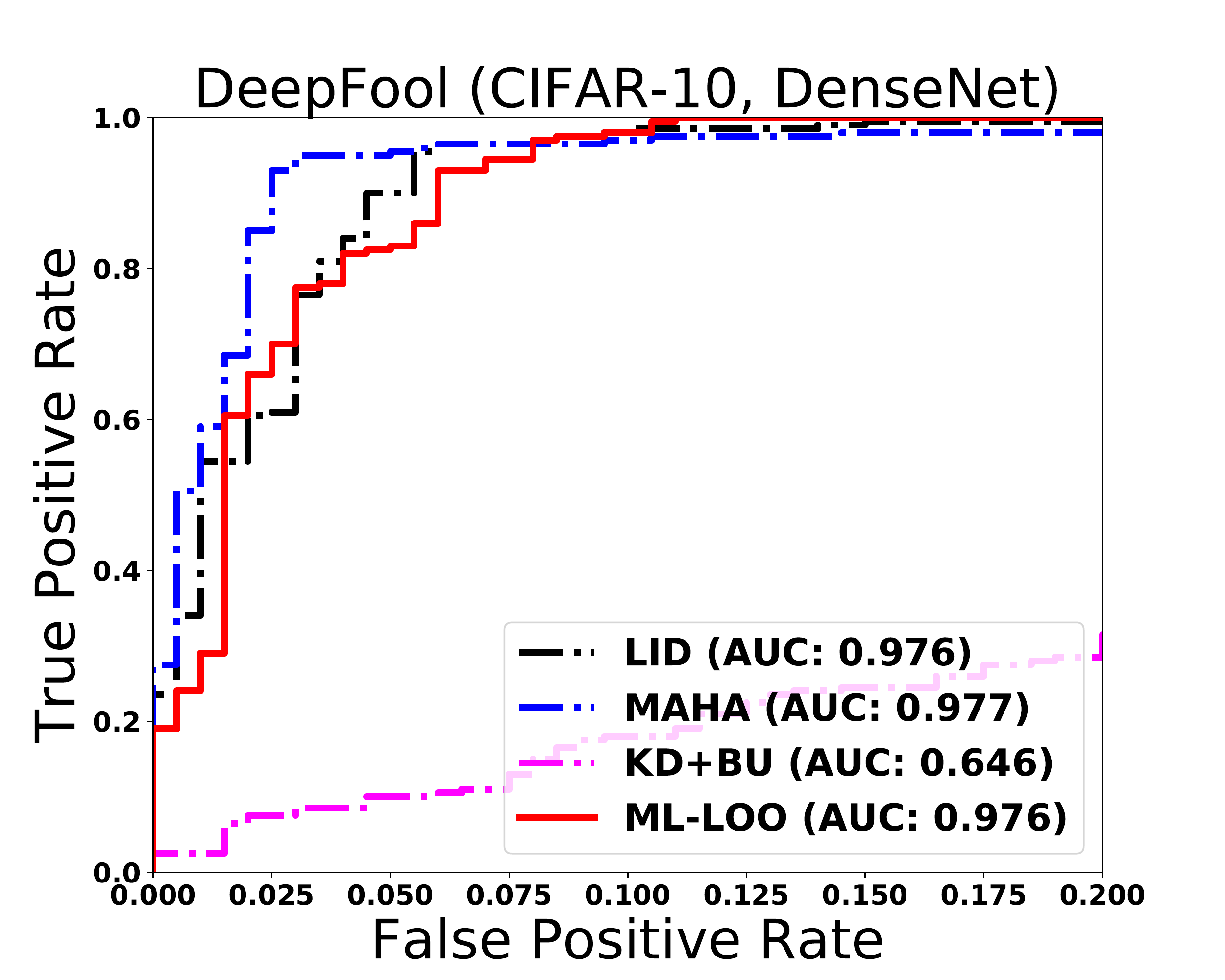} 
\includegraphics[width=0.3\linewidth]{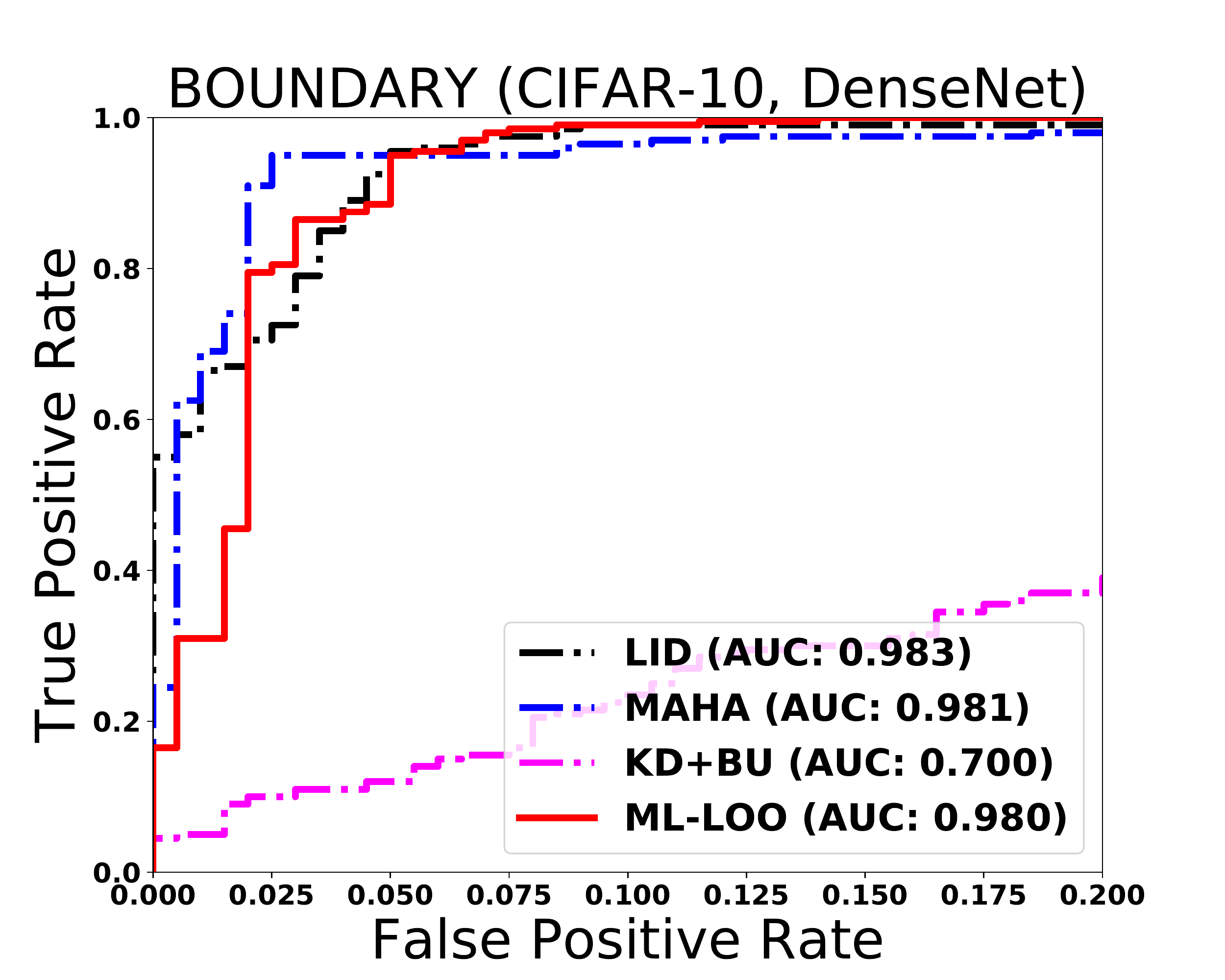} 
\includegraphics[width=0.3\linewidth]{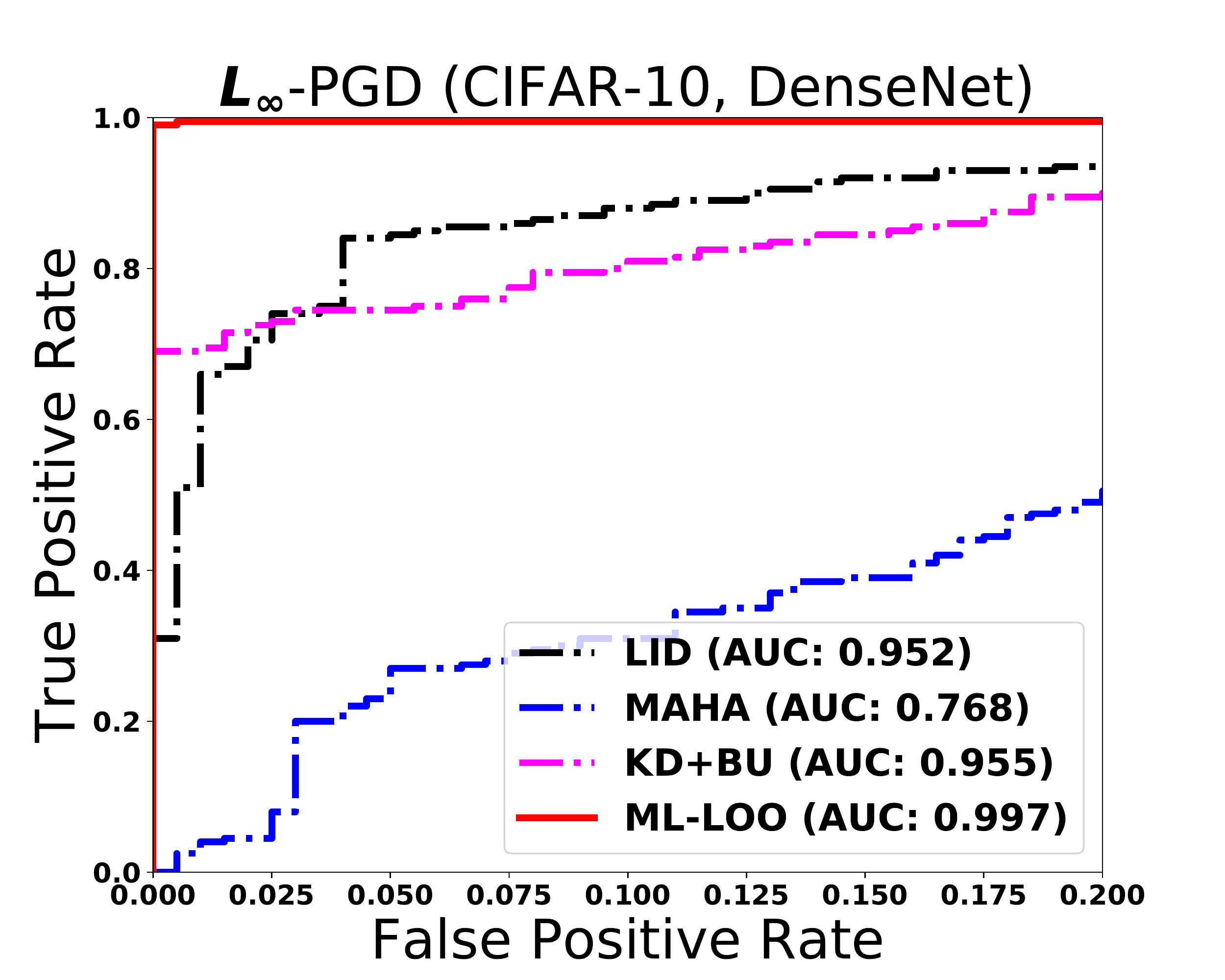}
\caption{ROC curves of detection methods on CIFAR-10 dataset with DenseNet}
\label{fig:CIFAR10DENSENET}
\end{figure}

\begin{figure}[H]
\centering 
\includegraphics[width=0.3\linewidth]{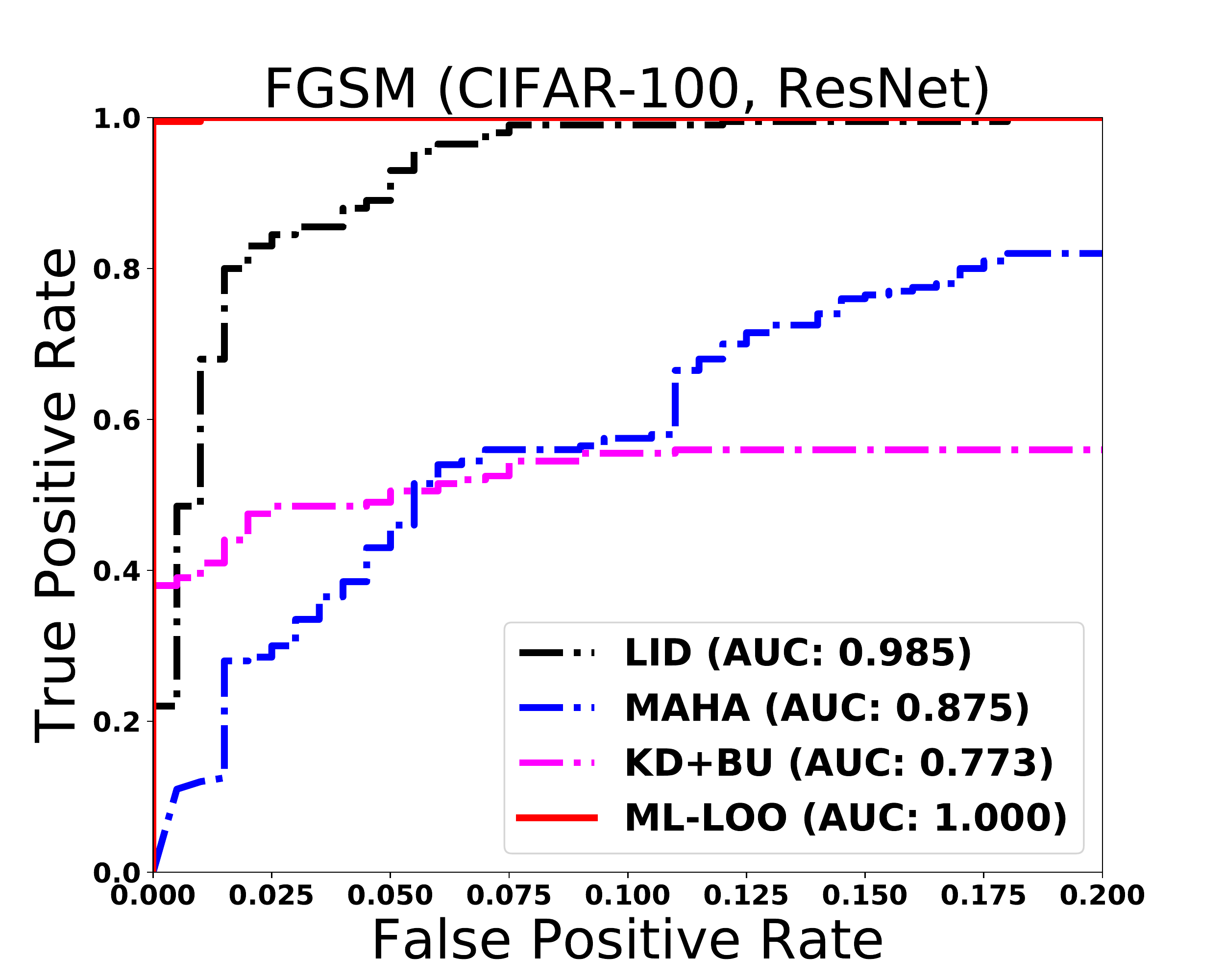} 
\includegraphics[width=0.3\linewidth]{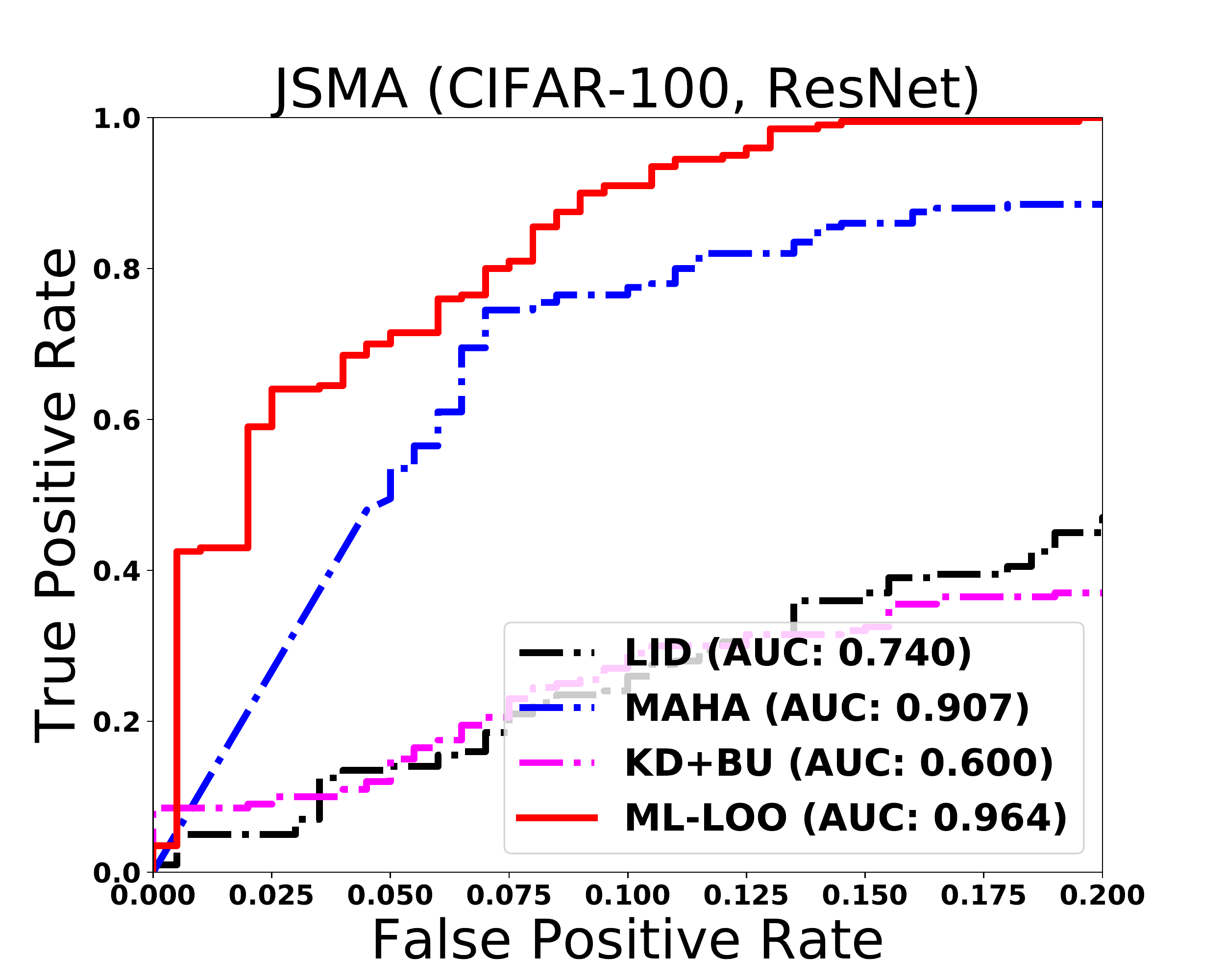} 
\includegraphics[width=0.3\linewidth]{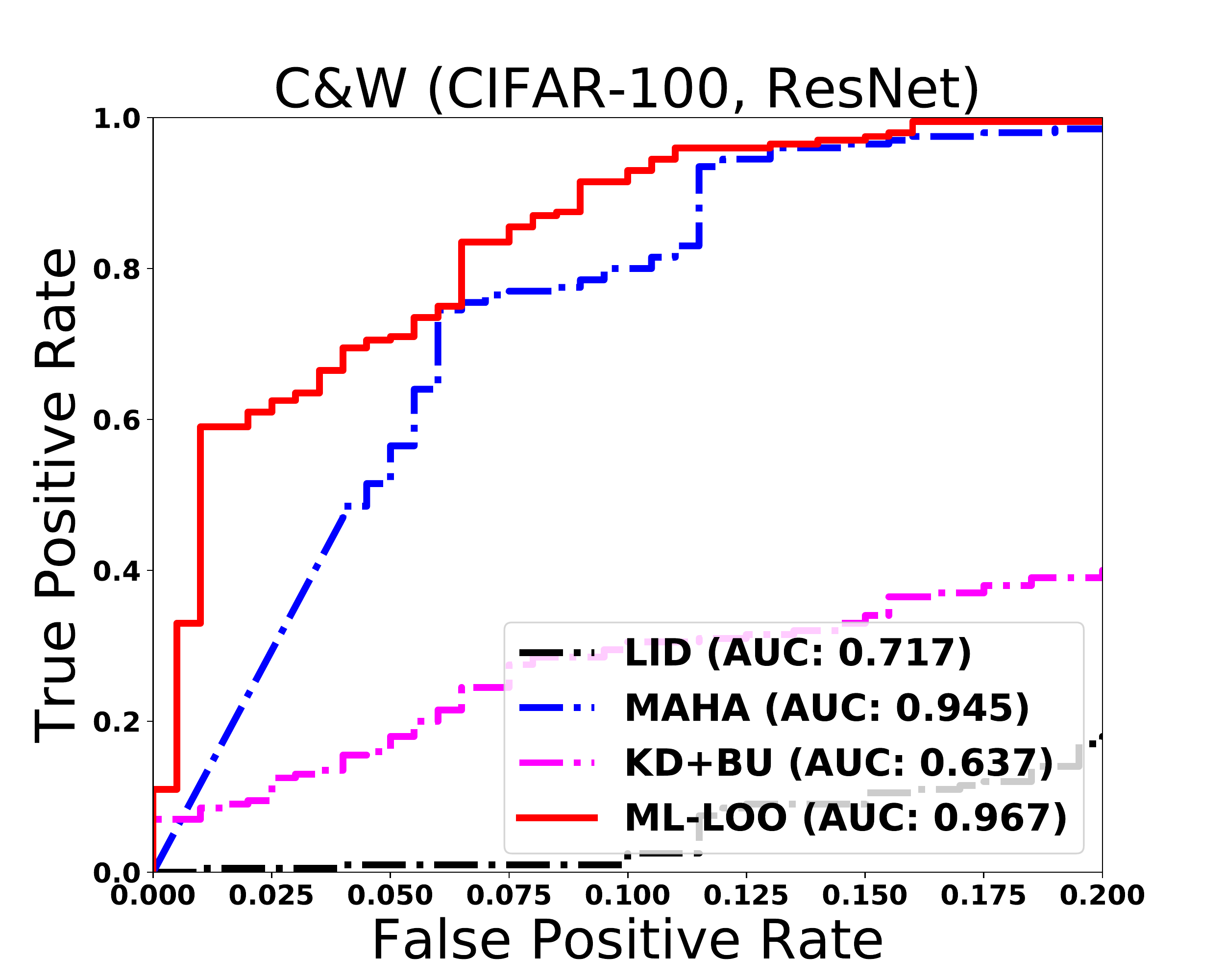} 
\includegraphics[width=0.3\linewidth]{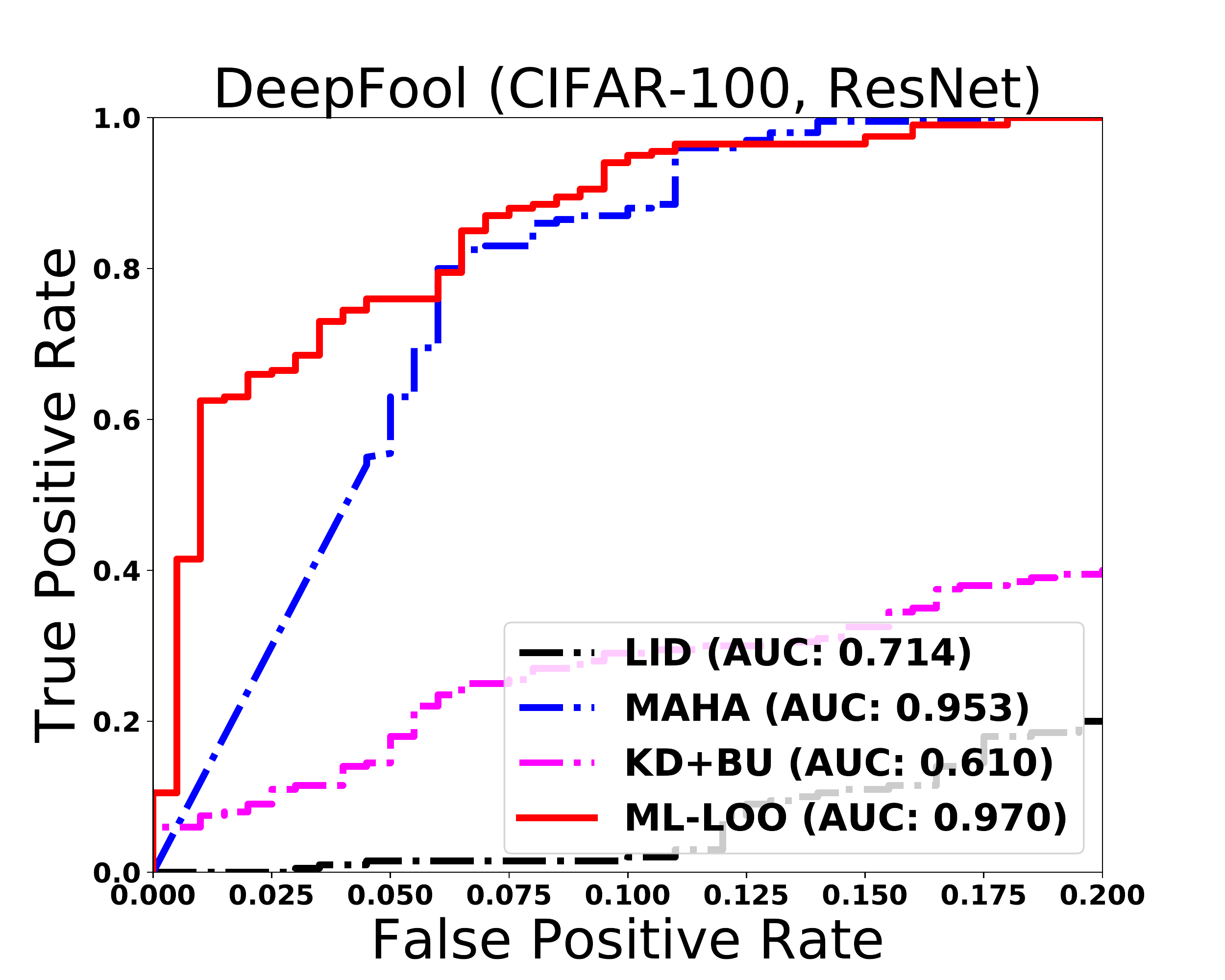} 
\includegraphics[width=0.3\linewidth]{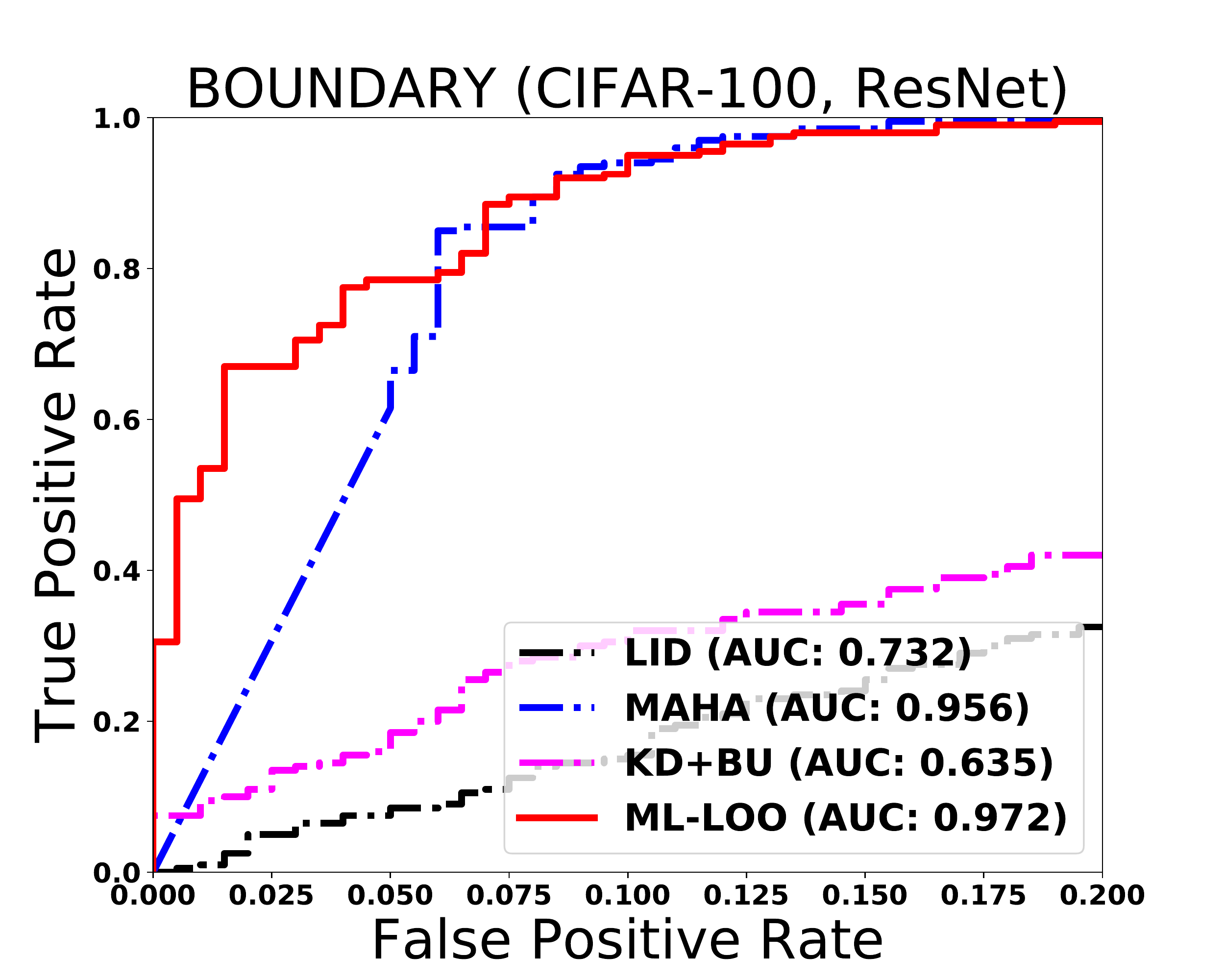} 
\includegraphics[width=0.3\linewidth]{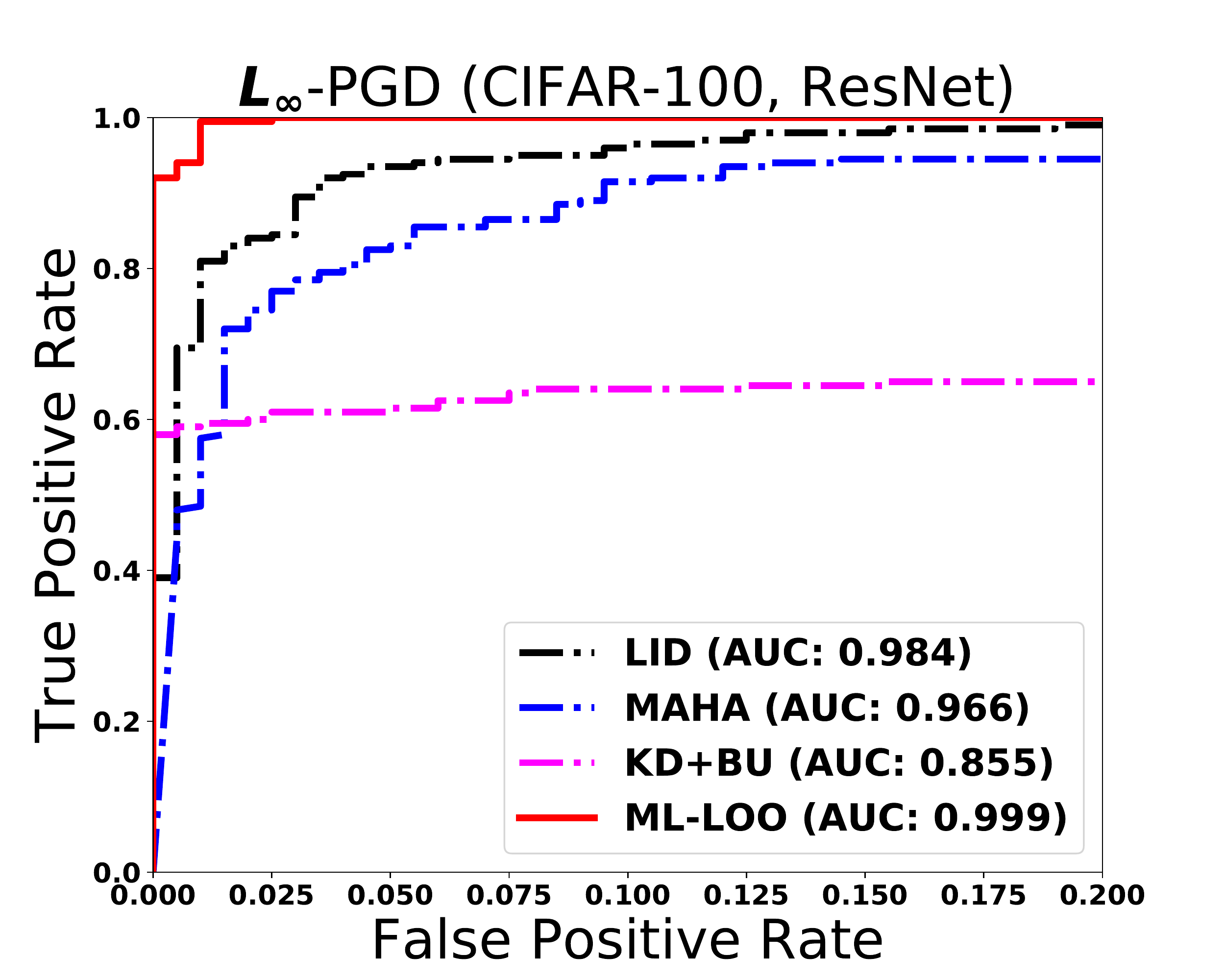}
\caption{ROC curves of detection methods on CIFAR-100 dataset with ResNet}
\label{fig:CIAFR100RESNET}
\end{figure}

\begin{figure}[H]
\centering 
\includegraphics[width=0.3\linewidth]{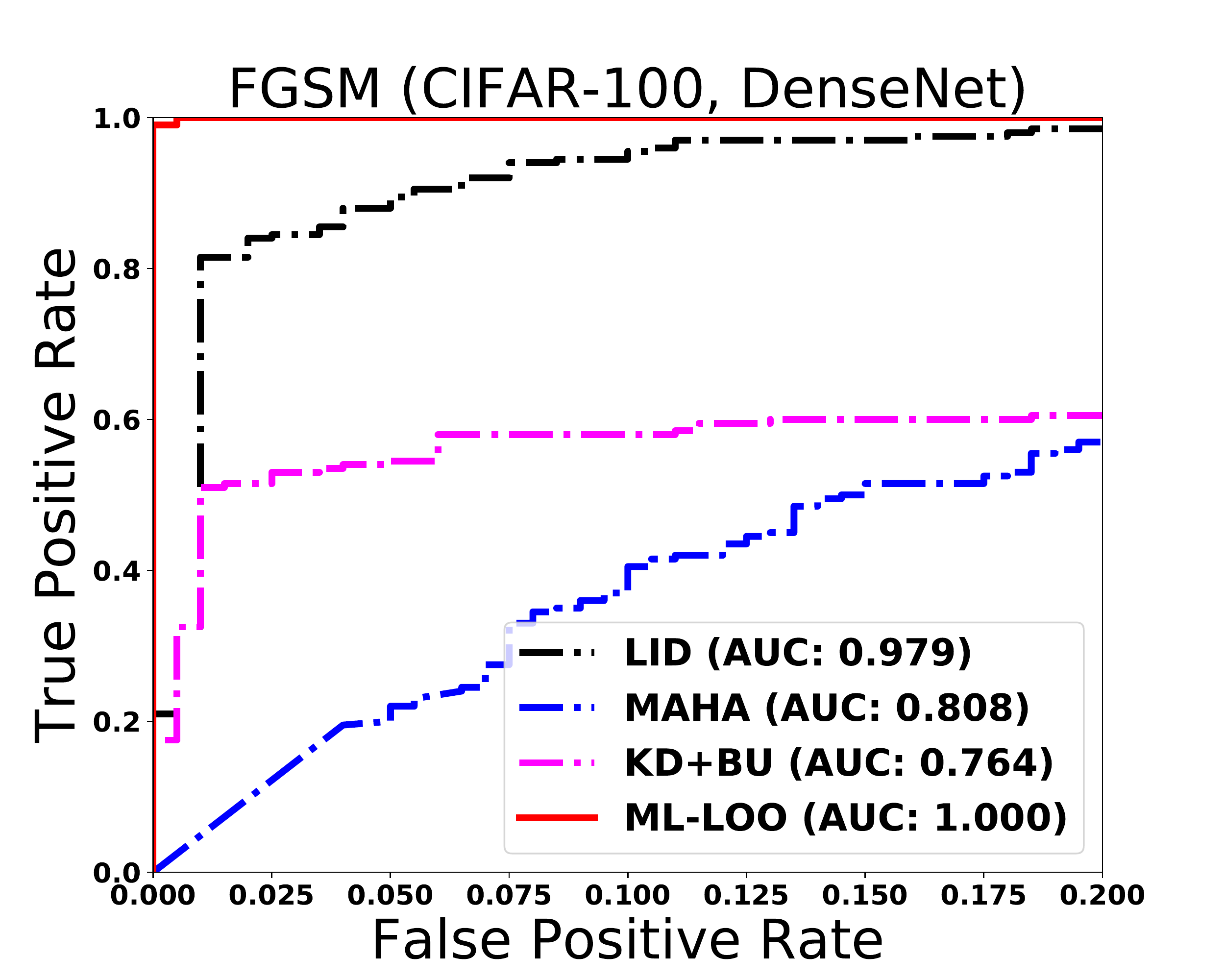} 
\includegraphics[width=0.3\linewidth]{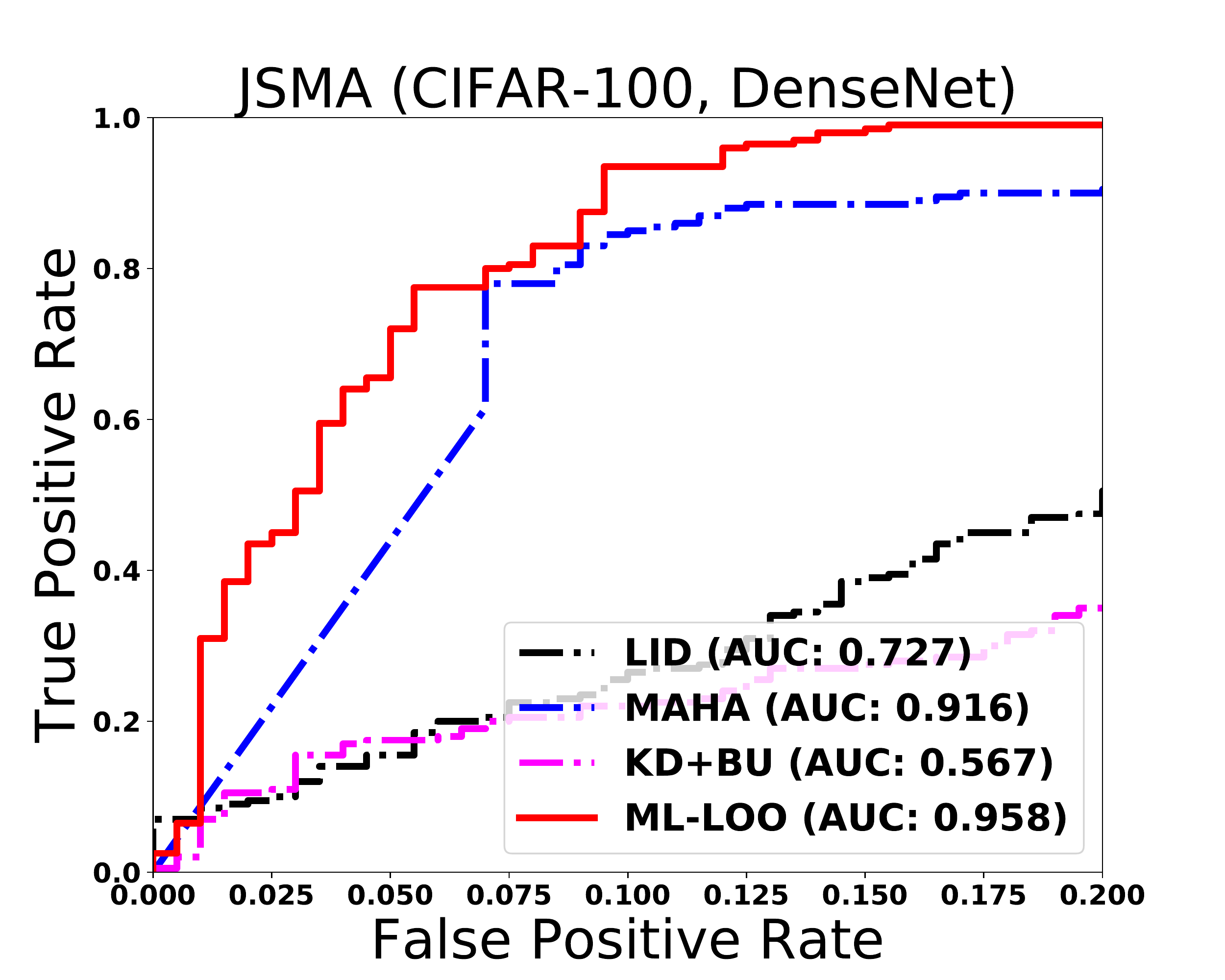} 
\includegraphics[width=0.3\linewidth]{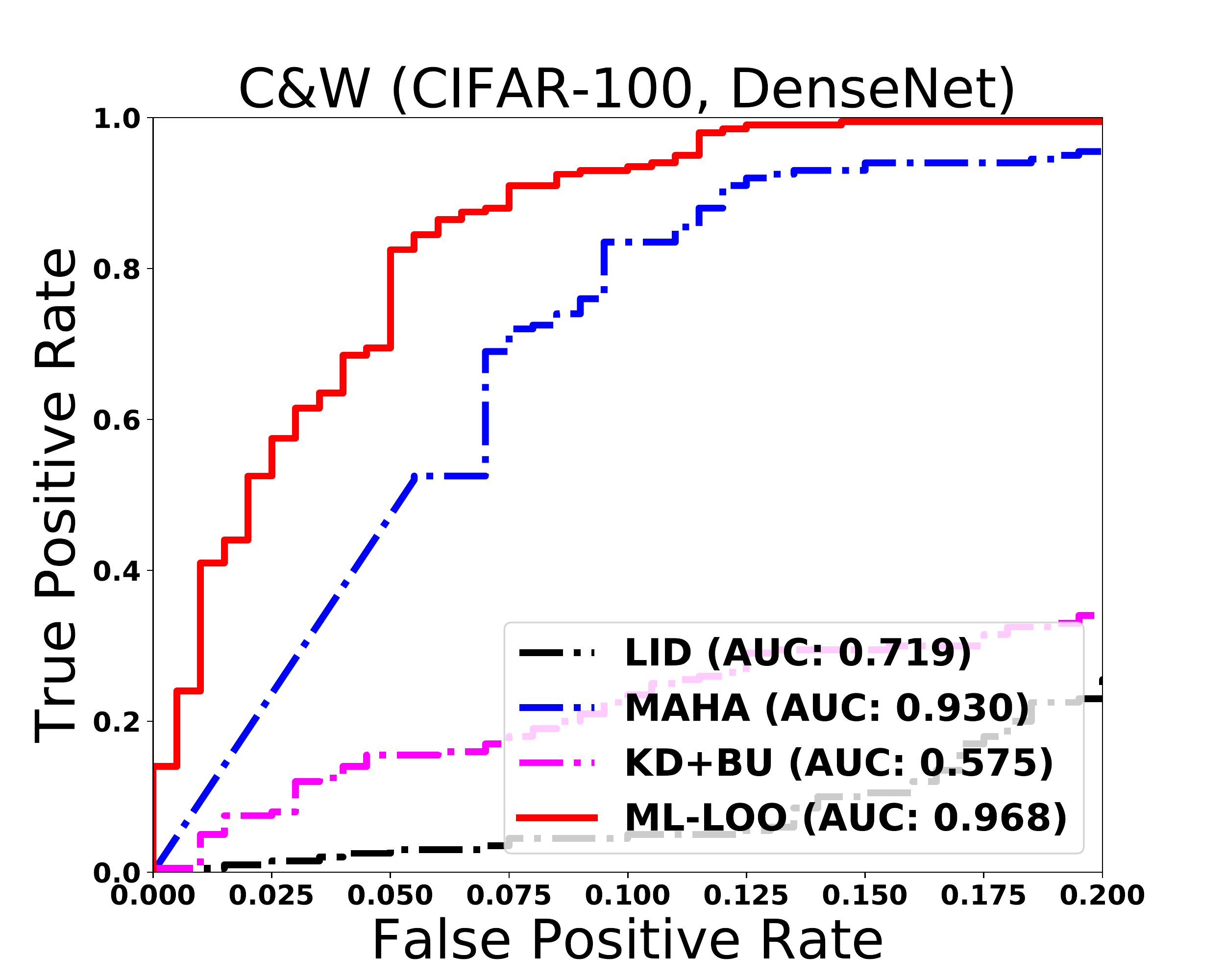} 
\includegraphics[width=0.3\linewidth]{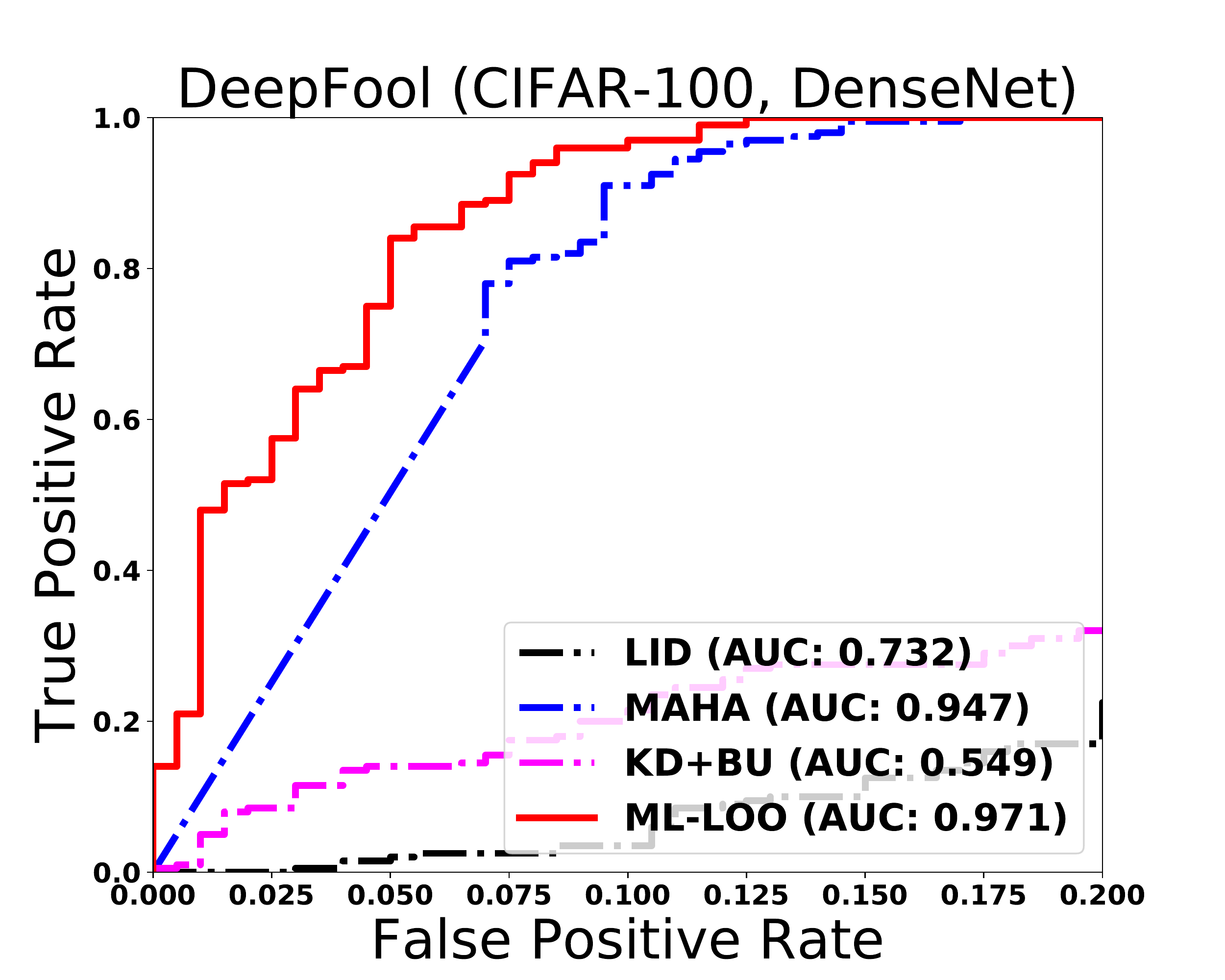} 
\includegraphics[width=0.3\linewidth]{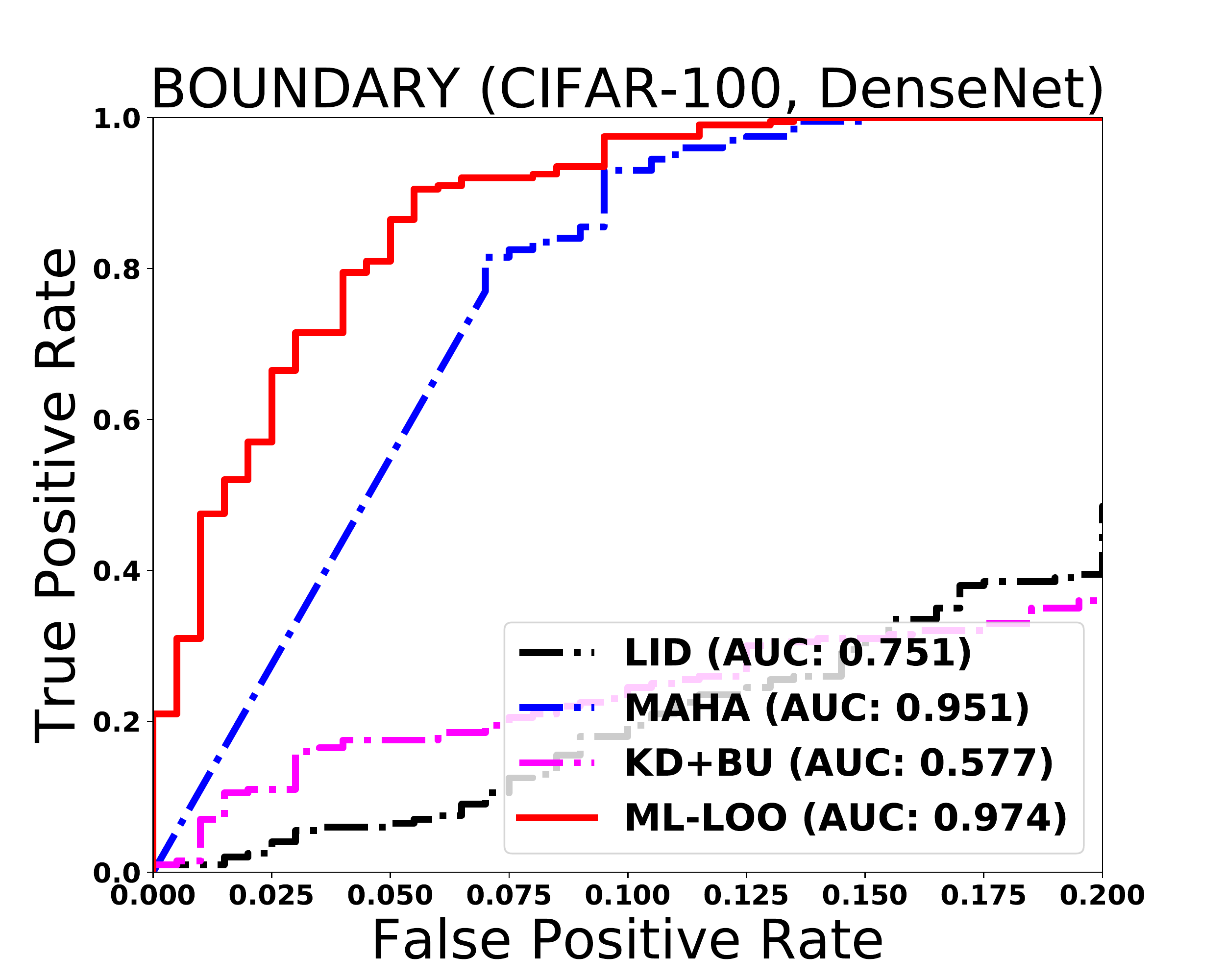} 
\includegraphics[width=0.3\linewidth]{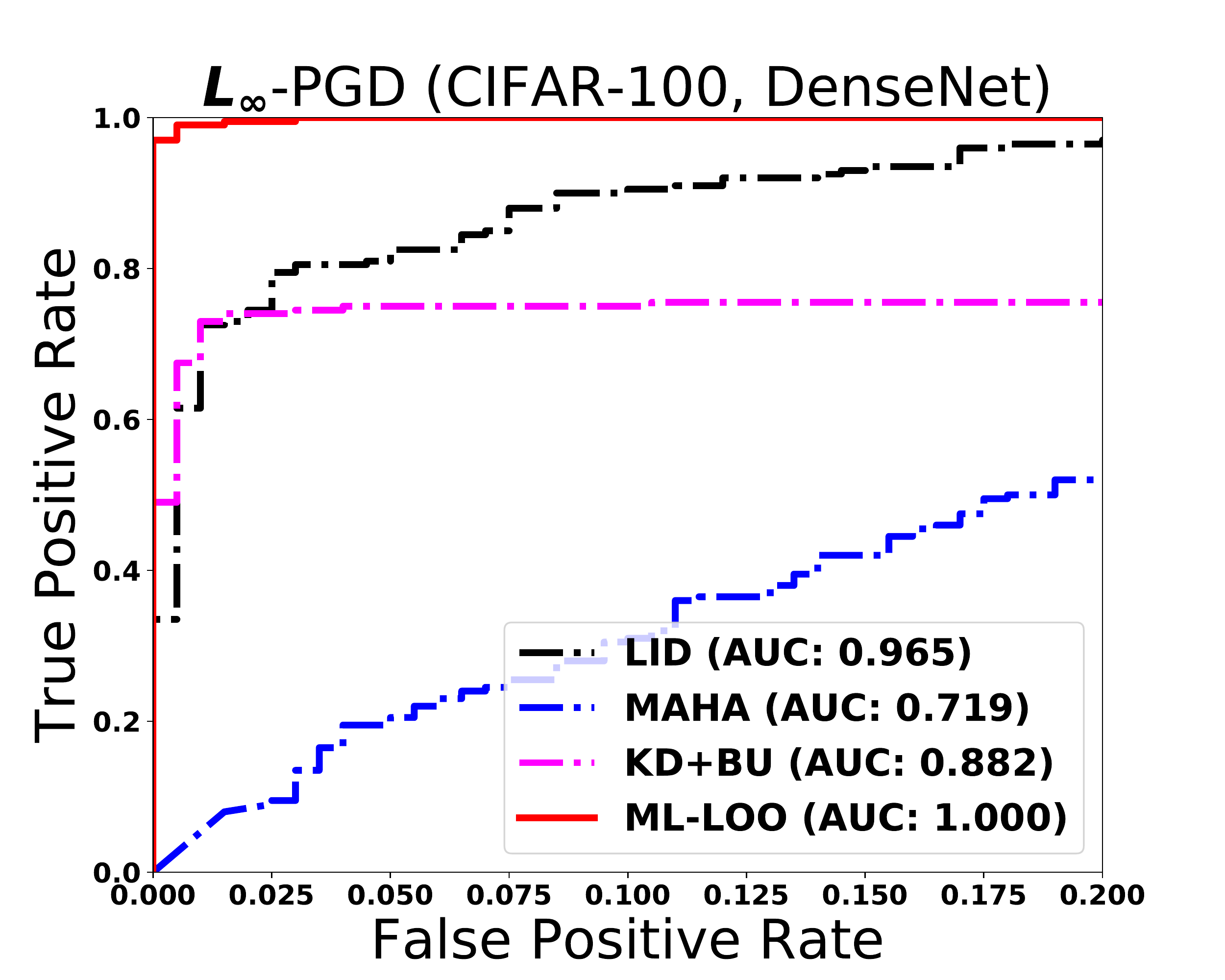}
\caption{ROC curves of detection methods on CIFAR-100 dataset with DenseNet}
\label{fig:CIAFR100DENSENET}
\end{figure}

Figure~\ref{fig:MNISTSCNN}, Figure~\ref{fig:CIFAR10RESNET2}, Figure~\ref{fig:CIFAR10DENSENET}, Figure~\ref{fig:CIAFR100RESNET}, and Figure~\ref{fig:CIAFR100DENSENET} show the ROC curves of four detection method (LID, MAHA, KD+BU, ML-LOO) on three data sets (CIFAR-10, MNIST, CIFAR-100) with three models (CNN, ResNet, DenseNet) under six attacks (FGSM, JSMA, C\&W, DeepFool, Boundary, $L_\infty$-PGD) where FPR is from 0.0 to 0.2, which is the setting of practical interest. The ROC curves where FPR is from 0.0 to 1.0 are shown in Appendix~\ref{sec:roc_plots}. 

\vspace{-5mm}

\newpage
\subsection{ROC curves of detection methods on CIFAR-10, MNIST and CIFAR-100 data sets with FPR from 0.0 to 1.0}
\label{sec:roc_plots}

In this section, we show the ROC curves of four detection method (LID, MAHA, KD+BU, ML-LOO) on three data sets (CIFAR-10, MNIST, CIFAR-100) with three models (CNN, ResNet, DenseNet) under six attacks (FGSM, JSMA, C\&W, DeepFool, Boundary, $L_\infty$-PGD) where FPR is from 0.0 to 1.0.


\begin{figure}[H]
\centering 
\includegraphics[width=0.3\linewidth]{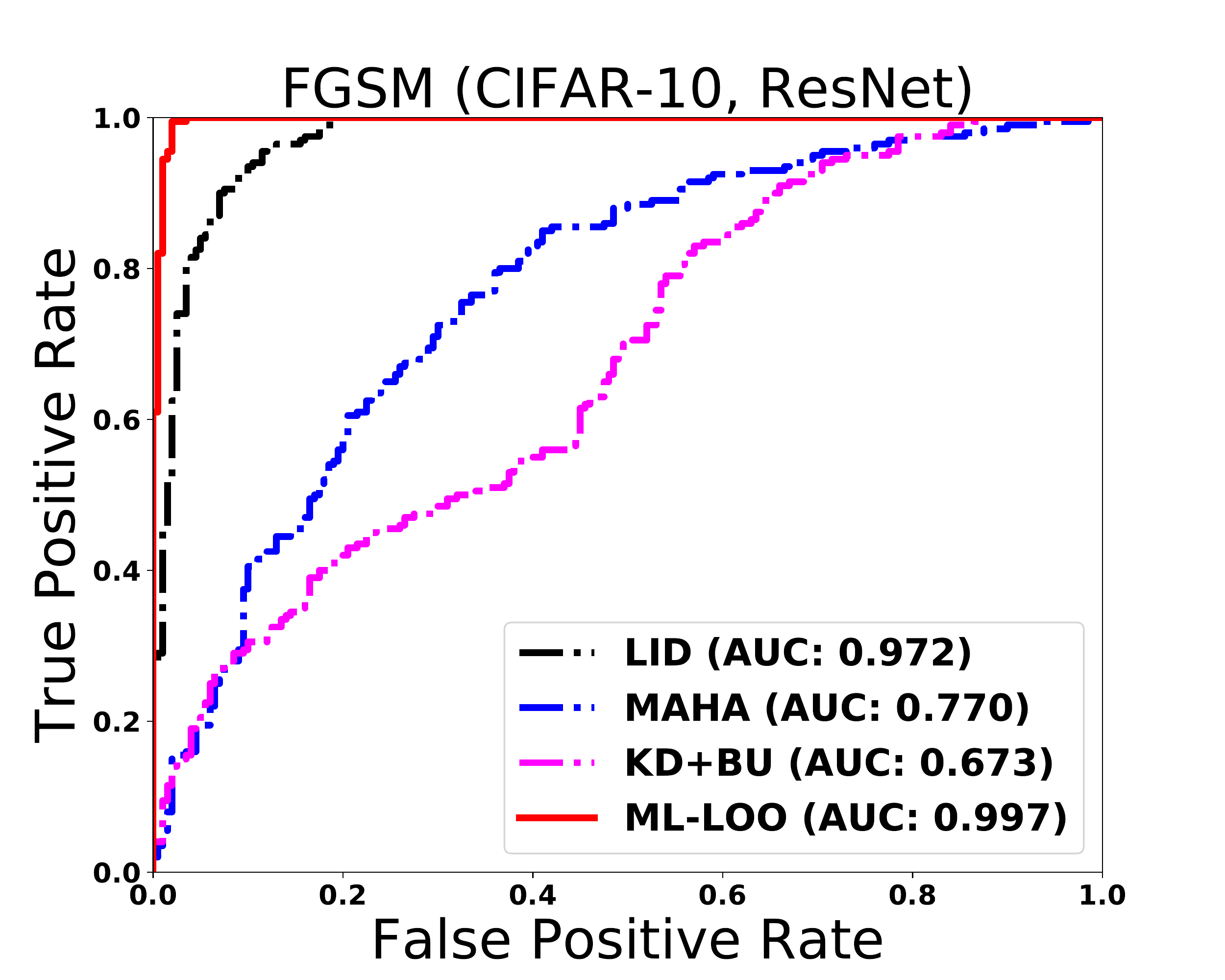} 
\includegraphics[width=0.3\linewidth]{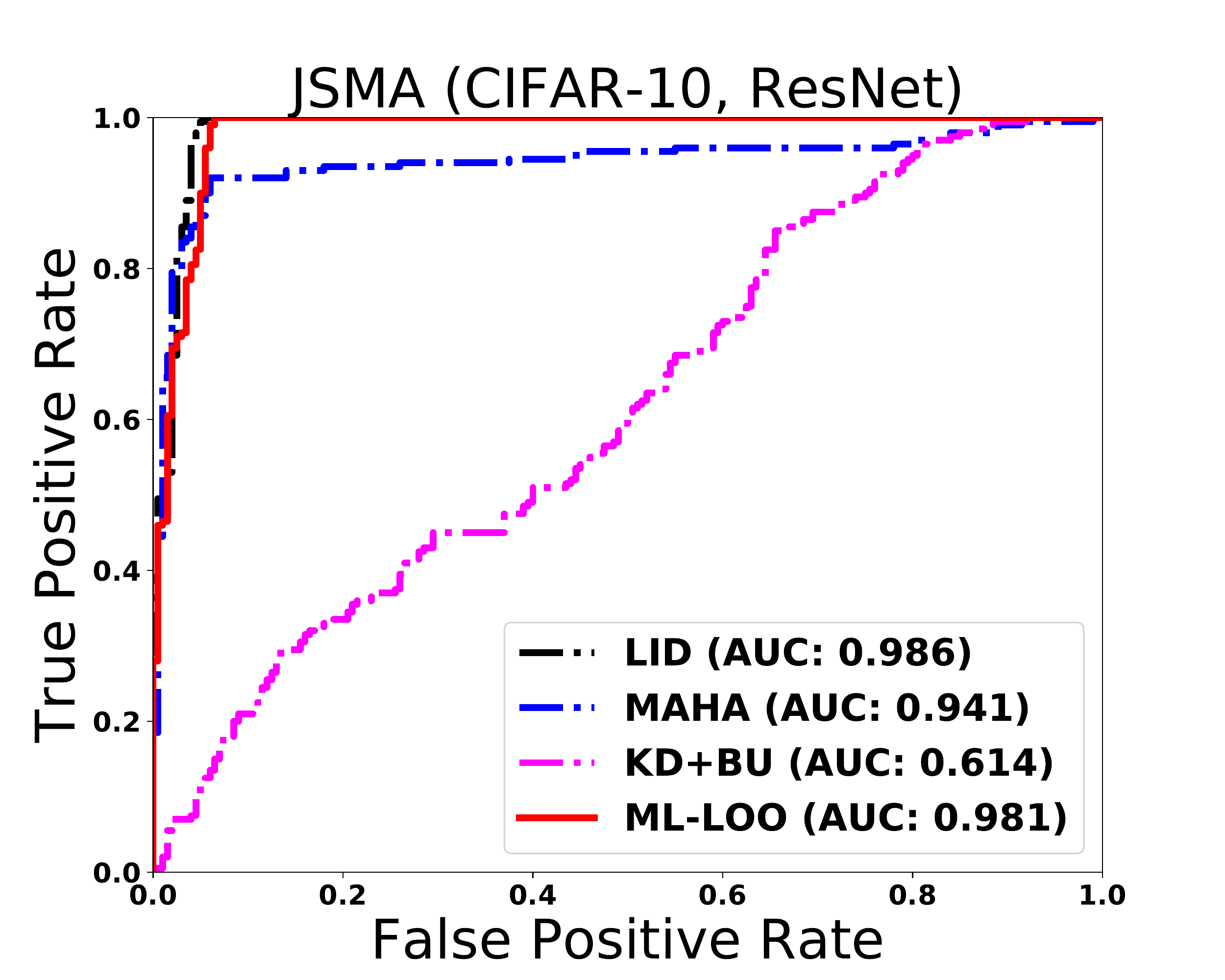} 
\includegraphics[width=0.3\linewidth]{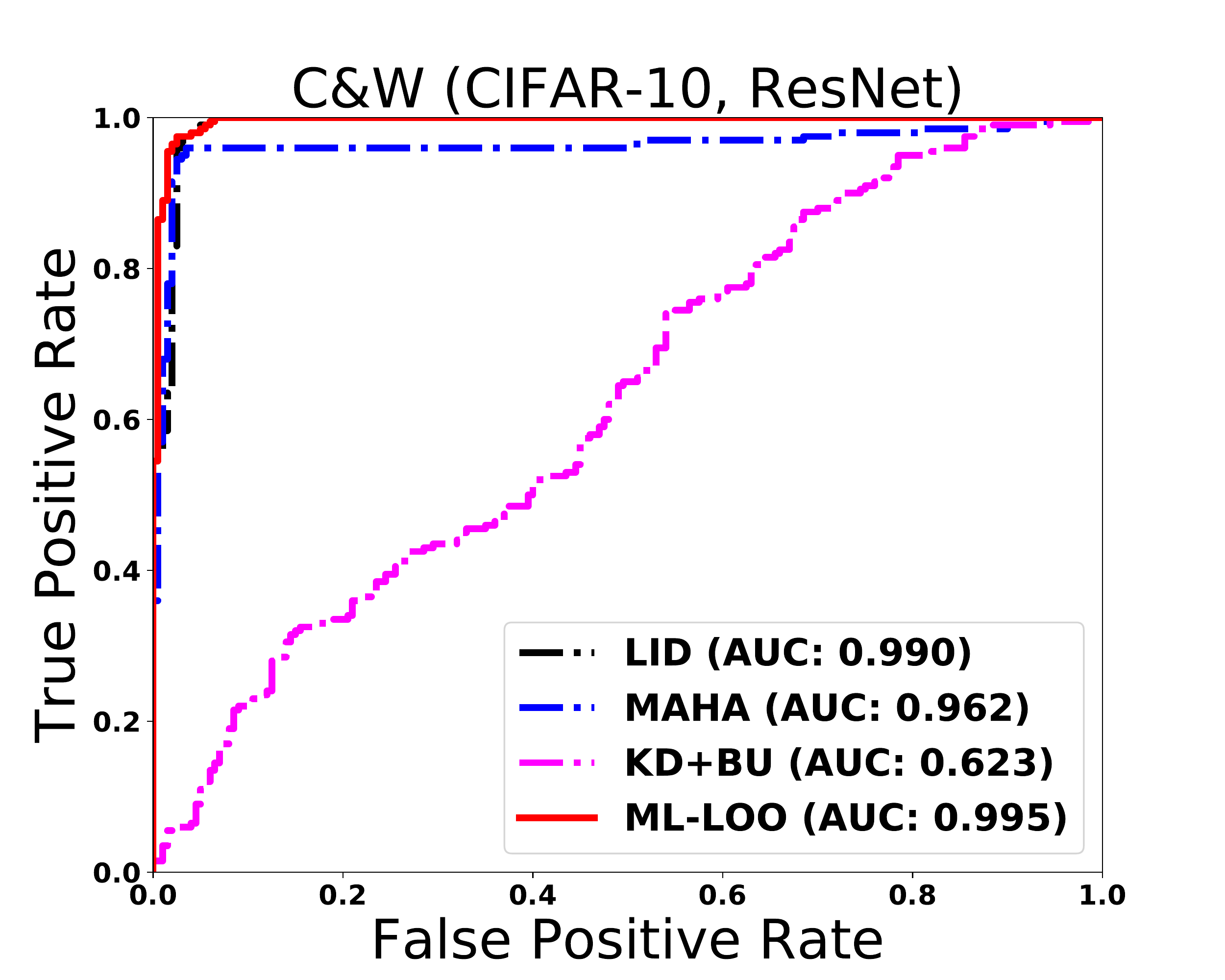} 
\includegraphics[width=0.3\linewidth]{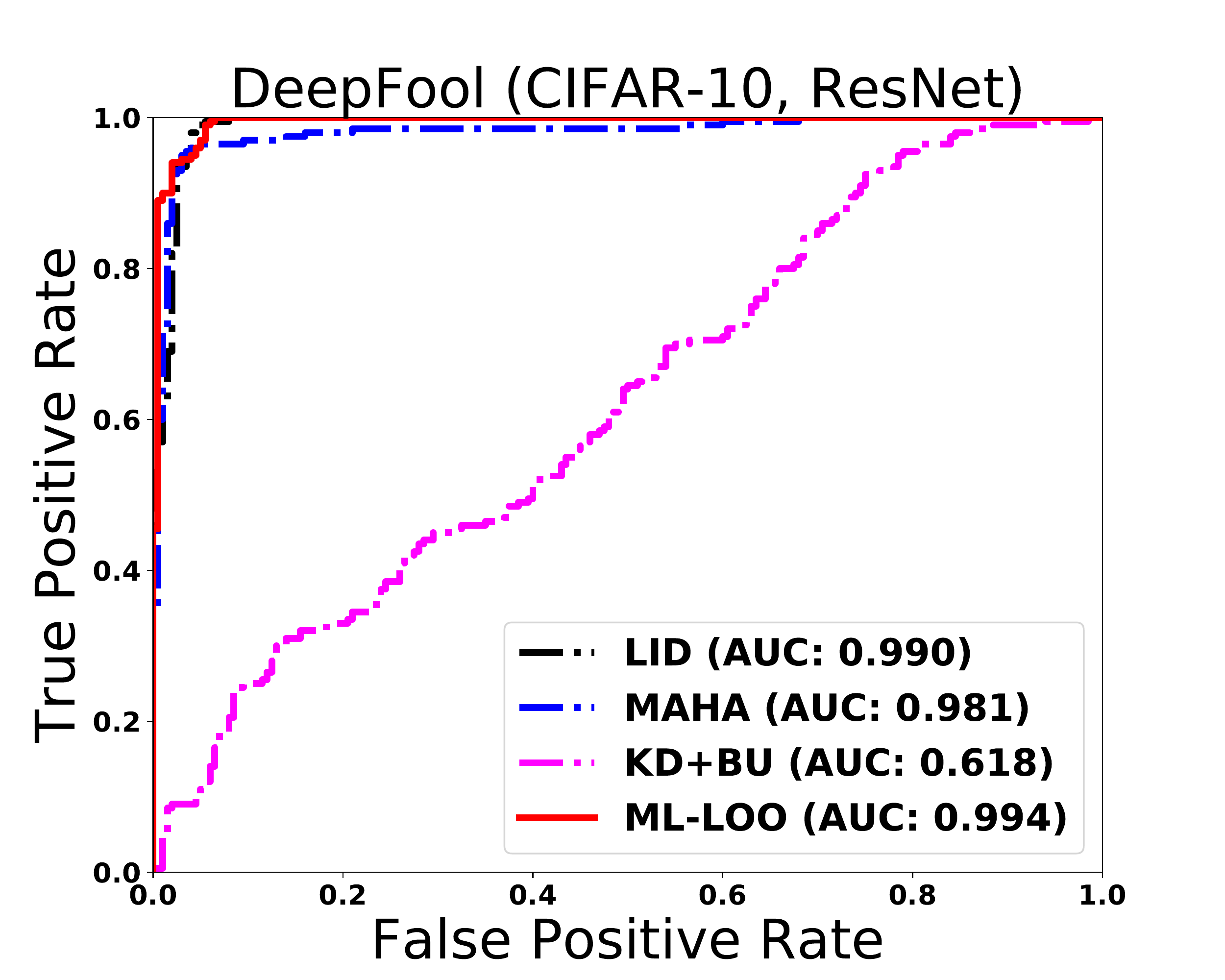} 
\includegraphics[width=0.3\linewidth]{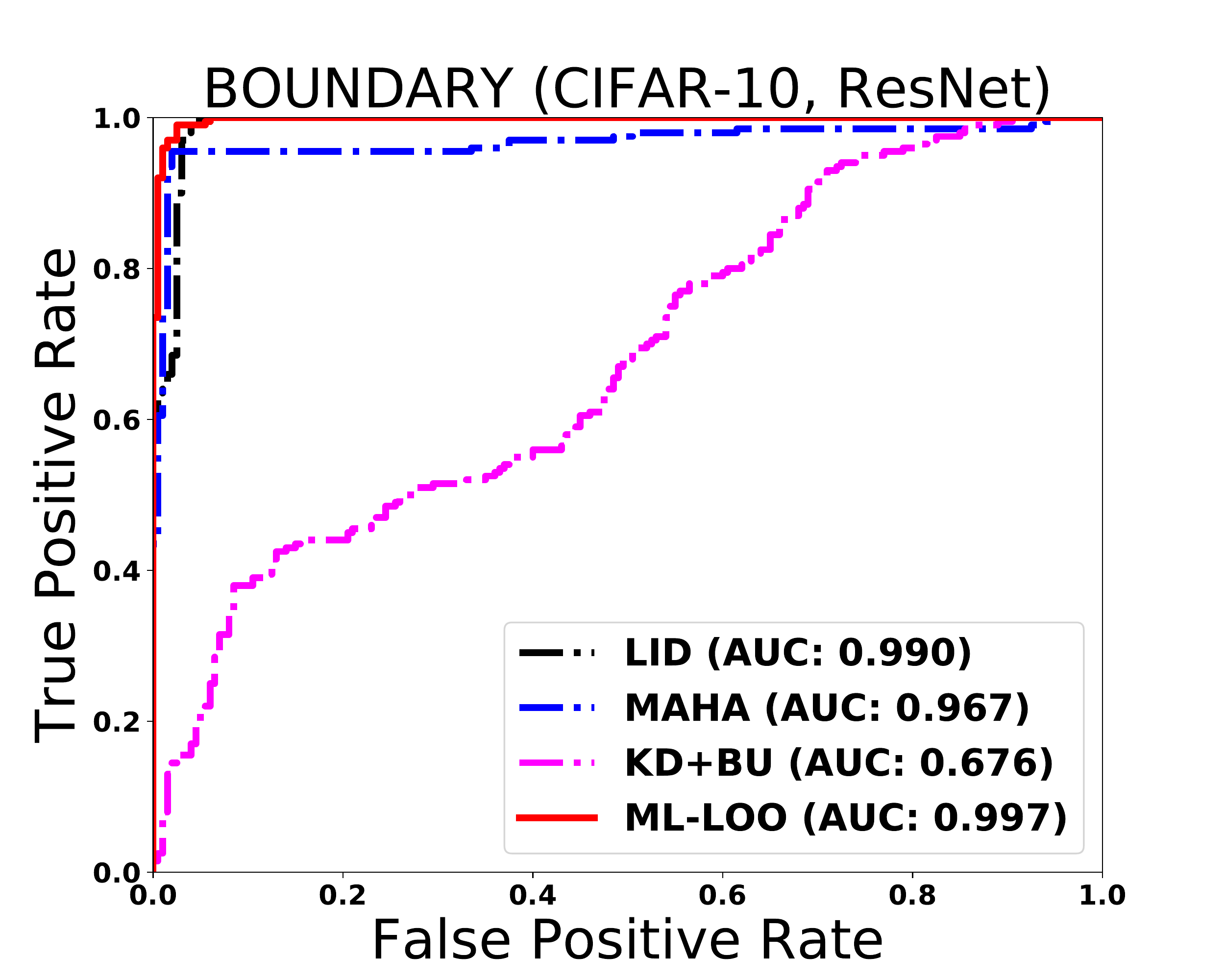} 
\includegraphics[width=0.3\linewidth]{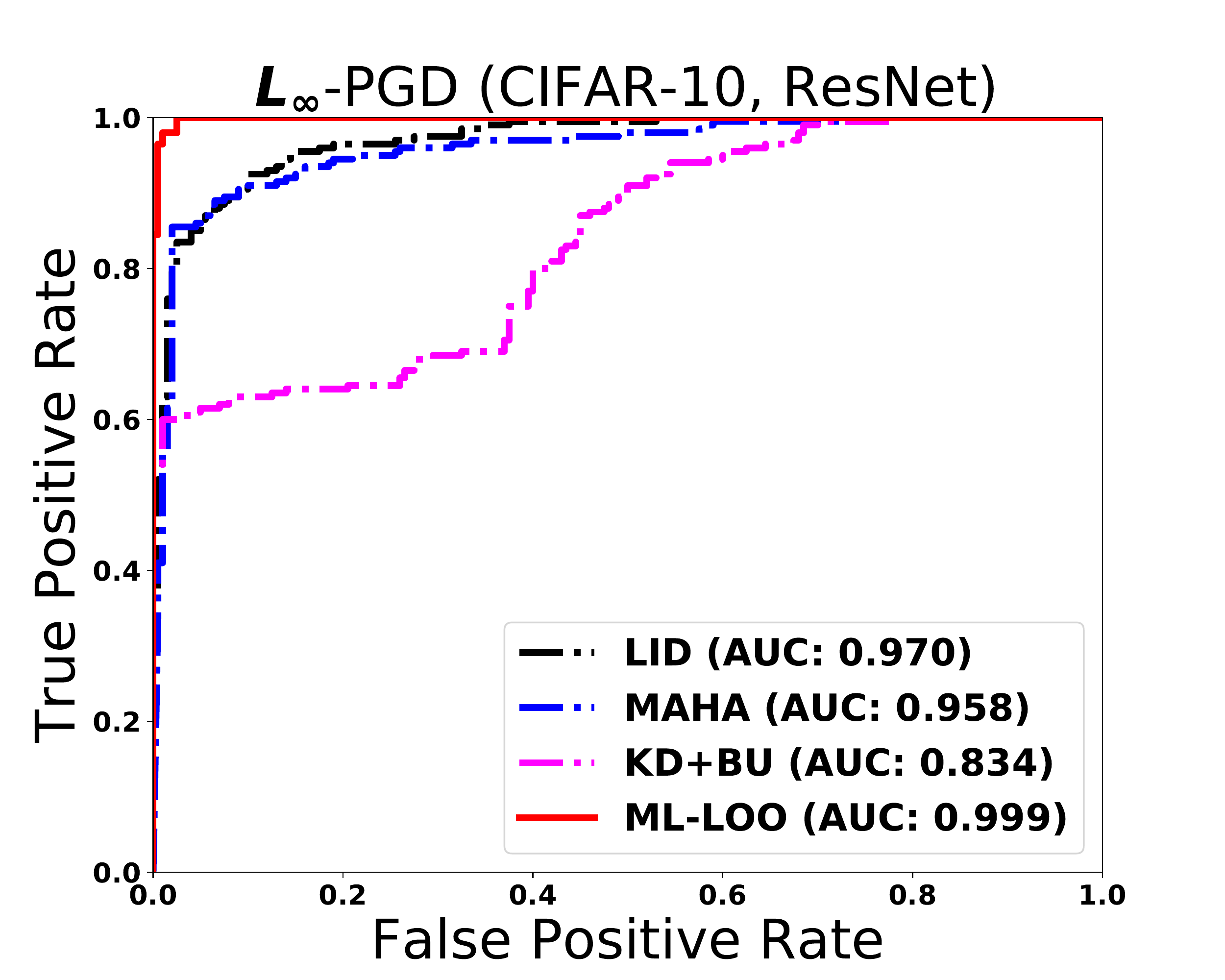}
\caption{ROC curves of detection methods on CIFAR-10 dataset with ResNet}
\label{fig:CIFAR10RESNET1}
\end{figure} 

\begin{figure}[H]
\centering 
\includegraphics[width=0.3\linewidth]{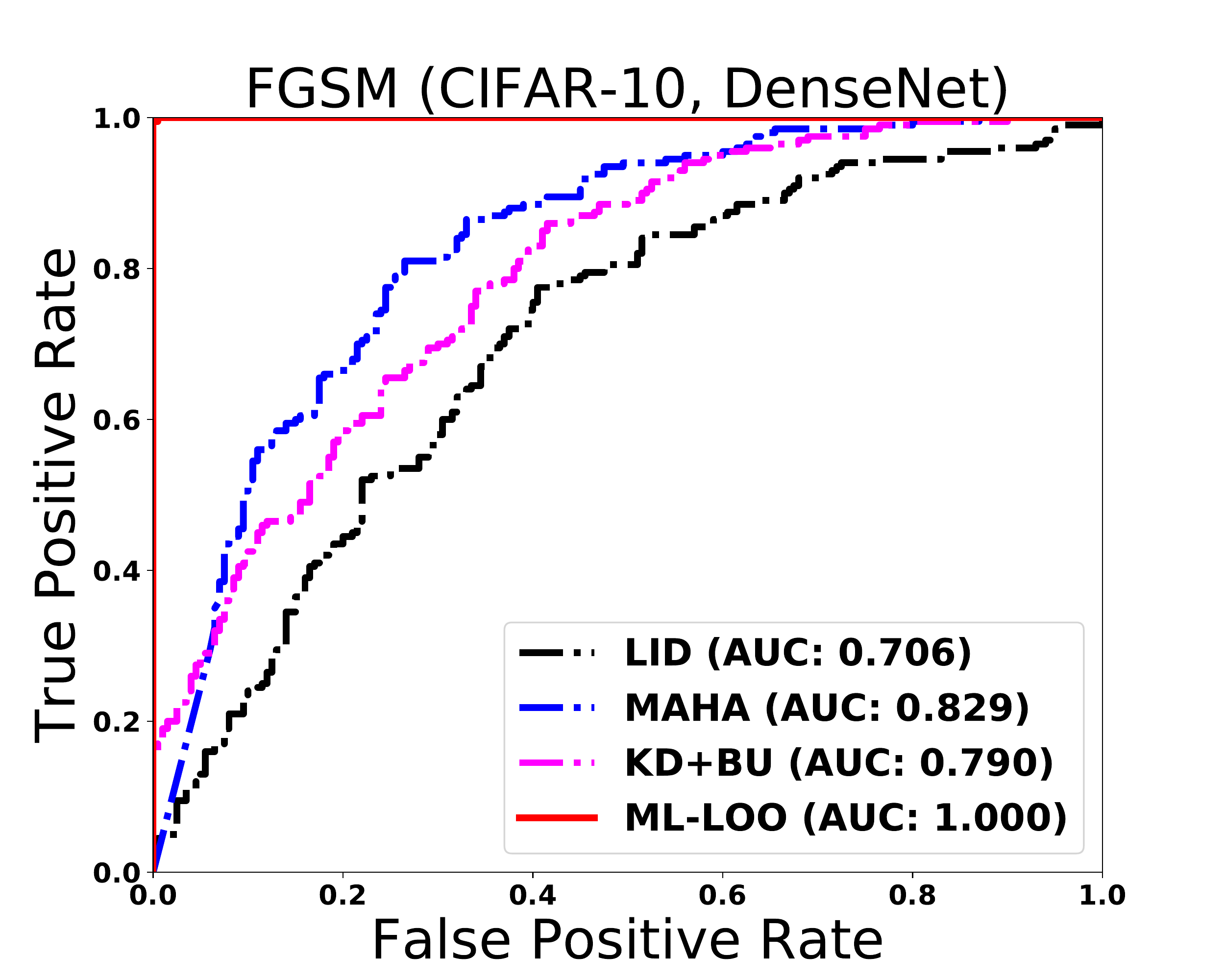} 
\includegraphics[width=0.3\linewidth]{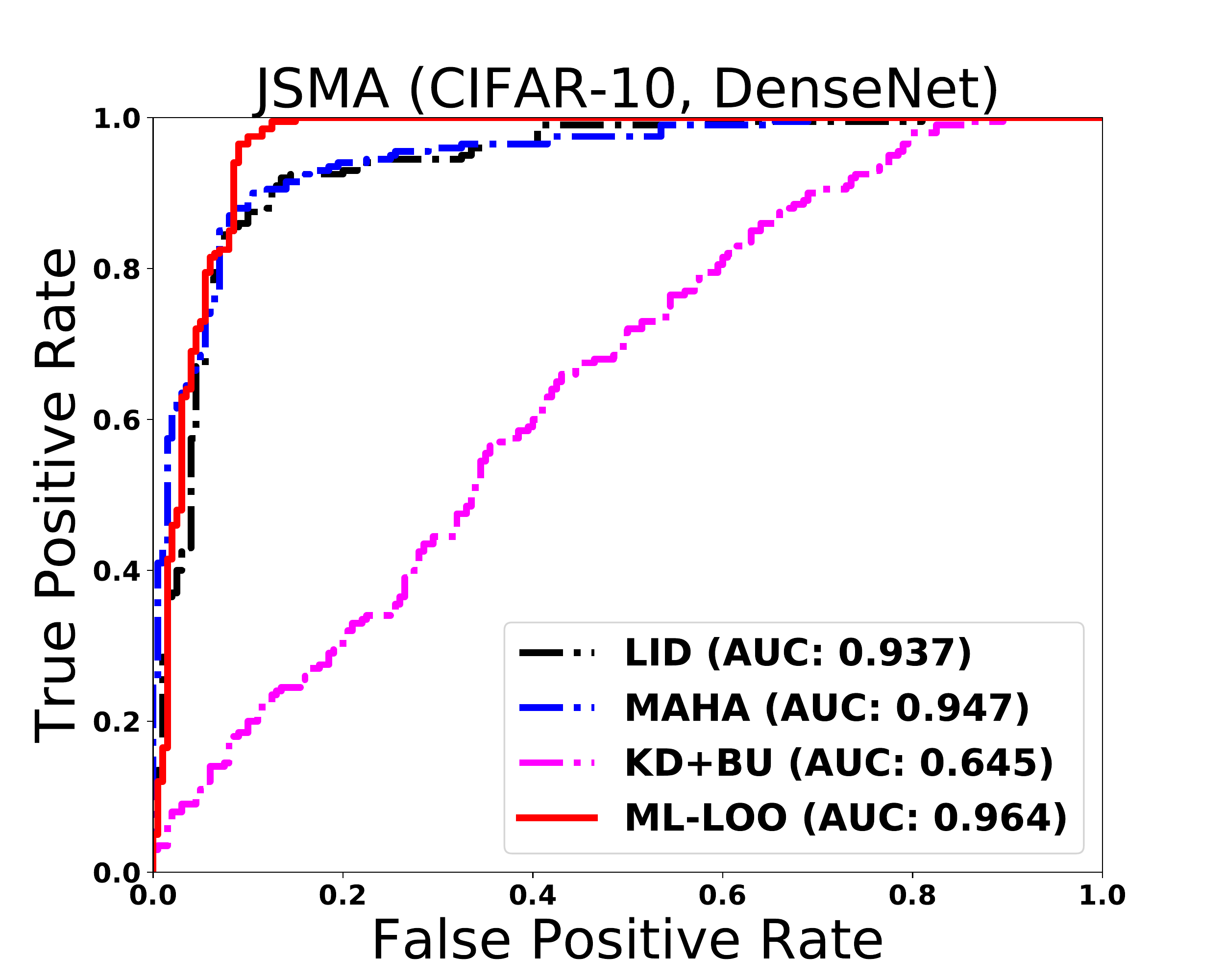} 
\includegraphics[width=0.3\linewidth]{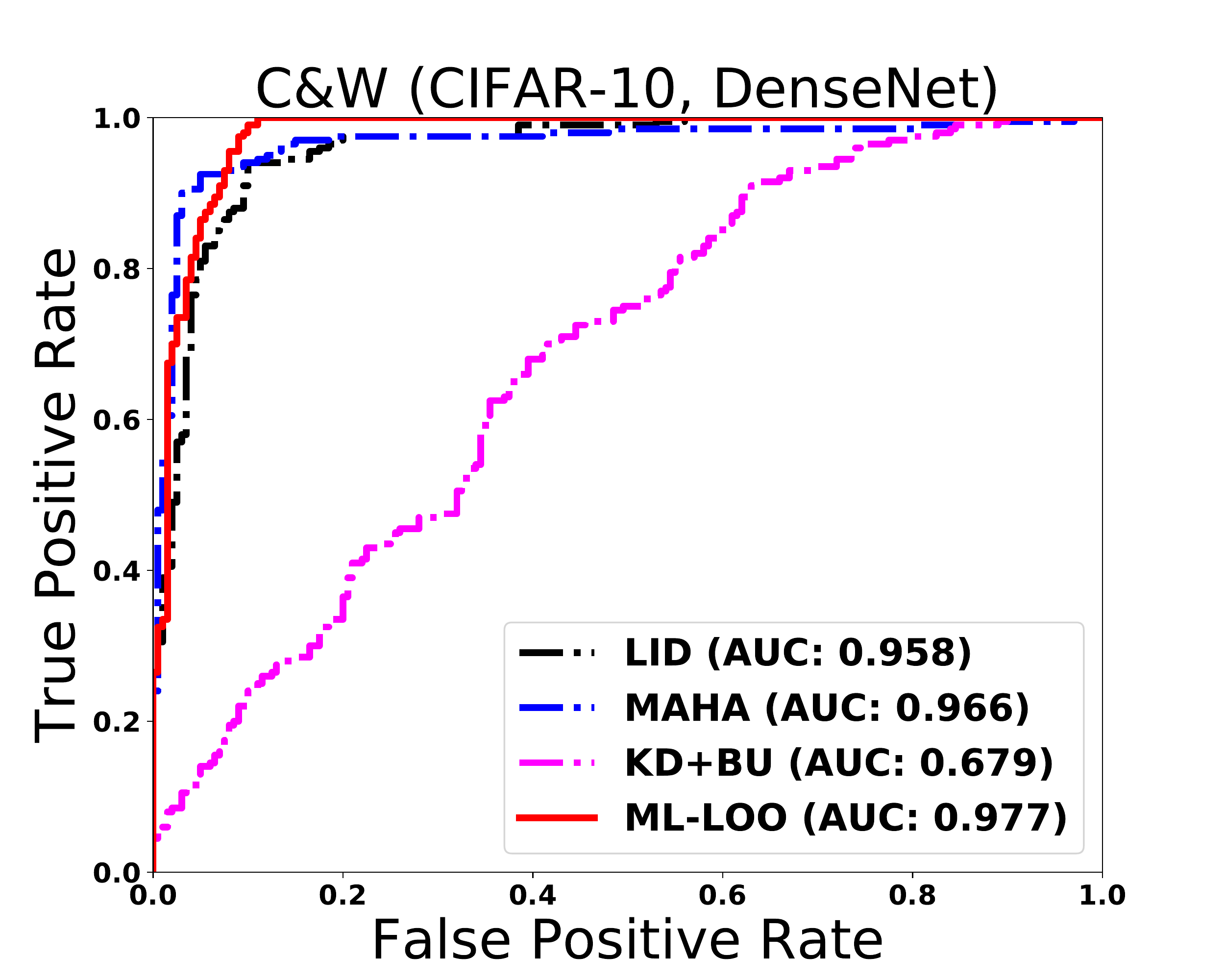} 
\includegraphics[width=0.3\linewidth]{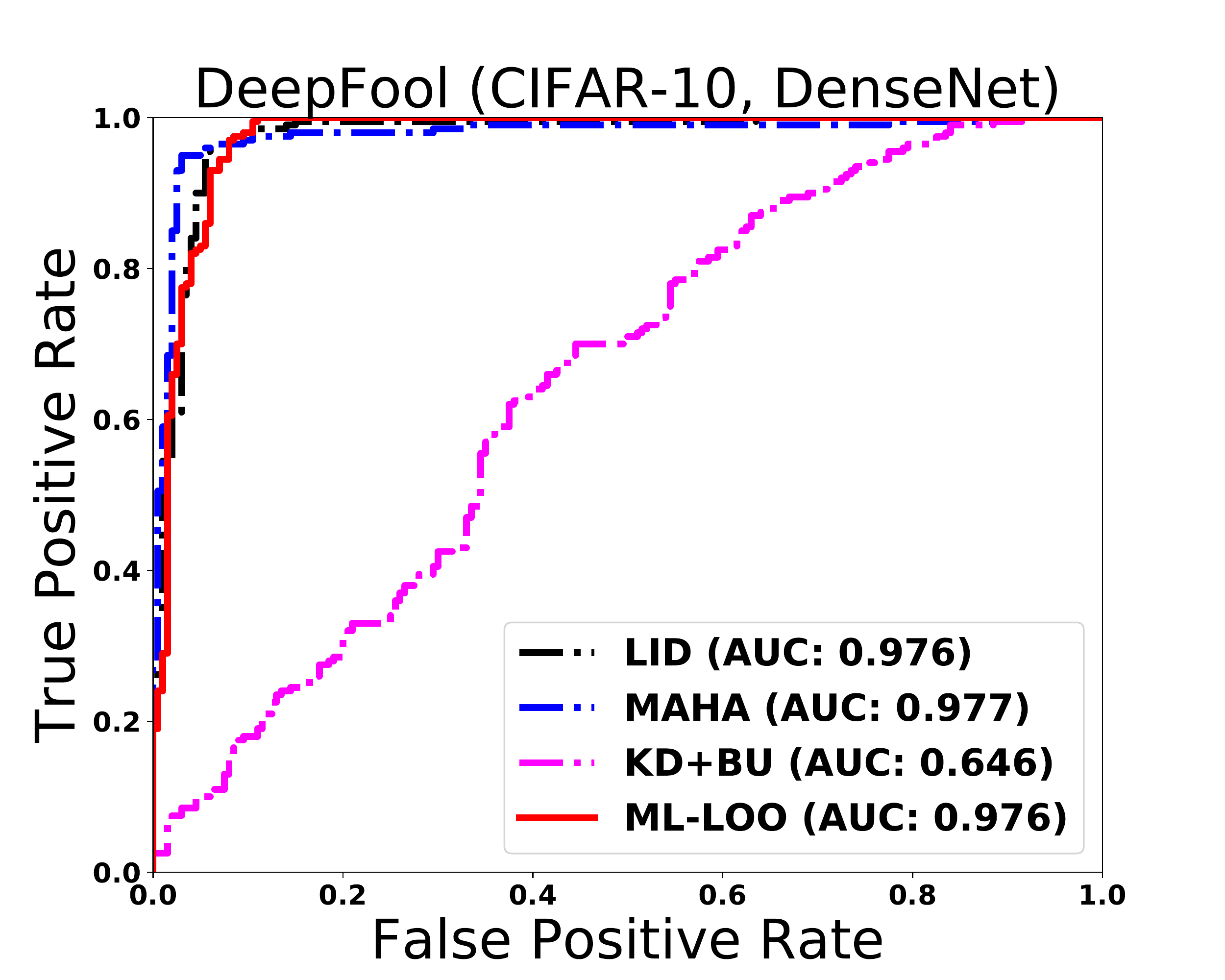} 
\includegraphics[width=0.3\linewidth]{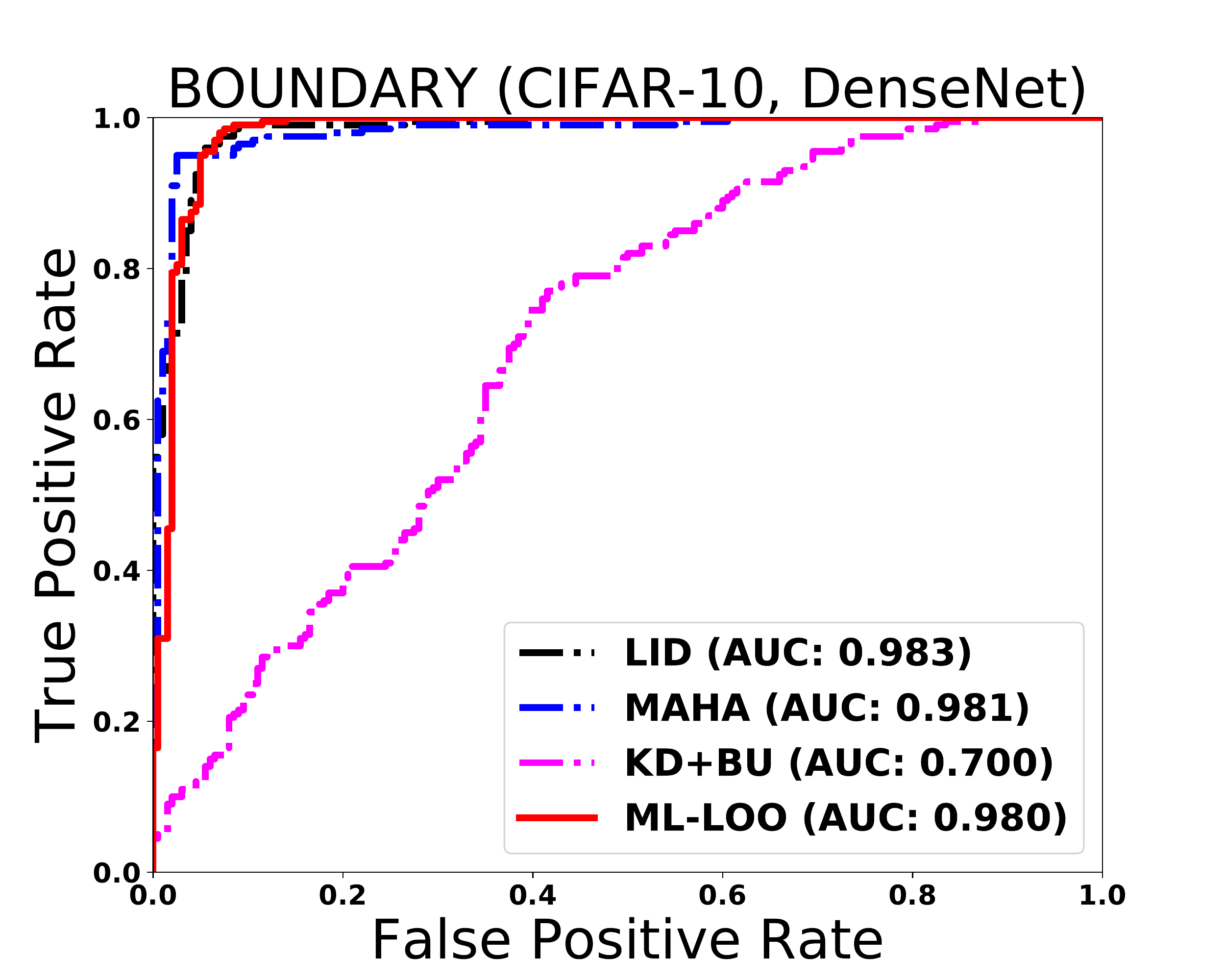} 
\includegraphics[width=0.3\linewidth]{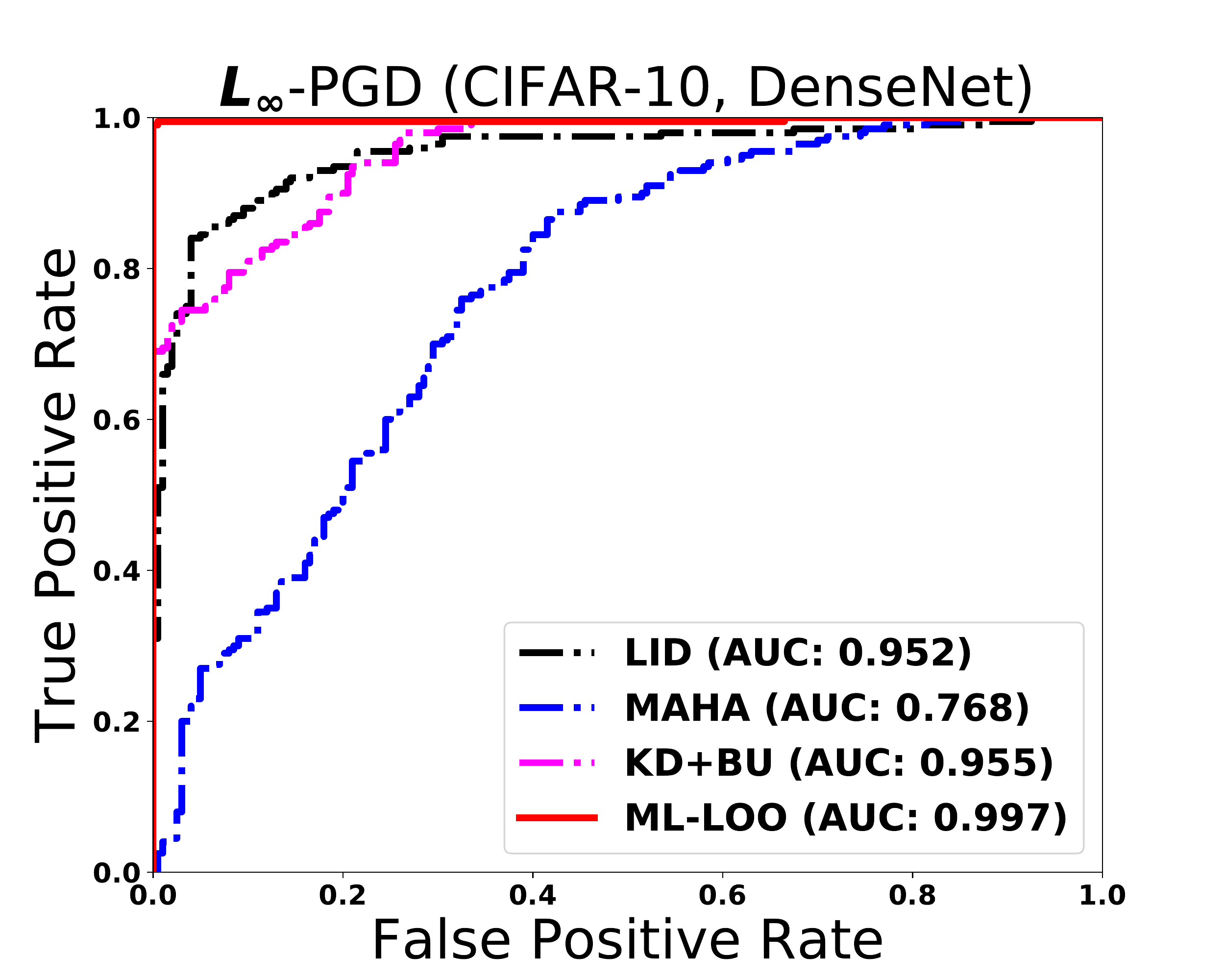}
\caption{ROC curves of detection methods on CIFAR-10 dataset with DenseNet}
\label{fig:CIFAR10DENSENET1}
\end{figure} 

\begin{figure}[H]
\centering 
\includegraphics[width=0.3\linewidth]{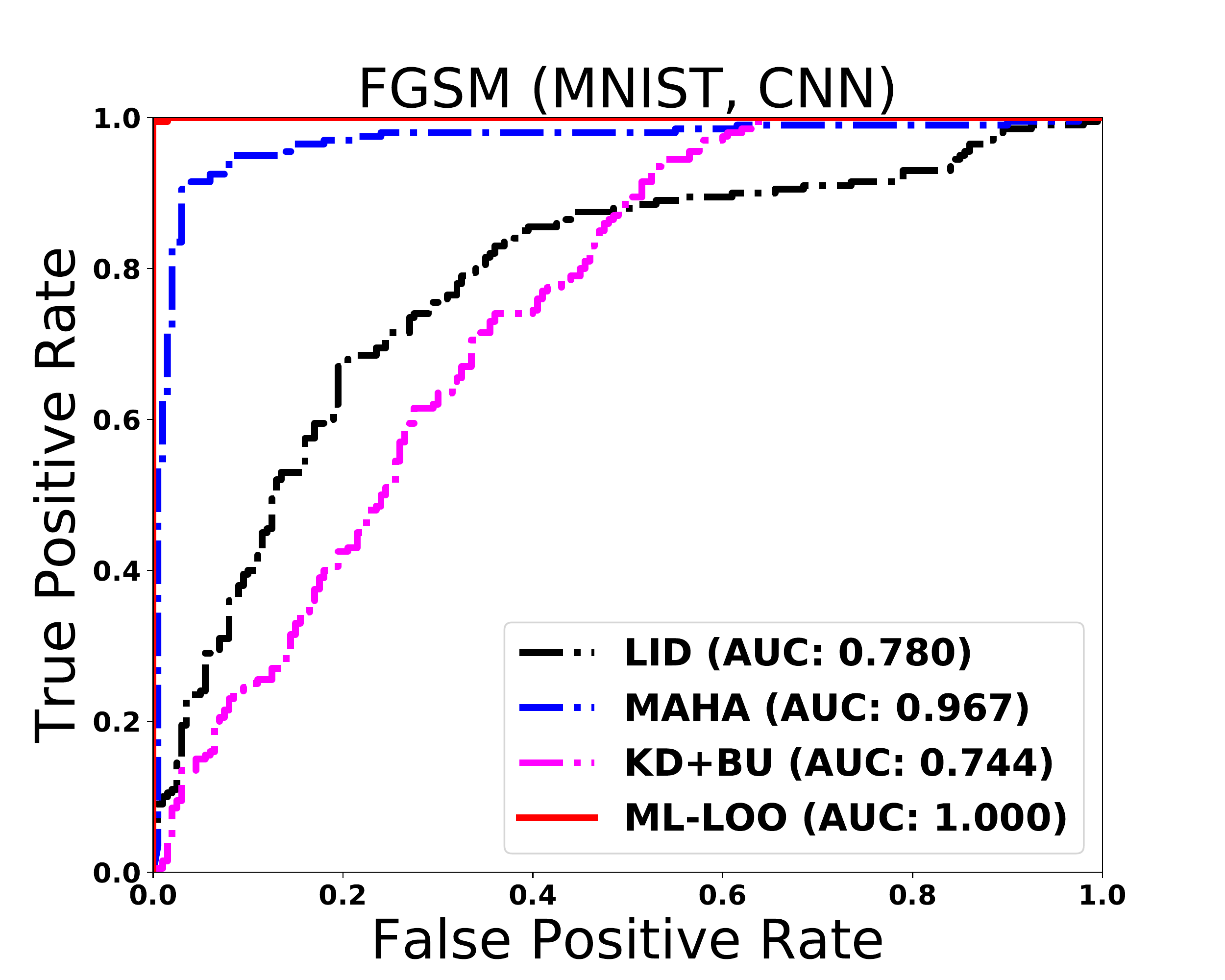} 
\includegraphics[width=0.3\linewidth]{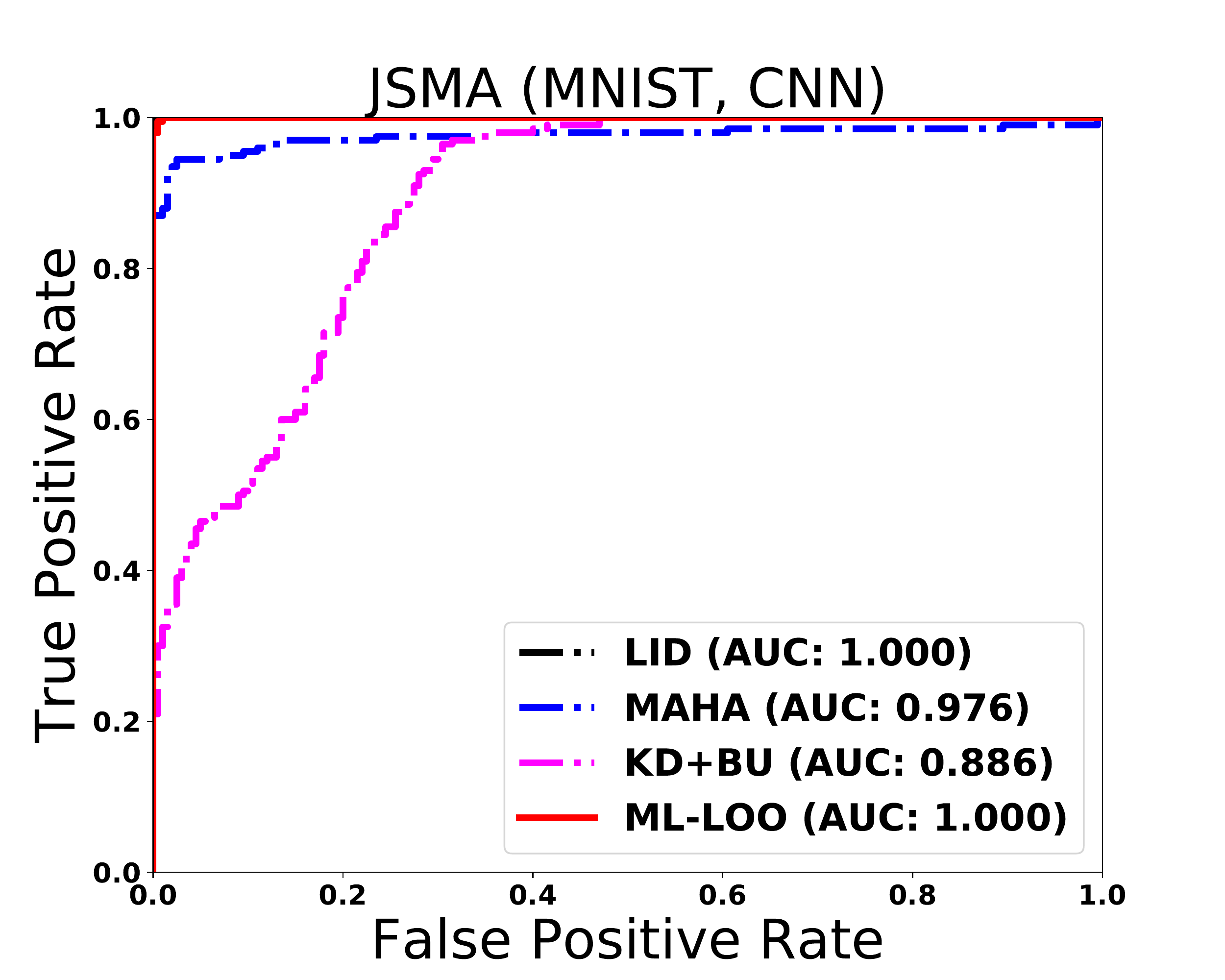} 
\includegraphics[width=0.3\linewidth]{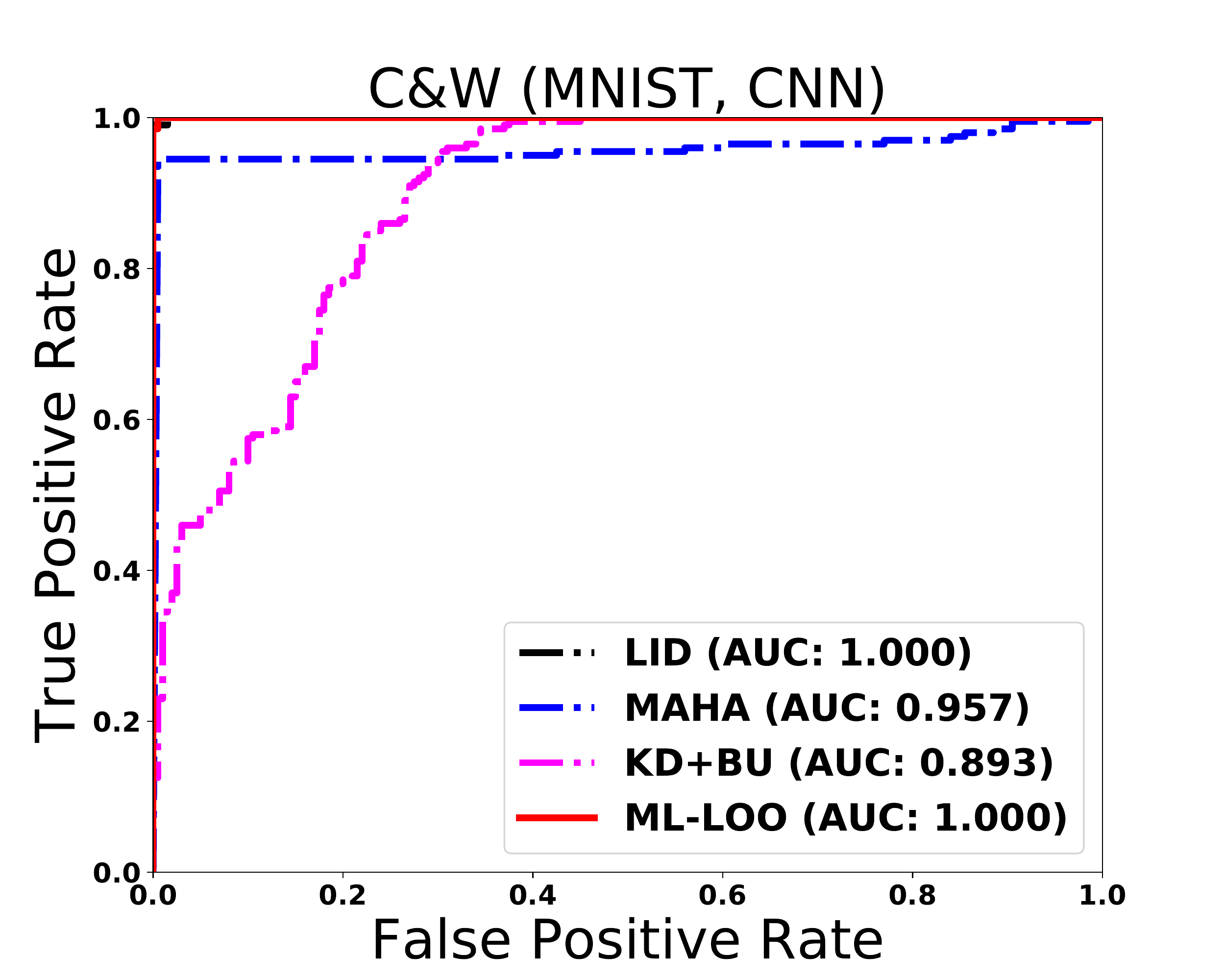} 
\includegraphics[width=0.3\linewidth]{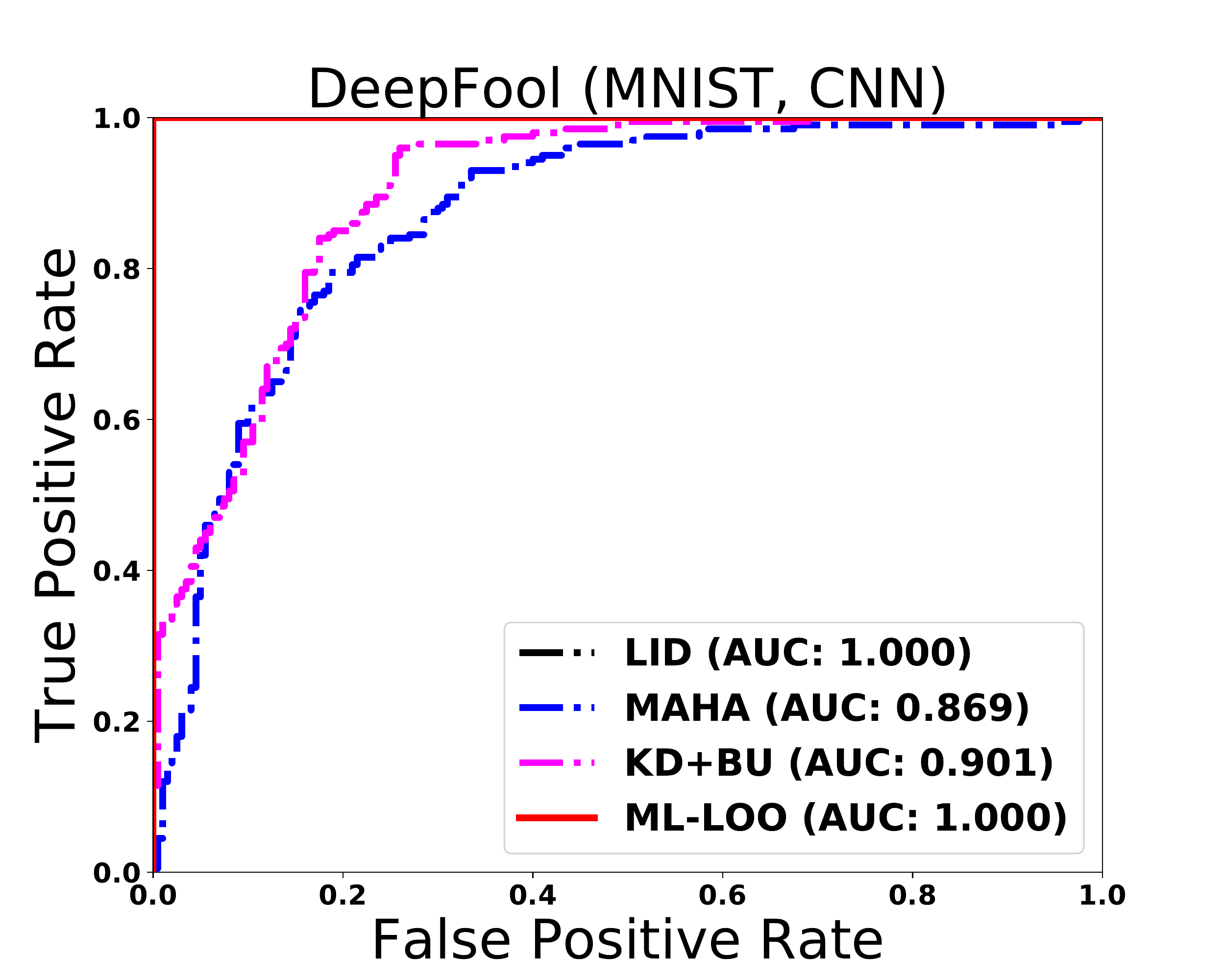} 
\includegraphics[width=0.3\linewidth]{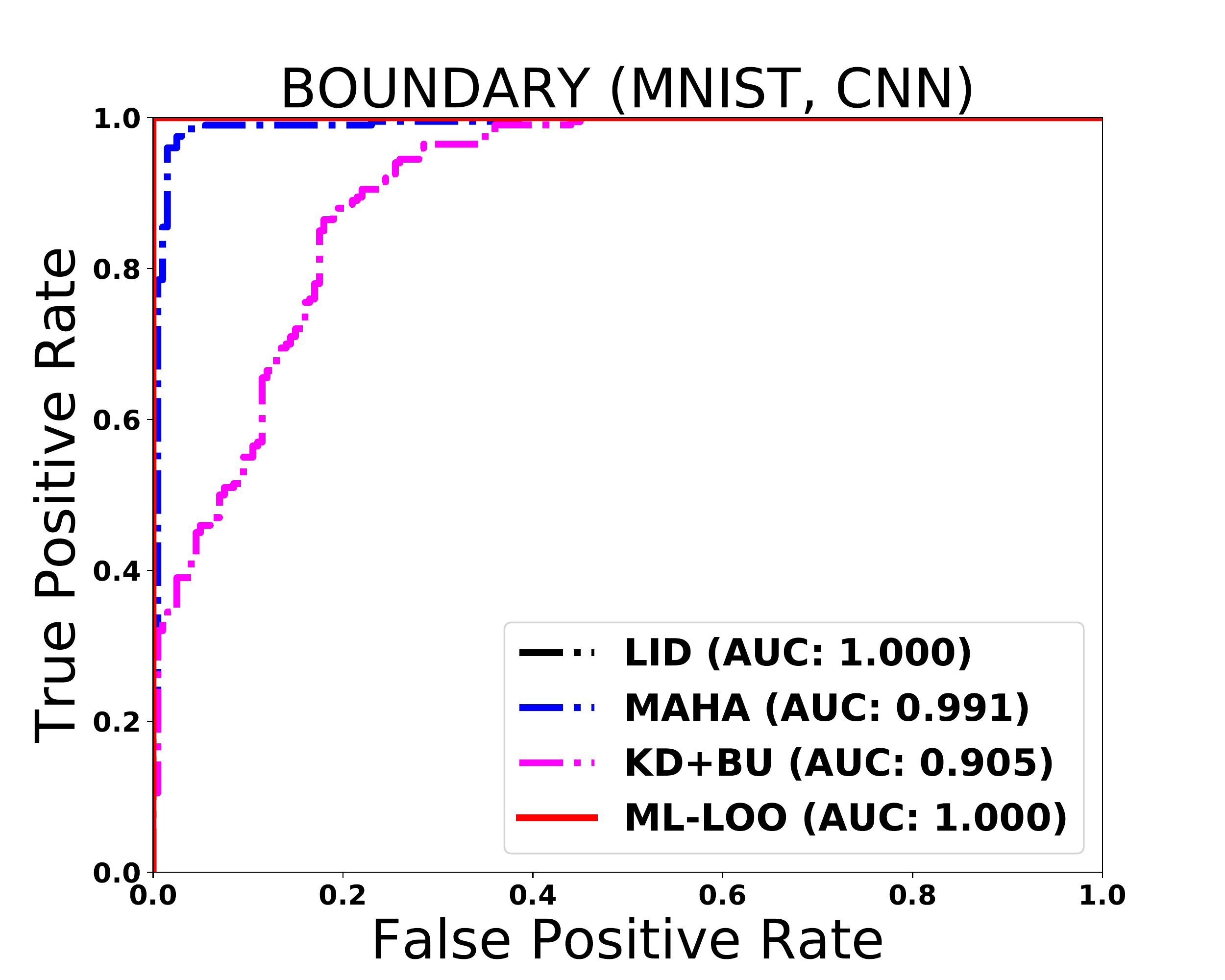} 
\includegraphics[width=0.3\linewidth]{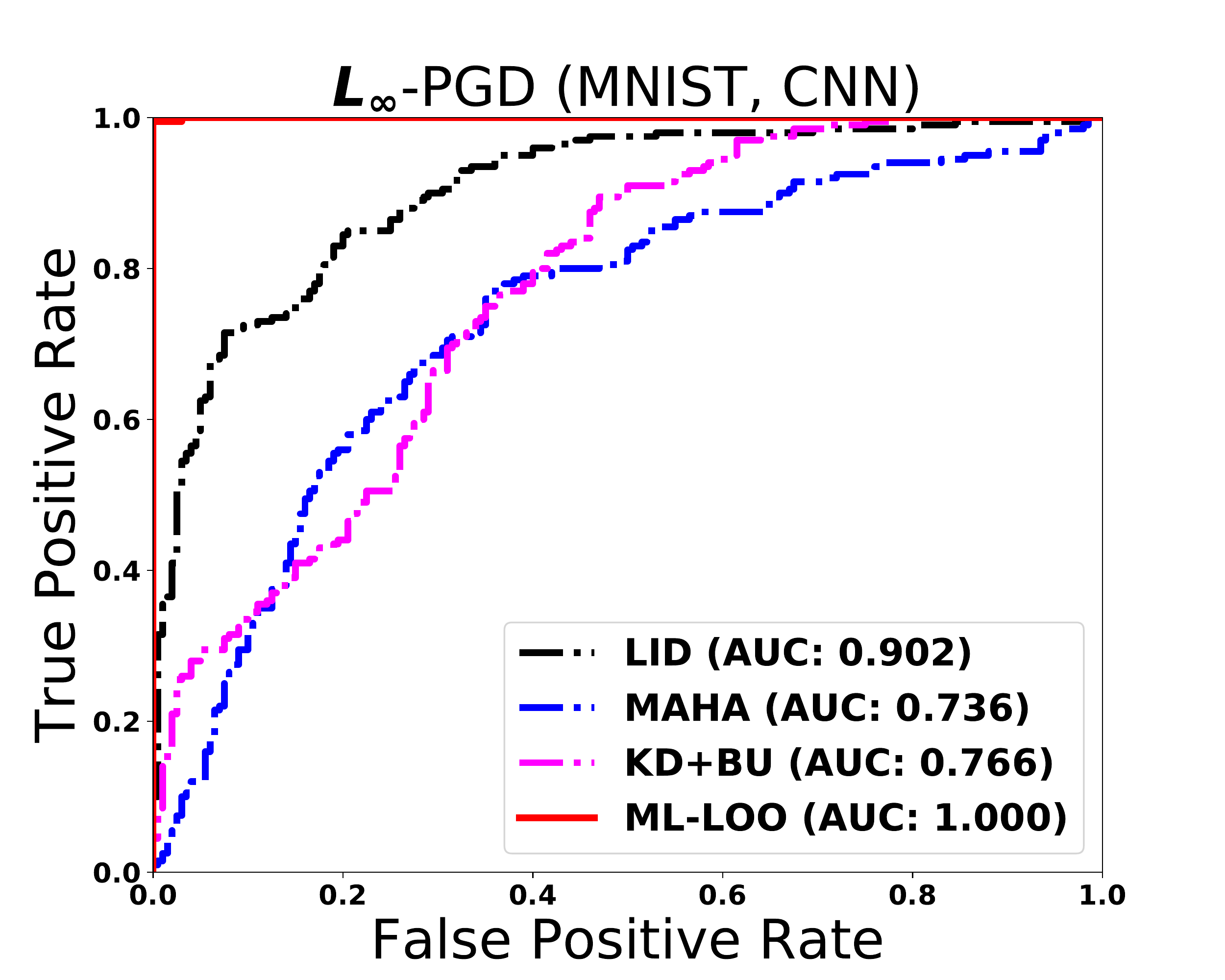}
\caption{ROC curves of detection methods on MNIST dataset with CNN}
\label{fig:MNISTSCNN1}
\end{figure}

\begin{figure}[H]
\centering 
\includegraphics[width=0.3\linewidth]{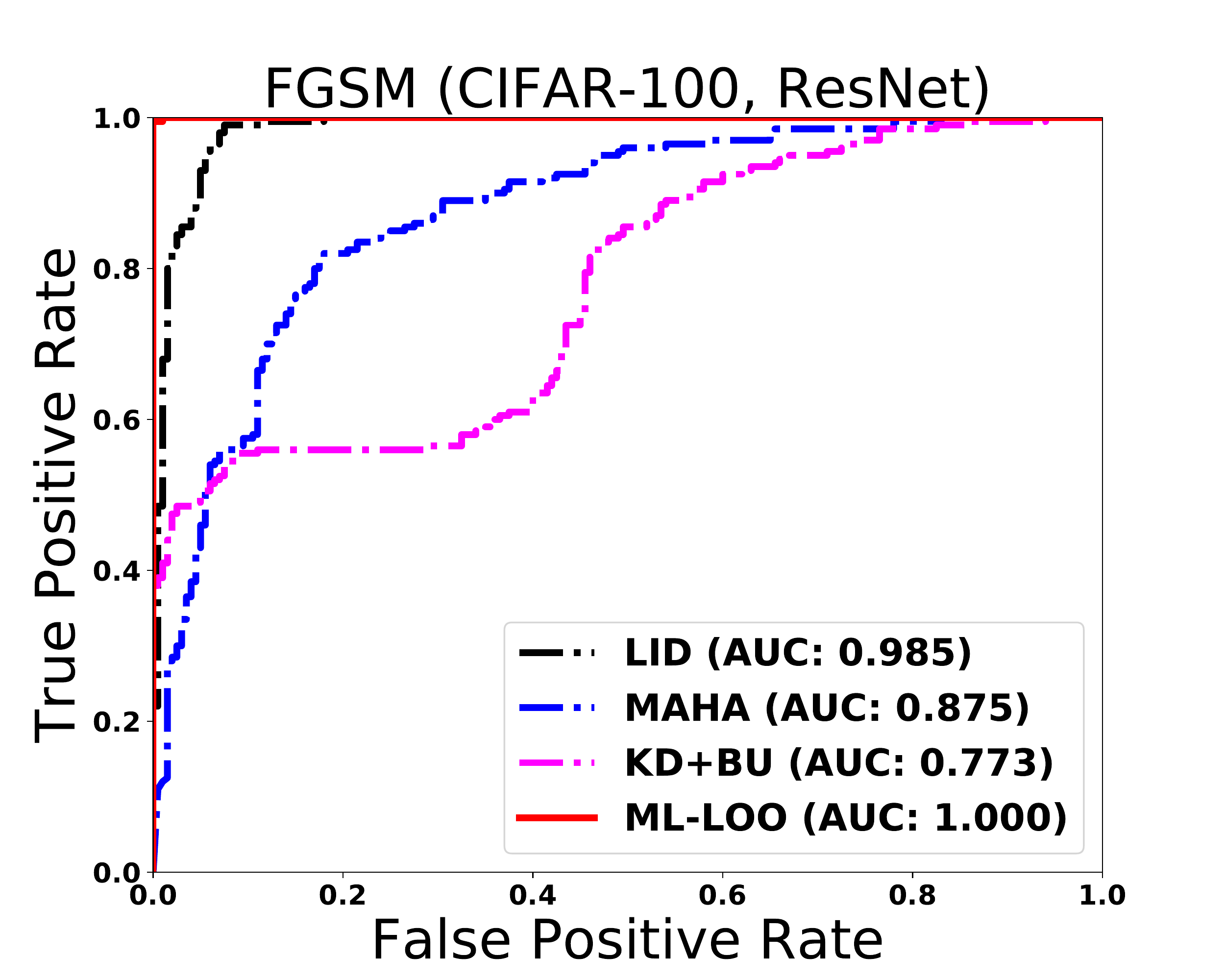} 
\includegraphics[width=0.3\linewidth]{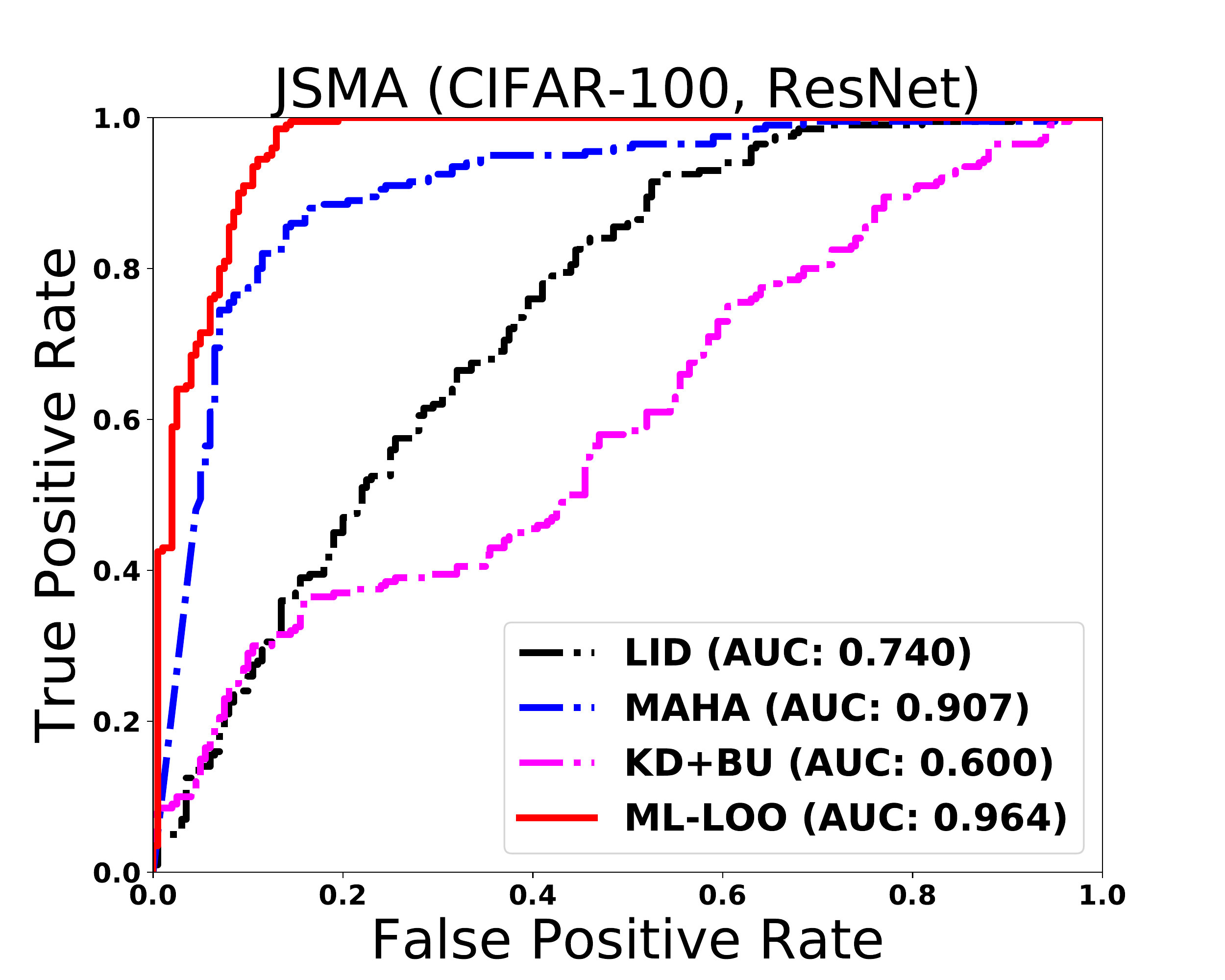} 
\includegraphics[width=0.3\linewidth]{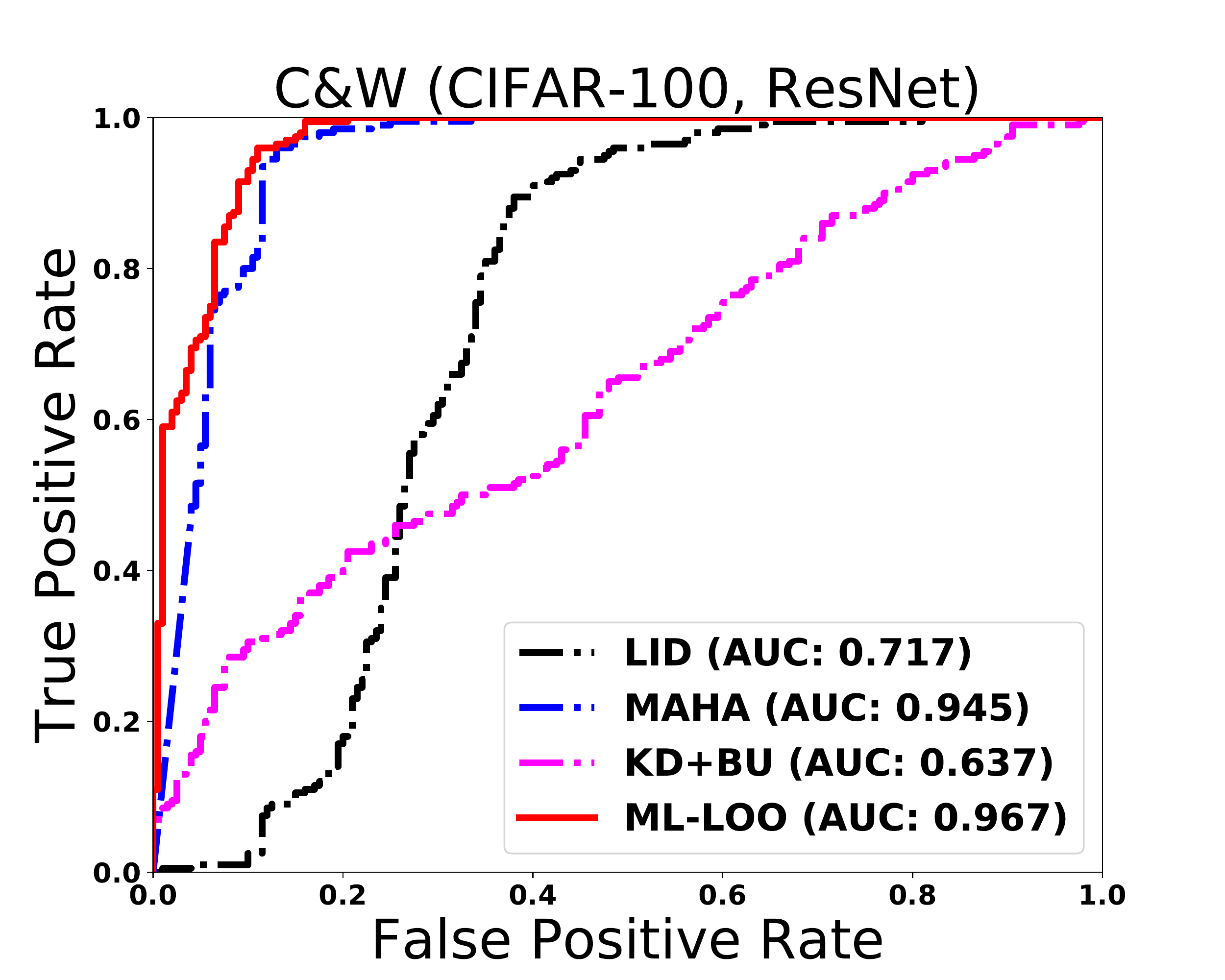} 
\includegraphics[width=0.3\linewidth]{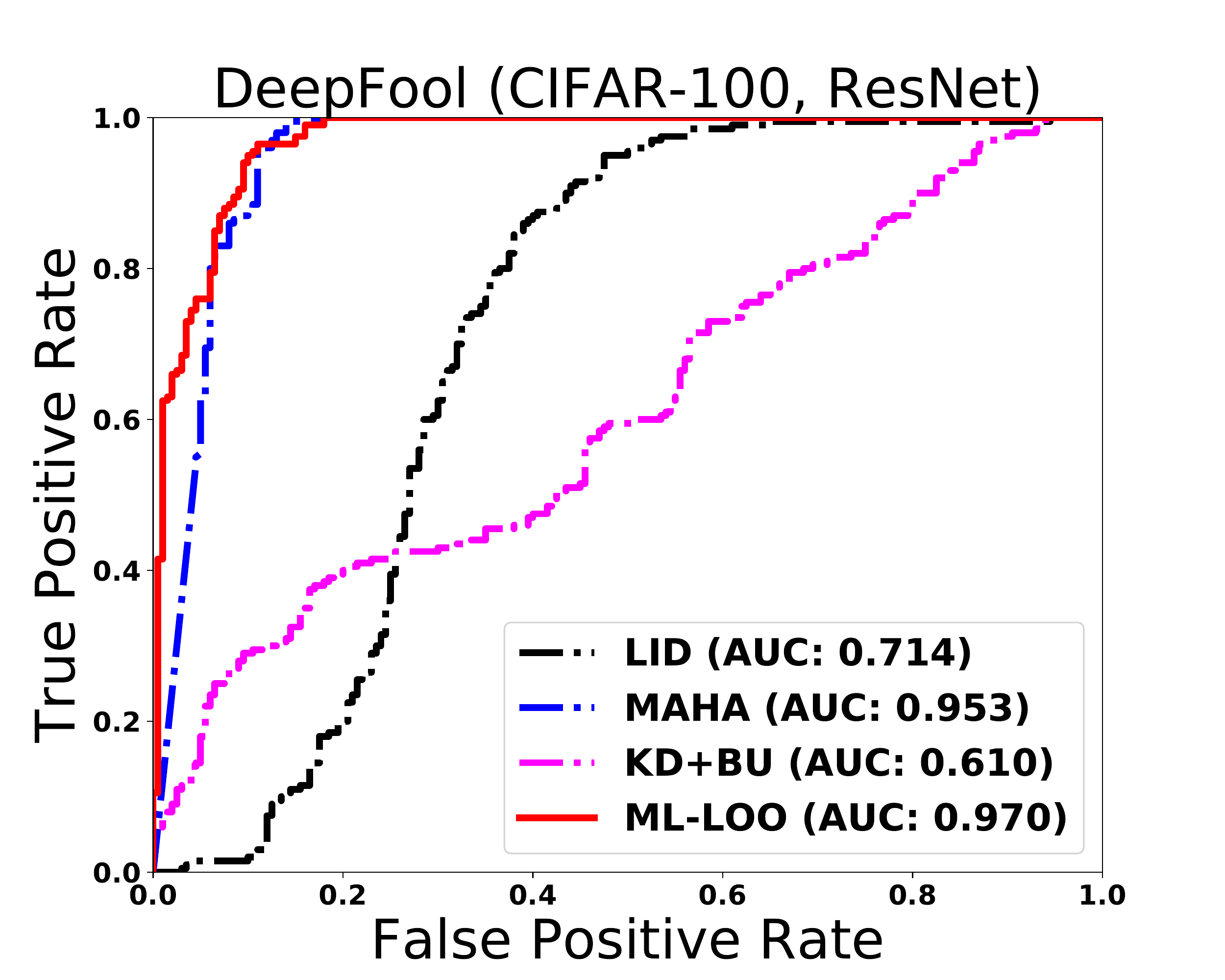} 
\includegraphics[width=0.3\linewidth]{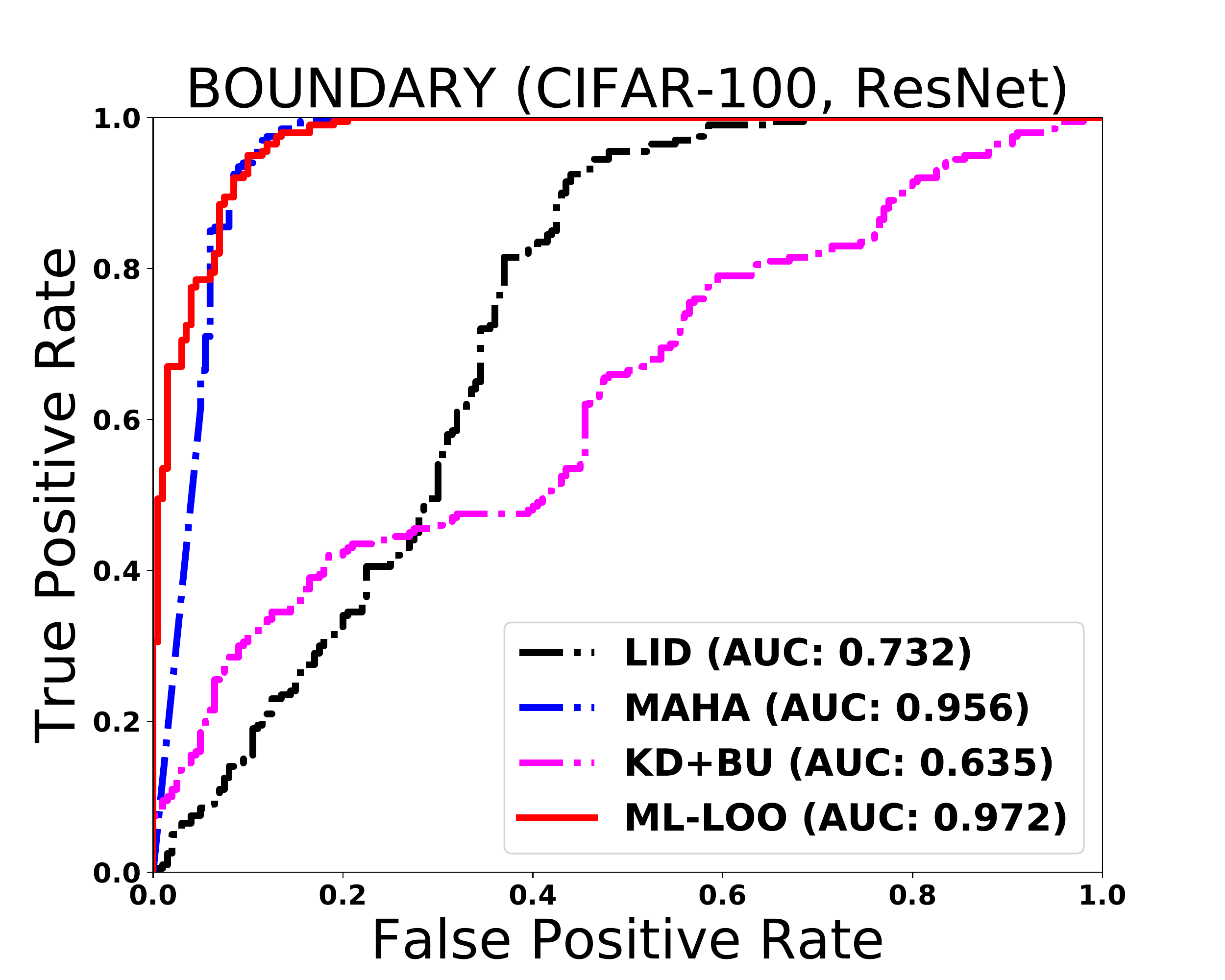} 
\includegraphics[width=0.3\linewidth]{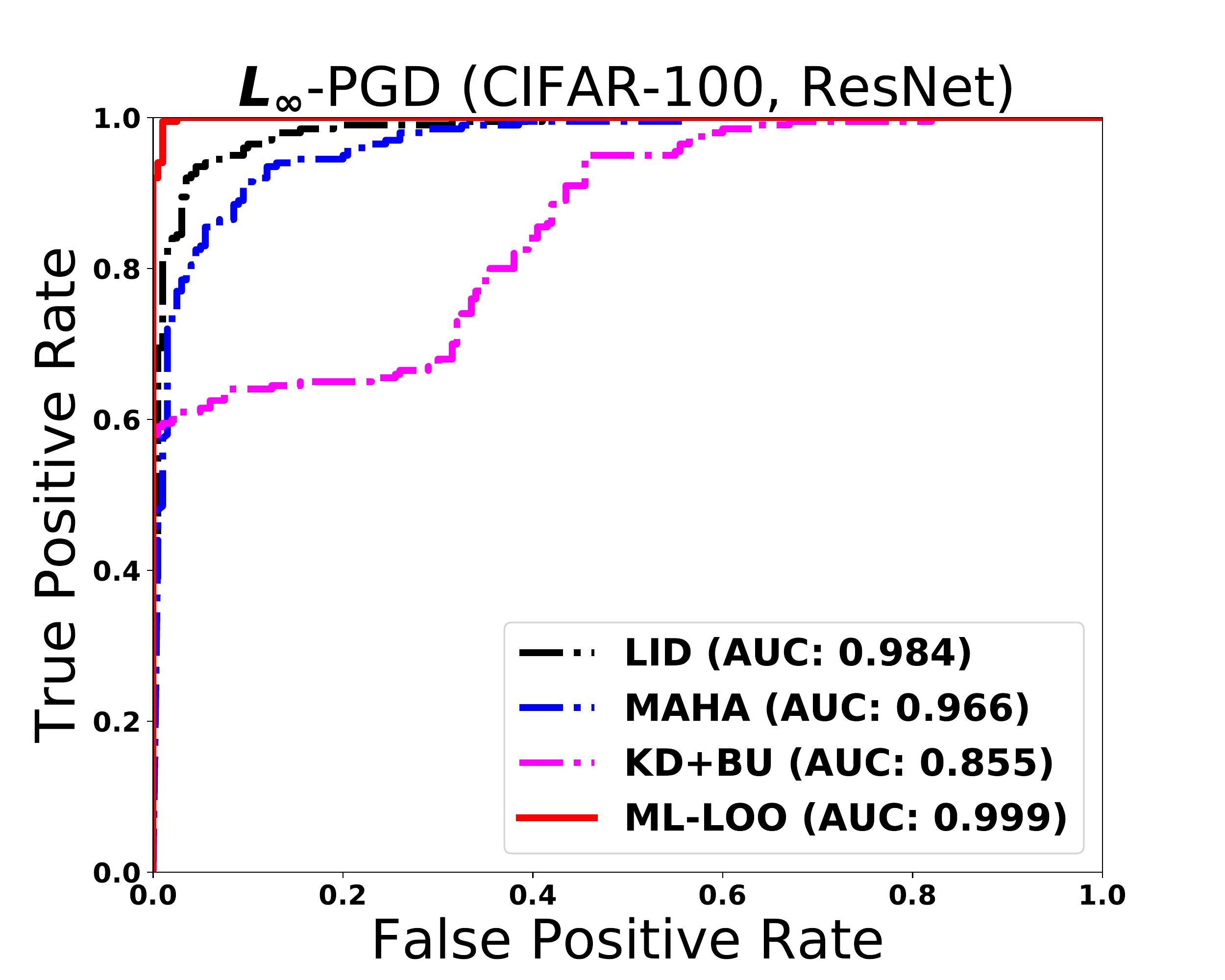}
\caption{ROC curves of detection methods on CIFAR-100 dataset with ResNet}
\label{fig:CIAFR100RESNET1}
\end{figure}

\begin{figure}[H]
\centering 
\includegraphics[width=0.3\linewidth]{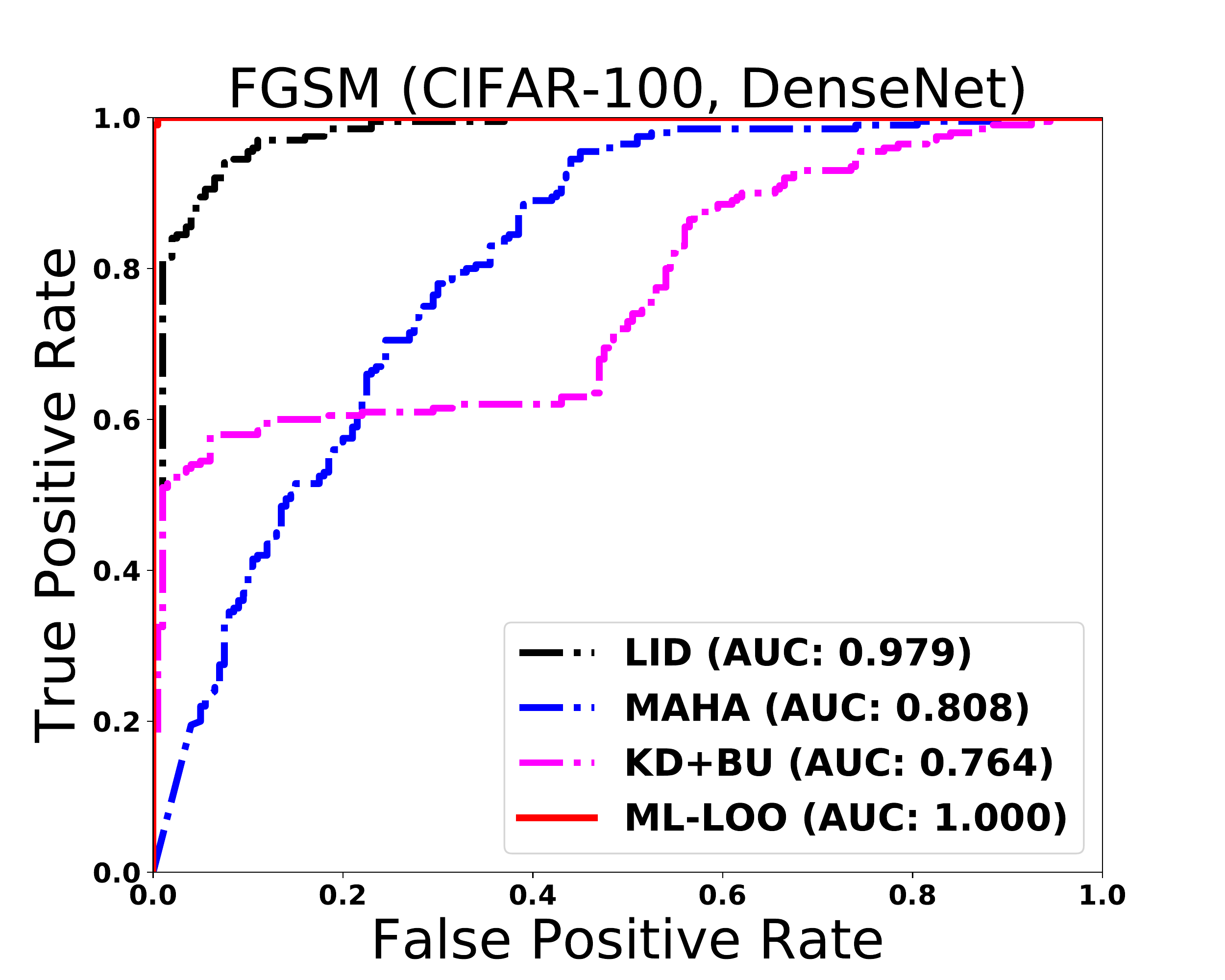} 
\includegraphics[width=0.3\linewidth]{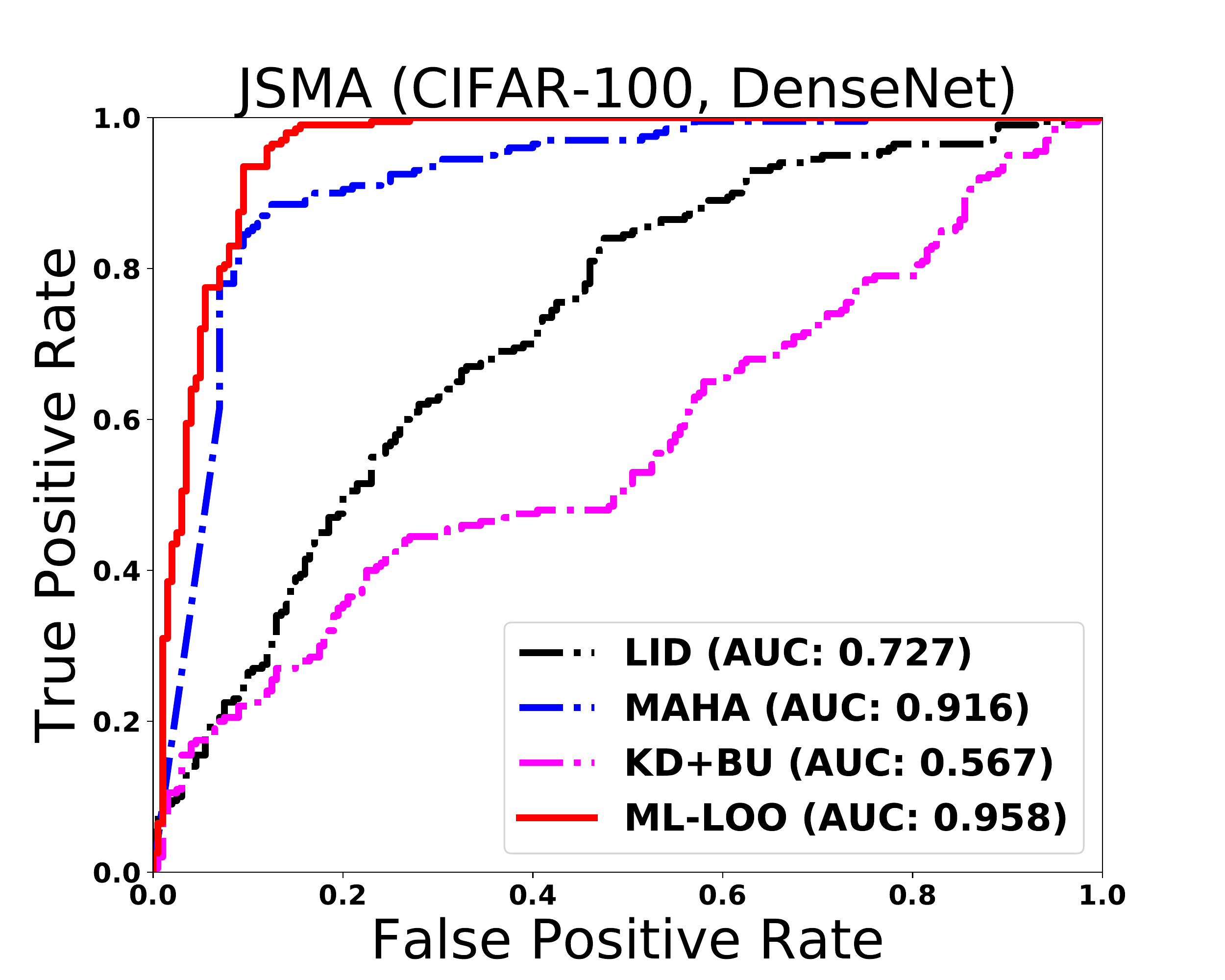} 
\includegraphics[width=0.3\linewidth]{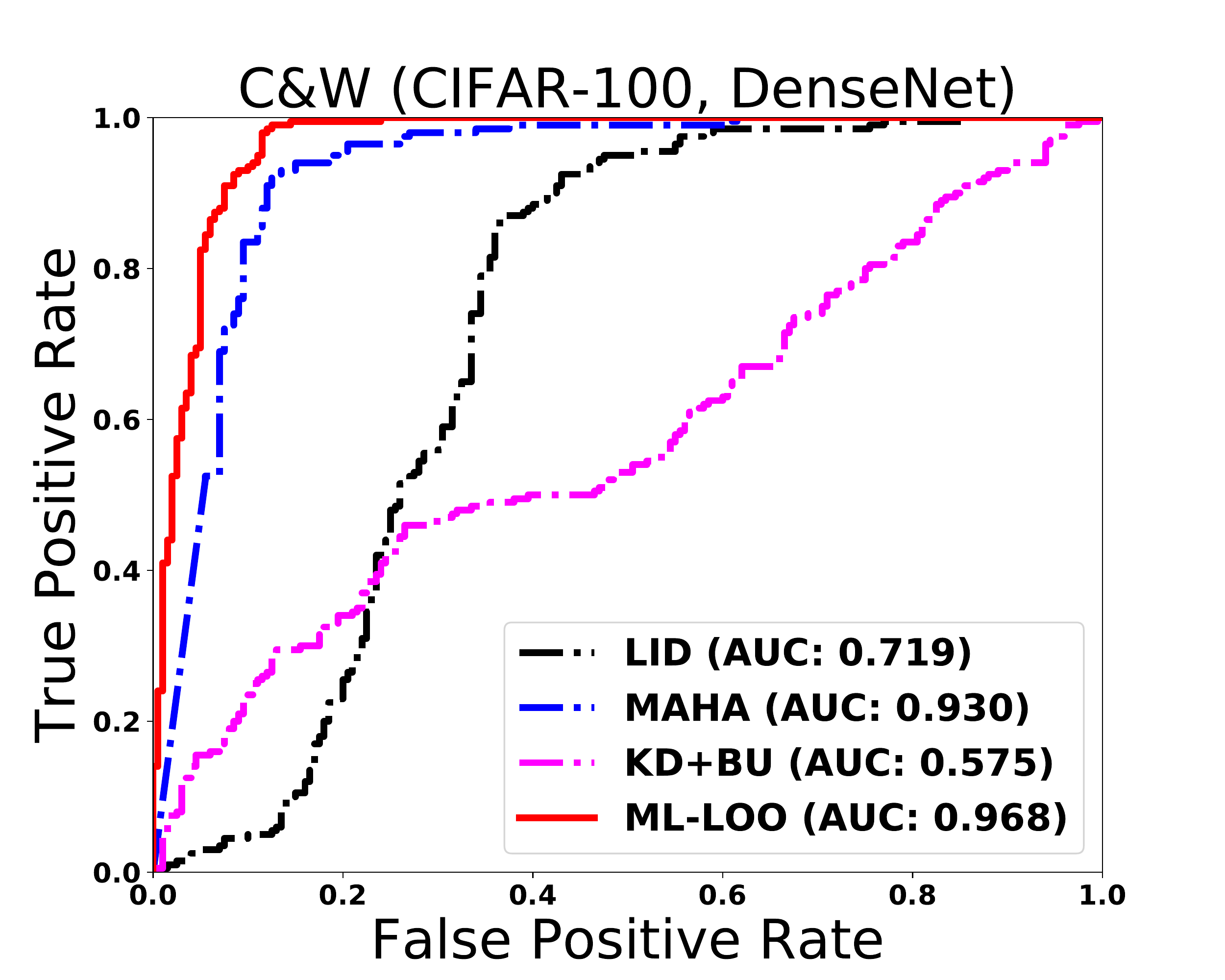} 
\includegraphics[width=0.3\linewidth]{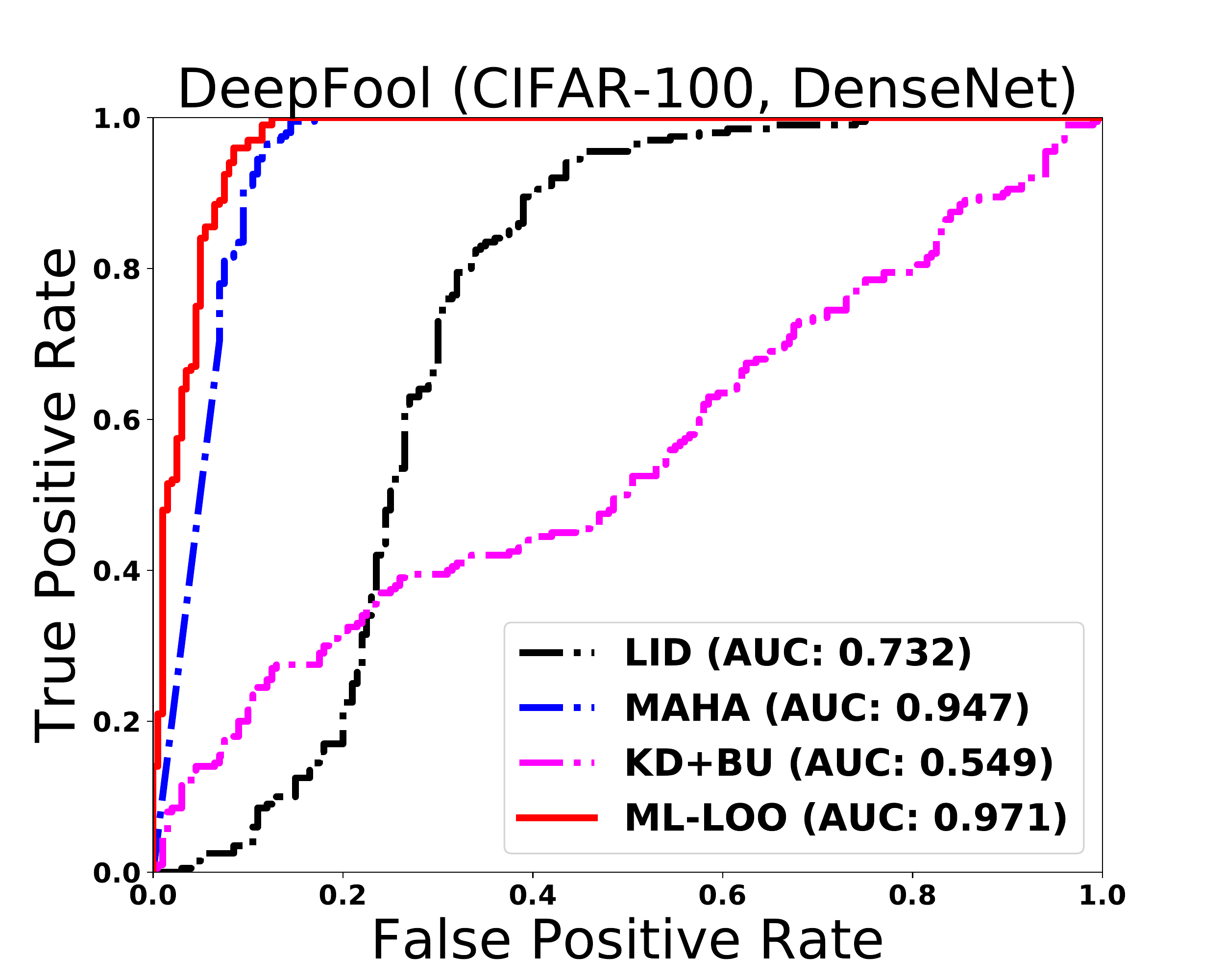} 
\includegraphics[width=0.3\linewidth]{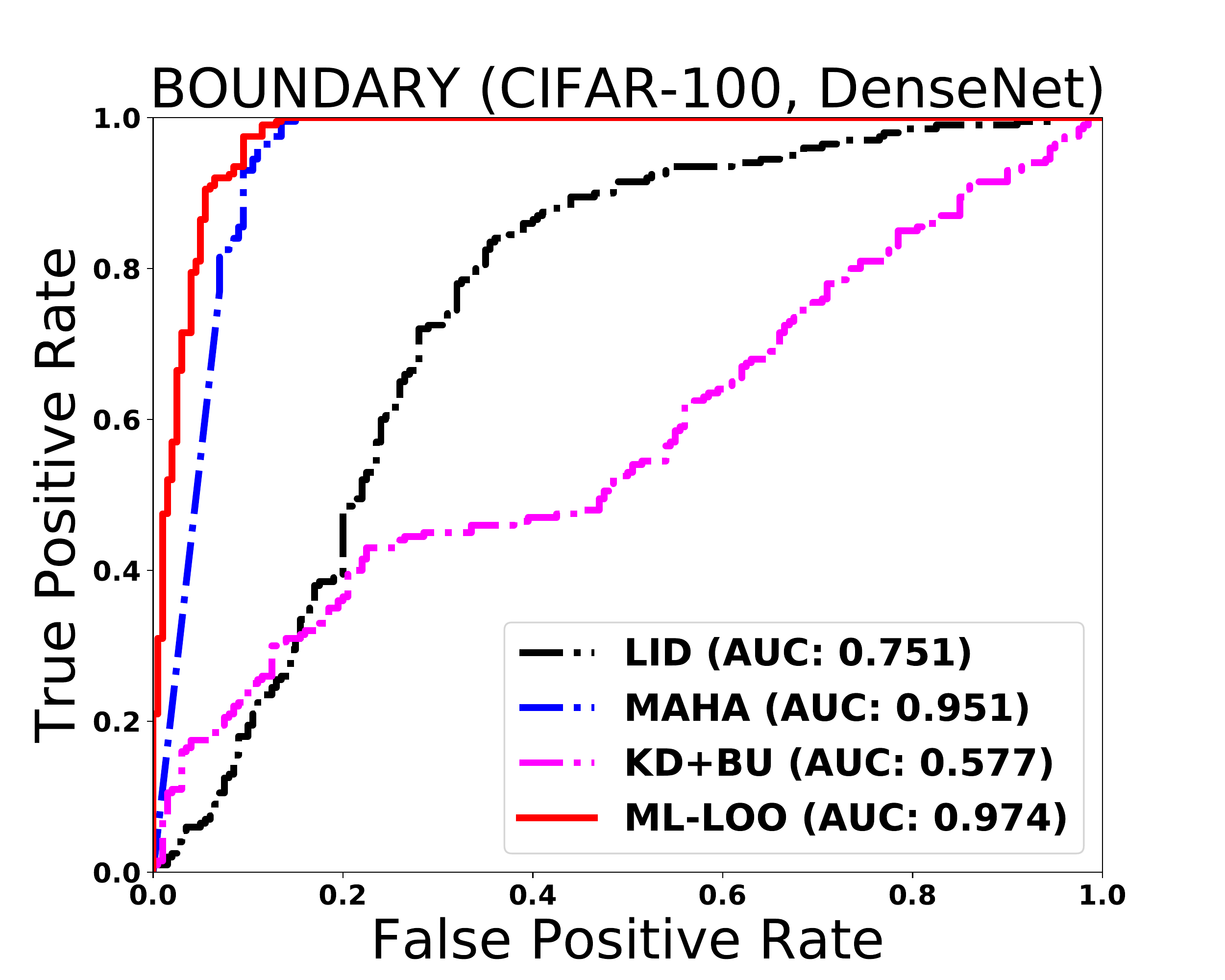} 
\includegraphics[width=0.3\linewidth]{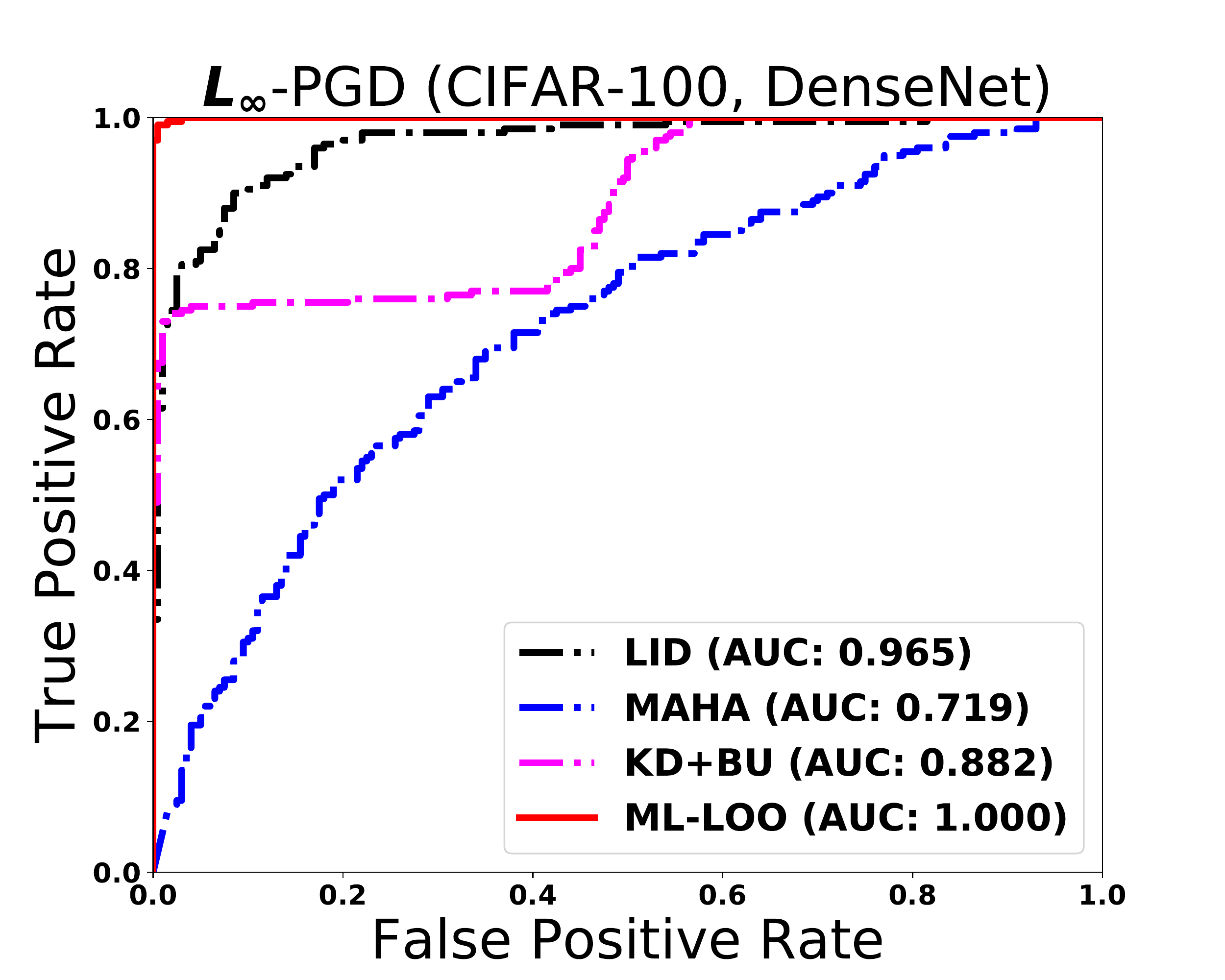}
\caption{ROC curves of detection methods on CIFAR-100 dataset with DenseNet}
\label{fig:CIAFR100DENSENET1}
\end{figure}

\end{document}